%% file: neurips_2026.tex
\documentclass{article}

\PassOptionsToPackage{numbers,compress,sort&compress}{natbib} 
\usepackage[preprint]{neurips_2026}
\usepackage{longtable}
\usepackage[utf8]{inputenc}
\usepackage[T1]{fontenc}
\usepackage{hyperref}
\usepackage{url}
\usepackage{booktabs}
\usepackage{amsfonts}
\usepackage{amssymb}
\usepackage{amsthm}
\usepackage{nicefrac}
\usepackage{microtype}
\usepackage{graphicx}
\usepackage{multirow}
\usepackage{subcaption}
\usepackage{siunitx}
\usepackage[ruled,vlined,linesnumbered]{algorithm2e}
\usepackage{float}
\usepackage[table]{xcolor}
\usepackage{wrapfig}
\usepackage{siunitx} 
\usepackage{amsmath}
\usepackage{fvextra}
\usepackage{booktabs}      
\usepackage{multirow}      
\usepackage{subcaption}    
\usepackage[table]{xcolor} 
\usepackage{makecell}
\usepackage{mathtools} 
\newcommand{\sub}[1]{\textsubscript{\fontsize{4.5pt}{5pt}\selectfont #1}}

\newcommand{\ourmethod}{\texttt{VAR-Q}}
\newcommand{\ourmethodandspace}{\texttt{VAR-Q} }
\definecolor{advantagegreen}{HTML}{E8F5E9}
\definecolor{LightRed}{RGB}{255, 214, 214}

\title{\ourmethod: Tuning-free KV Cache Quantization for Visual Autoregressive Image and Video Generation}

\author{%
  \textbf{Boxun Xu\textsuperscript{1,\dag}}\thanks{Correspondence: \texttt{boxunxu@ucsb.edu}}%
  \textbf{\quad
    Jiaji Lu\textsuperscript{1,\dag}\quad
    Yuxuan Yin\textsuperscript{1}\quad
    Zihu Wang\textsuperscript{1}} \\[2pt]
  \textbf{Ziyue Liu\textsuperscript{1}\quad
    Yu Wang\textsuperscript{1}\quad
    Zirui Liu\textsuperscript{2}\quad
    Peng Li\textsuperscript{1}} \\[6pt]
  \textsuperscript{1}University of California, Santa Barbara\quad
  \textsuperscript{2}University of Minnesota \\[2pt]
  \textsuperscript{\dag}Equal contribution.
}

\begin{document}

\maketitle
\begin{abstract}

Visual autoregressive generation spans two dominant paradigms: next-scale prediction for images and video, and next-frame prediction for video.
Both decode block-by-block rather than token-by-token, yielding KV caches whose statistics diverge from those of language models. At scale, KV memory exceeds the weights and limits large-scale deployment caused by a large amount of visual tokens. 
Our analysis of these caches reveals three findings. 
First, a small set of \emph{massive-value channels} carries the functional load: ablating these outlier channels collapses generation, whereas ablating a matched set of inlier channels leaves quality intact. 
Second, the outliers organize along two axes, feature and block, with magnitudes persistent within each cell and shifting across cells. 
Third, this structure arises from two independent mechanisms: projection bias and RoPE-induced scattering. Guided by these findings, we propose \ourmethod, a tuning-free KV-cache quantization framework that groups along the same two axes. The generation loop provides block boundaries for free, removing the need for outlier detection or calibration. Paired with bit-packing kernels for runtime memory reduction, \ourmethodandspace delivers low-bit quantization across nine models spanning both paradigms, cutting KV memory by up to 78.5\% with negligible quality degradation and enabling 6$\times$ larger inference batches.

\end{abstract}


\begin{figure}[htbp]
    \centering
    \includegraphics[width=0.95\columnwidth]{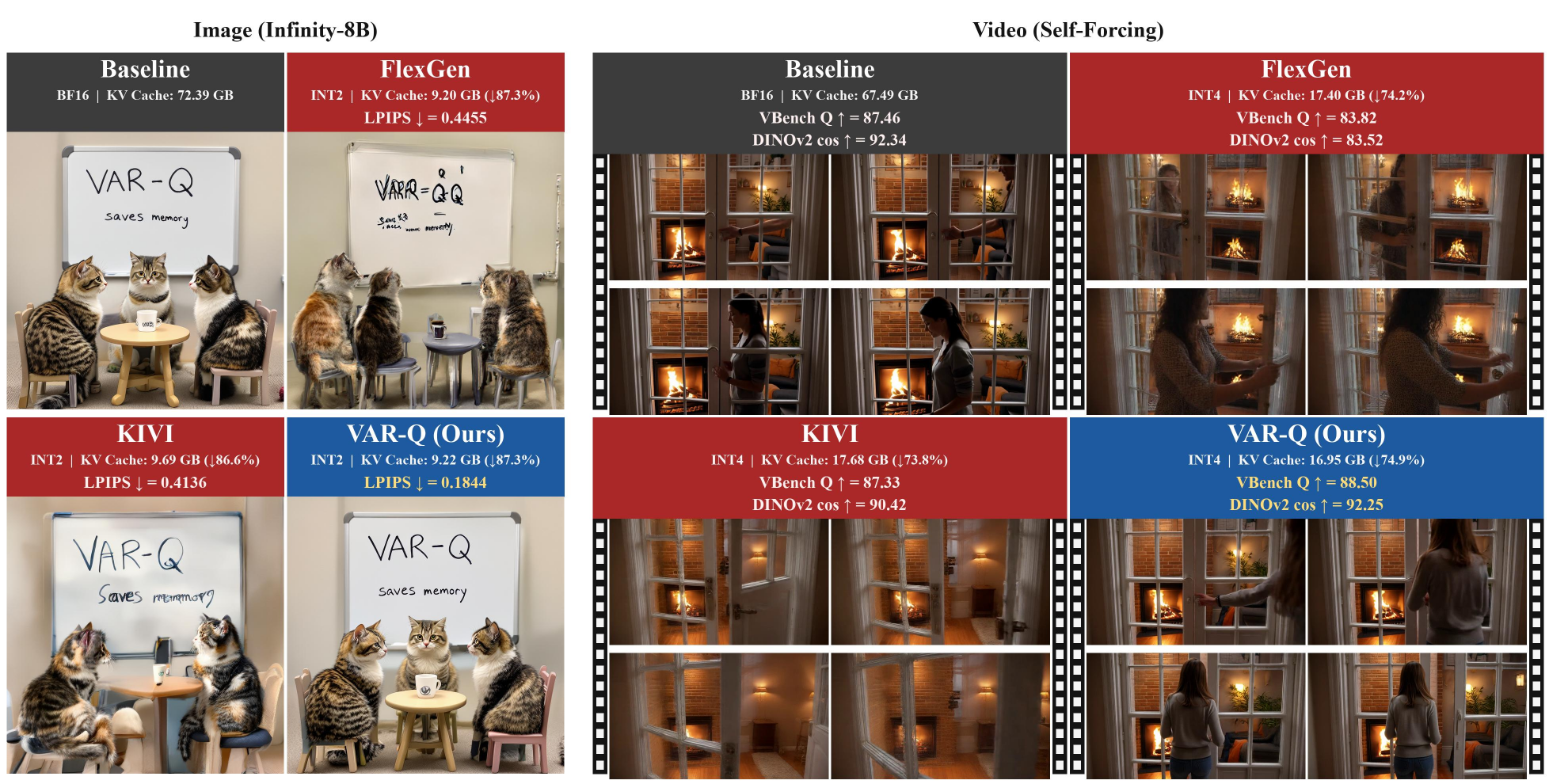}
    \caption{\textbf{Generation quality under quantized KV caches.} Top: On Infinity-8B at INT2, VAR-Q reproduces the BF16 rendering of text and scene layout that FlexGen and KIVI corrupt. On Self-Forcing at INT4, VAR-Q tracks the camera trajectory properly.}
    \label{fig:top}
\end{figure}

\section{Introduction}

Visual generation has consolidated around two autoregressive paradigms. \emph{Next-scale prediction} decodes a multi-scale token map one resolution at a time, and now spans both image~\citep{tian2024visual, han2025infinity} and video~\citep{infinitystar} synthesis. \emph{Next-frame prediction} decodes a video one frame at a time, conditioned on the past frames~\citep{huang2025selfforcing, yang2025longlive}. Both decode in semantically coherent \emph{blocks}, a scale or a frame, rather than individual tokens. As representation dimensions increase and generation horizons reach hundreds of frames, the Key-Value (KV) cache accumulated across these blocks rivals or exceeds the model weights themselves, easily saturating single-GPU memory. KV-cache quantization has become the standard remedy for language models~\citep{liu2024kivi, hooper2024kvquant, liu2024spinquant, su2025rotatekv, baek2025fireq, yao2025vecinfer, li2024snapkv, zhang2023h2o, kang2024gear}, where per-channel grouping for keys and per-token grouping for values now compress caches to 2--4 bits with negligible perplexity loss. None of this work targets visual AR. The cache structure of these models is not understood, and recipes calibrated to language models do not transfer: applied directly, they collapse generation.

The KV cache of a visual AR model is not a uniform memory of past tokens. Less than $1\%$ of its outlier channels carry its functional load in a numerical manner. In LongLive~\citep{yang2025longlive}, zeroing just the top 0.5\% of channels by magnitude collapses generation entirely, while zeroing ten times as many inlier channels leaves generation quality largely intact. The same asymmetry holds across Infinity~\citep{han2025infinity}, InfinityStar~\citep{infinitystar}, and SelfForcing~\citep{huang2025selfforcing}, in both image and video synthesis, and at every layer we measure. Whatever the cache stores for future retrieval lives in a thin, magnitude-distinguished subspace.

This subspace is not a scattered set of outliers but a structured one, organized along two orthogonal axes. A small set of channels carries large magnitudes throughout decoding, producing persistent stripes along the channel axis. Within each block magnitudes are stationary, and across block boundaries they shift sharply, producing bands along the block axis. Both axes have architectural causes. 
On the channel axis, key outliers arise from a composition of three multiplicative sources: heavy-tailed $\mathbf{W}_K$ row norms seed an outlier baseline, post-projection RMSNorm gain $\gamma_K$ amplifies it where present, and RoPE pair rotation~\citep{su2024roformer} redistributes energy at the pair level. 
On the block axis, position-varying RoPE projects channel magnitudes through a sinusoidal modulation, especially at high-frequency channels in the block index, producing the cross-block shifts directly from the rotation schedule rather than from any change in input. Value caches exhibit the same two-axis structure with weaker per-axis attribution, sufficient to justify the same grouping rule without a separate mechanism.

However, existing quantization rules align with at most one axis: KIVI~\citep{liu2024kivi} groups keys per-channel and values per-token, leaving the block axis unprotected on both. To this end, we propose \textbf{VAR-Q}, a tuning-free KV quantization framework that groups the cache jointly along channels and blocks, with one quantization unit per channel and block pair. The block boundaries come for free from the generation loop itself, without detection, calibration, or retraining. We pair the algorithm with bit-packing kernels that turn nominal bit-width reduction into actual memory savings on current GPUs.
Across nine visual AR models spanning both paradigms, VAR-Q matches the BF16 baseline at INT4 on both next-scale image and next-scale video generation, and remains usable on Infinity-8B down to INT2. On next-frame video, it sustains quality at INT4 and outperforms KV quantization work transferring from LLM domains directly such as KIVI~\citep{liu2024kivi} and FlexGen~\citep{sheng2023flexgen}. Complementary to concurrent cache eviction~\citep{li2025scalekv, xu2025amskv} and attention sparsification~\citep{xi2025sparsevideogen} for visual generation, \ourmethod{} reduces KV memory by up to $78.5\%$ relative to BF16, and inference batch size grows by up to $6\times$ on a single NVIDIA A100 80GB.

\section{Background}
\label{sec:background}

We unify the two visual autoregressive paradigms under a common block-wise notation. \emph{Next-scale AR}~\citep{tian2024visual, han2025infinity, infinitystar} emits one resolution of a multi-scale token map per step; \emph{next-frame AR}~\citep{huang2025selfforcing, yang2025longlive} emits one or more frames per step, and we treat each frame as a separate block. We refer to the tokens emitted at step $i$ as block $i$. After $B$ steps, the key cache is
\begin{equation}
\mathbf{K} = \big\{\mathbf{K}^{(i)}\big\}_{i=1}^{B}, 
\qquad 
\mathbf{K}^{(i)} \in \mathbb{R}^{n_i \times D},
\end{equation}
where $n_i$ is the number of tokens in block $i$ and $D$ is the head-channel dimension; the value cache $\mathbf{V}$ is defined analogously. In next-scale AR, $n_i$ grows geometrically across a fixed number of scales; in next-frame AR, $n_i$ is constant and $B$ grows with the video length. 
Both caches are produced by linear projections $\mathbf{K}^{(i)} = f_K(\mathbf{x}^{(i)} \mathbf{W}_K)$ and $\mathbf{V}^{(i)} = \mathbf{x}^{(i)} \mathbf{W}_V$, where $f_K$ denotes any model-specific post-projection transforms applied to keys.

\section{Methodology}
\label{sec:method}
The element-wise magnitudes of the KV cache form a two-axis grid over feature channels and decoding blocks, stable within cells and shifting sharply across block boundaries. A small subset of cells carries values orders of magnitude larger than the rest, and these dominate generation. Group-wise quantization must therefore align with both axes.
Section~\ref{sec:massive_values} establishes that the massive values are critical for generation through controlled ablation; Section~\ref{sec:structure} characterizes the two-axis grid; Section~\ref{sec:interpretability} attributes both axes to specific architectural sources; and Section~\ref{sec:algorithm} translates the grid into \ourmethod.


\subsection{Massive Values in KV Caches}
\label{sec:massive_values}

\paragraph{Outliers in the KV cache are functional, not just
numerical.}

\begin{wrapfigure}{r}{0.5\columnwidth}
    \centering
    \vspace{-18pt}
    \includegraphics[width=0.48\columnwidth]{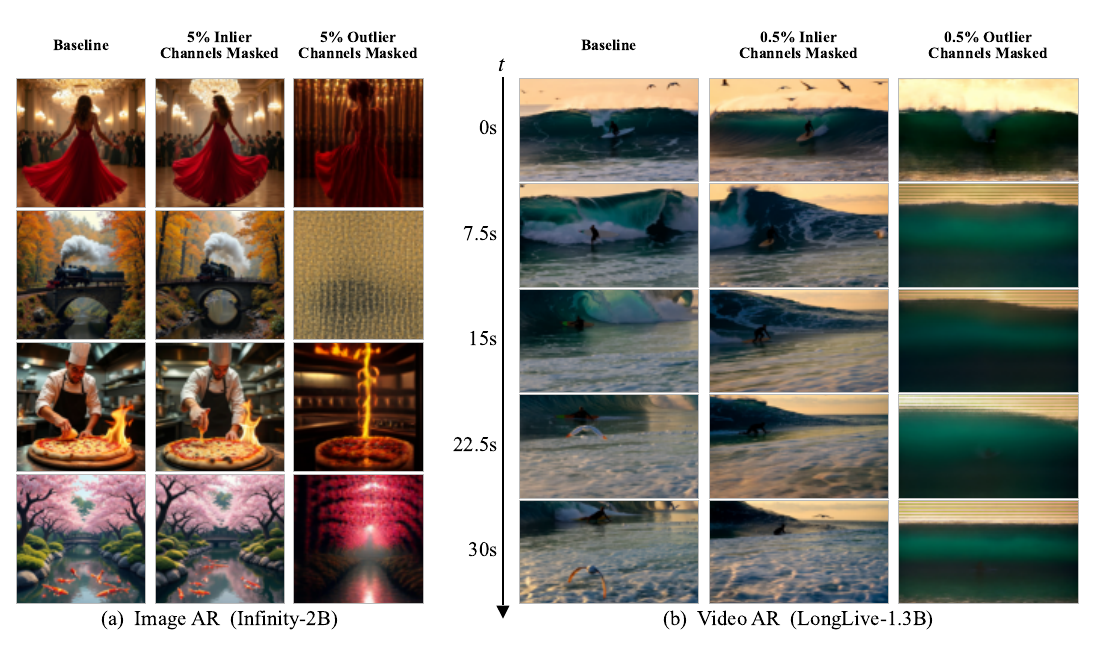}
    \caption{Generation Comparison under channel-wise outlier vs.\ inlier KV masking.}
    \label{fig:main_fig}
    \vspace{-14pt}
\end{wrapfigure}

\begin{figure}[t]
    \centering
    \includegraphics[width=\columnwidth]{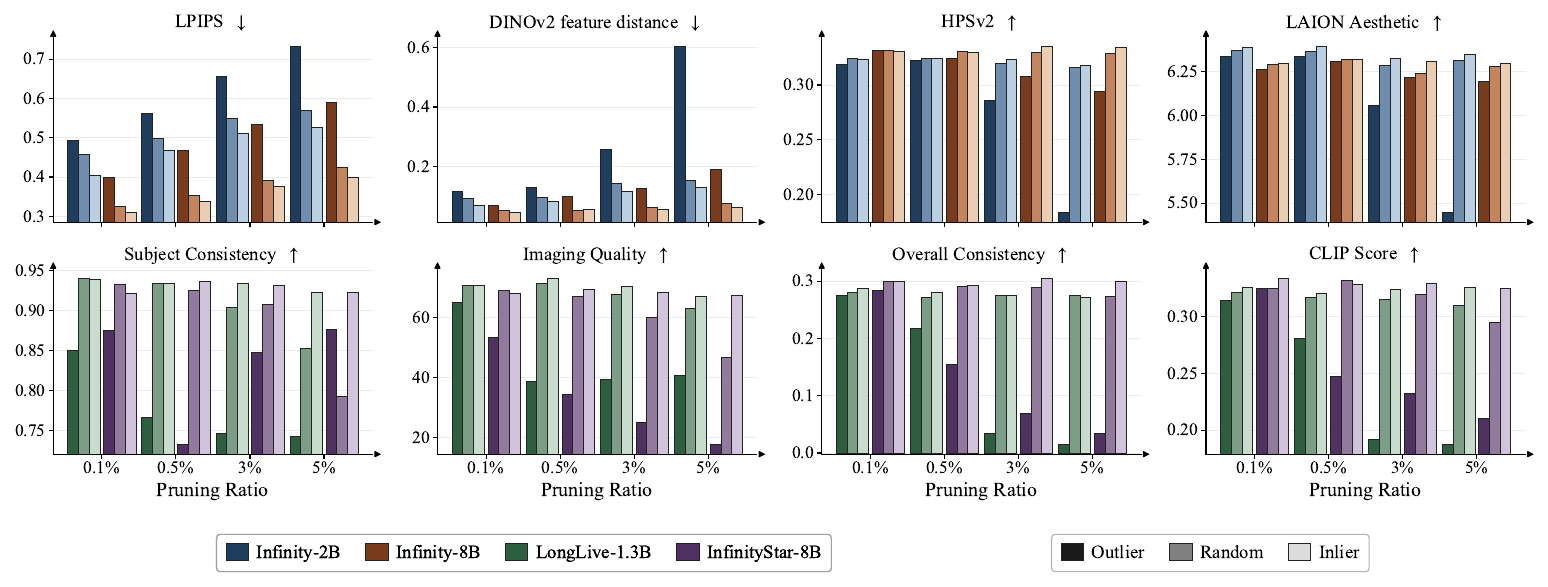}
    \caption{Channel-wise outlier, random, and inlier KV ablation on four visual AR models. Channels are ranked by maximum absolute value over cached tokens.}
    \label{fig:two_axis_ablation}
\end{figure}

Massive activations have been documented in the residual streams~\citep{sunmassive} 
and MLP weights~\citep{yu2024super} of language models, where direct ablations 
confirm them as functionally indispensable. KV cache outliers have been observed 
across LLMs~\citep{liu2024kivi, hooper2024kvquant} and VLMs~\citep{su2025akvq}, 
motivating protect-by-isolation strategies such as per-channel 
grouping~\citep{liu2024kivi}, rotation-based redistribution~\citep{ashkboos2024quarot}, 
and high-precision retention for dense outliers~\citep{hooper2024kvquant}. 
This protection establishes utility, not necessity: per-channel grouping and 
rotation succeed whether outliers carry load-bearing signal, dominate the 
quantization range of co-grouped channels, or both. Only direct ablation can 
adjudicate. Visual AR has had neither: no characterization of its KV cache 
outliers, no ablation of their function. Whether load-bearing outliers exist 
in this regime, and how they are organized across long visual sequences, 
remains open.


We test necessity directly through ablation. For each layer, we rank 
channels by their maximum magnitude across the cached sequence. The top 
$p$ percent form the \emph{outlier} set. From the remaining channels, 
we draw an equal-sized random sample to form the \emph{inlier} set. Each 
ablation zeros one set.
Figure~\ref{fig:main_fig} makes the contrast visible: on Infinity-2B at 
$p=5\%$ and LongLive-1.3B at $p=0.5\%$, outlier ablation collapses 
generation into uniform texture or solid blocks; inlier ablation leaves 
outputs intact. Figure~\ref{fig:two_axis_ablation} extends this to four 
models and eight metrics. On LongLive-1.3B, masking the top $0.5\%$ drops 
CLIP from $0.32$ to $0.28$, while ablating ten times as many inlier channels leaves CLIP at baseline ($0.32$). We call this small, magnitude-defined subset the \emph{massive-value channels}.
\footnote{A per-tensor decomposition of the ablation reveals that video 
models are more sensitive to $\mathbf{K}$ outliers and image models to 
$\mathbf{V}$ outliers (Appendix~\ref{app:kv_separate}).}



\subsection{Massive Values are Structured Along Two Axes in KV Caches}
\label{sec:structure}
Massive values impose a structural constraint on group-wise quantization: any group that mixes massive and regular cells must stretch its scaling factor to cover the massive cell, starving the regular cells of resolution. Resolving this requires group boundaries that follow the spatial distribution of massive values. In visual AR KV caches, that distribution is two-dimensional: massive values concentrate along a small set of channels and a small set of blocks. Both axes must be covered.

\input{Tables/MSE}

\begin{figure*}[t]
\centering
\includegraphics[width=\textwidth]{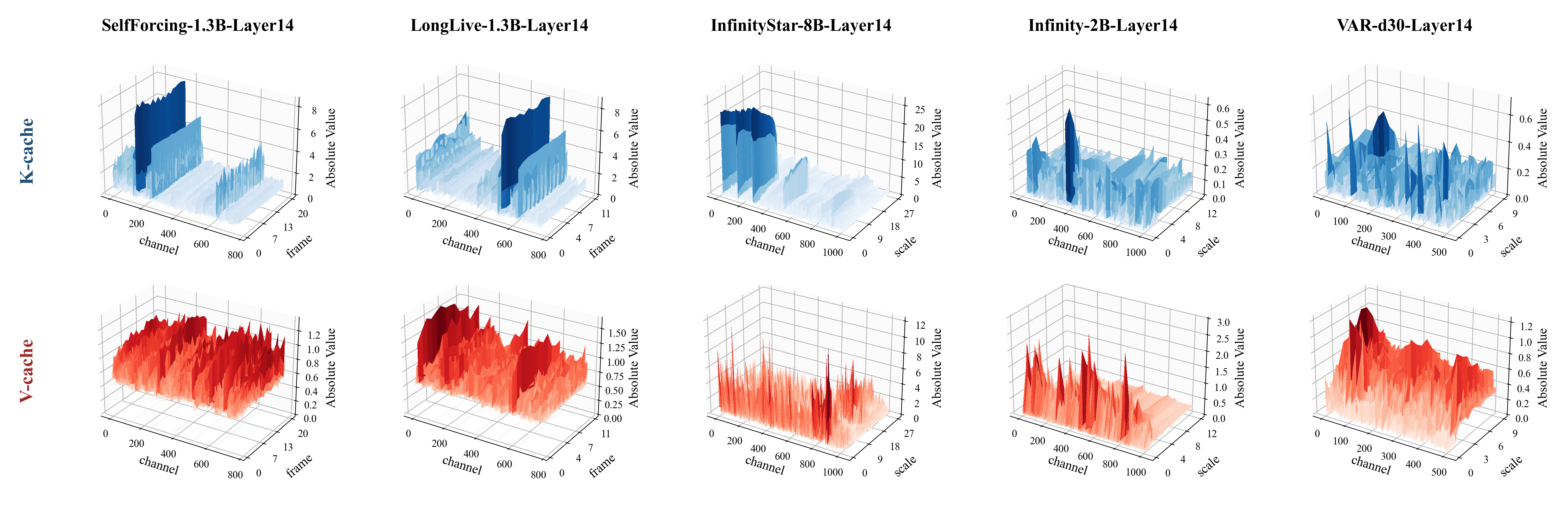}
\caption{Median absolute values of $K$ (top) and $V$ (bottom) caches across five visual AR models over the (channel, generation index) grid. Channel stripes persist along the generation axis in all models, and their magnitudes shift sharply at block boundaries. Color scales are per-panel.}
\label{fig:distribution}
\end{figure*}

\textbf{Channel stripes and block bands.}
The same structure appears in every model in Figure~\ref{fig:distribution}, organized along two orthogonal axes. A small subset of channels carries large magnitudes throughout the entire decoding trajectory, producing persistent stripes along the channel axis. Within each block, magnitudes are approximately stationary; across block boundaries, they shift sharply, producing bands along the block axis. The two axes are not redundant; they compose. Massive values live precisely at the intersections of high-magnitude channels and high-magnitude blocks. The block axis is specific to visual AR: language-AR KV caches form a homogeneous token stream with no analogous boundaries~\citep{liu2024kivi, hooper2024kvquant}, and prior LLM quantization methods accordingly exploit only the channel axis.

\textbf{Aligning with both axes minimizes error.}
Table~\ref{tab:grouping_err} measures how well four grouping rules contain the structure on Infinity-2B. Per-tensor collapses the entire cache into one group and incurs the largest error on every metric. Per-block aligns with the generation axis but lets each block mix high- and low-magnitude channels. Per-channel aligns with the hidden axis, the choice of KIVI~\citep{liu2024kivi} for LLM key caches, but forces a massive cell to share its scaling factor with regular cells of the same channel across all blocks. Only per-(channel, block) aligns with both axes, placing each massive cell in a group that contains only other massive cells from the same channel and the same block. Errors fall monotonically as more axes are covered, and the joint rule wins on every column: $\epsilon_O$ from $32.32\%$ to $10.13\%$, $\epsilon_K$ from $24.01\%$ to $10.04\%$. 

\subsection{Why Do Massive Values Emerge in KV Caches of Visual Autoregressive Models?}
\label{sec:interpretability}

\begin{wrapfigure}{l}{0.3\textwidth}
  \vspace{-\intextsep}
  \centering
  \includegraphics[width=0.3\textwidth]{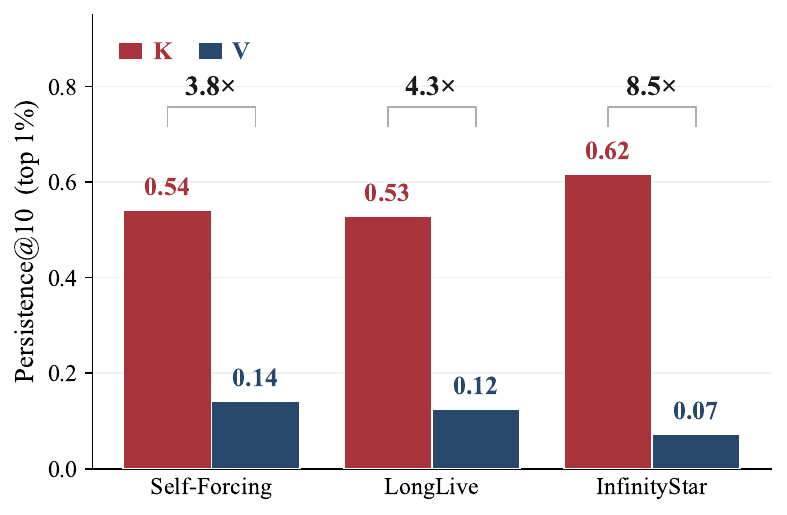}
  \caption{Outlier persistence across different prompts}
  \label{fig:persistence}
  \vspace{-\baselineskip}
\end{wrapfigure}

While Section~\ref{sec:structure} establishes \emph{where} massive outliers concentrate, this section investigates \emph{why} they emerge. We attribute these structural anomalies along the channel and block axes to specific operations within the K and V projection chains. Formally, we decompose the generation of $K_t$ and $V_t$ as follows:
\begin{equation}
K_t = \mathrm{RoPE}_t\!\Big(
\mathrm{RMSNorm}\!\big(x_t W_K^\top\big) \odot \gamma_K
\Big),
\qquad
V_t = x_t\, W_V^\top.
\label{eq:kv_chain}
\end{equation}

\paragraph{K outliers are architectural and V outliers are largely
input-driven.}
Whether an outlier is architectural or input-driven can be tested
empirically through \emph{cross-prompt persistence}: an architectural
outlier should persist across prompts, while an input-driven outlier
should drift. We sample $N$ prompts across models and define outliers as
the top-$p$ channels by magnitude. \textbf{Persistence@N} is the fraction
of a single prompt's outlier channels that remain outliers under \emph{all}
$N$ prompts (random $\approx 0$). At $N{=}10$, $p{=}1\%$, K
Persistence@10 reaches $\mathbf{0.53\text{--}0.62}$, a hard architectural
core shared across inputs; V sits at $\mathbf{0.07\text{--}0.14}$,
a $\mathbf{3.8\times\text{--}8.5\times}$ gap(Fig.~\ref{fig:persistence}).
V's near-vanishing persistence indicates input activations dominate V
outlier identity.

\begin{wrapfigure}{r}{0.6\linewidth}
  \vspace{-\intextsep}
  \centering
  \includegraphics[width=\linewidth]{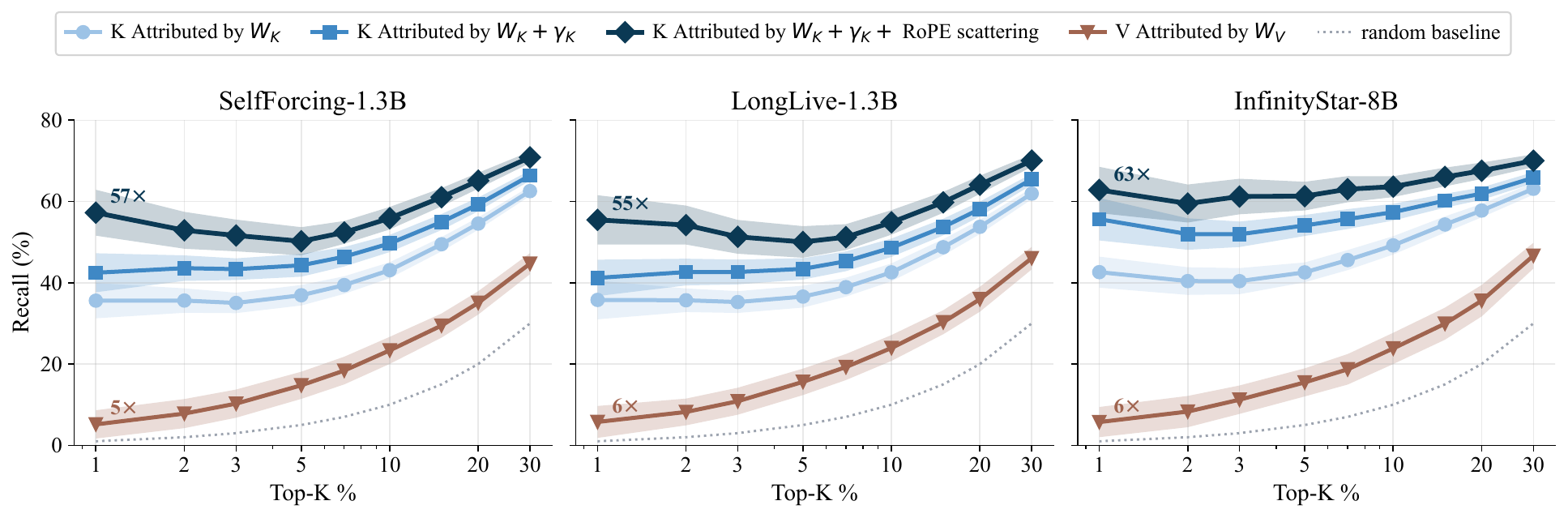}
  \caption{Recall of K outliers as projection-chain components are
  progressively added; $W_V$ shown as control.}
  \vspace{-6pt}
  \label{fig:recall}
\end{wrapfigure}

\paragraph{A Compounding Architectural Effect on K-Cache Channel Outliers.}
The architectural core of K identified above arises from three multiplicative sources along the projection chain (Eq.~\eqref{eq:kv_chain}): \textbf{(i)} heavy-tailed row norms of $W_K$ seed an initial outlier baseline; \textbf{(ii)} where RMSNorm-K is
present, the channel gain $\gamma_K$ amplifies this baseline; and
\textbf{(iii)} RoPE rotates each $(2i, 2i{+}1)$ pair, redistributing
energy between the two channels while preserving the pair's $\ell_2$ norm.

We quantify this compounding by progressively adding components and
measuring the recall of final K outliers (Fig.~\ref{fig:recall}). At
$p{=}1\%$, $W_K$ alone recovers $35\text{--}42\%$
($35\text{--}42\times$ random); adding $\gamma_K$ lifts recall further;
the full chain reaches $\mathbf{55\text{--}63\times}$ random. In contrast,
$W_V$ alone yields $5\text{--}6\times$ random recall on every model, an
order of magnitude weaker than the K chain.

\paragraph{Block-Axis Dynamics generated by RoPE Periodicity.}
A class of K block-axis outliers is generated by RoPE alone, independent of input dynamics. 
RoPE rotates each channel pair $(2i, 2i{+}1)$ at frequency $\omega_i$ by an angle $(b\,\omega_i)$ along the pair's RoPE axis, where $b$ is the block index. Within block $b$, every token shares the same rotation, and per-token magnitudes are
\begin{equation}
|K_{2i}(b, t)| = M_t\,|\cos(b\,\omega_i + \phi_t)|, \qquad
|K_{2i+1}(b, t)| = M_t\,|\sin(b\,\omega_i + \phi_t)|,
\label{eq:rope_modulation}
\end{equation}

\begin{wrapfigure}[13]{r}{0.6\textwidth}
  \vspace{-13pt}
  \centering
  \includegraphics[width=0.58\textwidth]{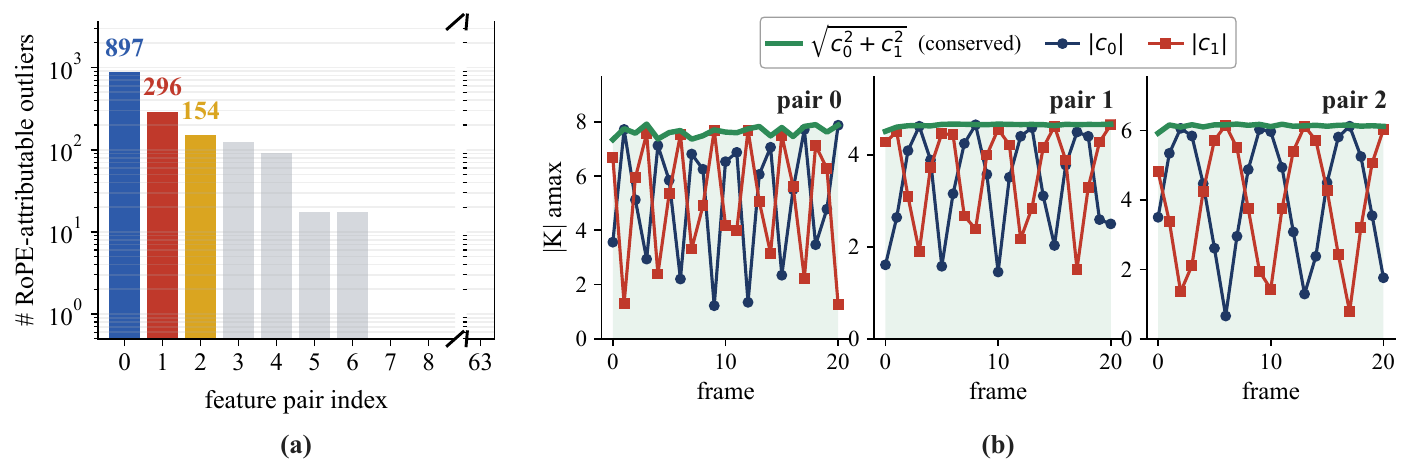}
  \caption{
  \textbf{(a)} On Self-Forcing, block-axis outliers concentrate on the lowest-index (fastest-rotating) pairs.
  \textbf{(b)} Within such a pair, $|c_0|$ and $|c_1|$ swap energy across blocks while pair $\ell_2$ stays approximately flat.
  }
  \label{fig:block_freq_link}
  \vspace{-14pt}
\end{wrapfigure}

with $M_t$ the pre-RoPE pair amplitude at token $t$ and $\phi_t$ its initial angle. Let $c_{2i}(b)$ denote the channel's per-token magnitudes within block $b$, with $\|c_{2i}(b)\|_\infty := \max_t |K_{2i}(b, t)|$. Two consequences follow and are visible in Fig.~\ref{fig:block_freq_link}.
\textit{(i) Pair envelope is approximately block-invariant.} RoPE preserves the per-token pair $\ell_2$ exactly, and in the massive-value regime this carries over to the pair of channel maxes (Appendix~\ref{app:bound}). The two channels move in anti-phase within each block.
\textit{(ii) Fast pairs dominate the block axis.} The total angle a pair traverses across the block sequence determines its swing: slow pairs barely rotate and stay flat, fast pairs complete multiple cycles and produce block bands.

\subsection{VAR-Q: Algorithm and System Support}
\label{sec:algorithm}

\begin{wrapfigure}{r}{0.5\columnwidth}
    \centering
    \vspace{-14pt}
    \includegraphics[width=\linewidth,trim={63pt 70pt 60pt 40pt}, clip]{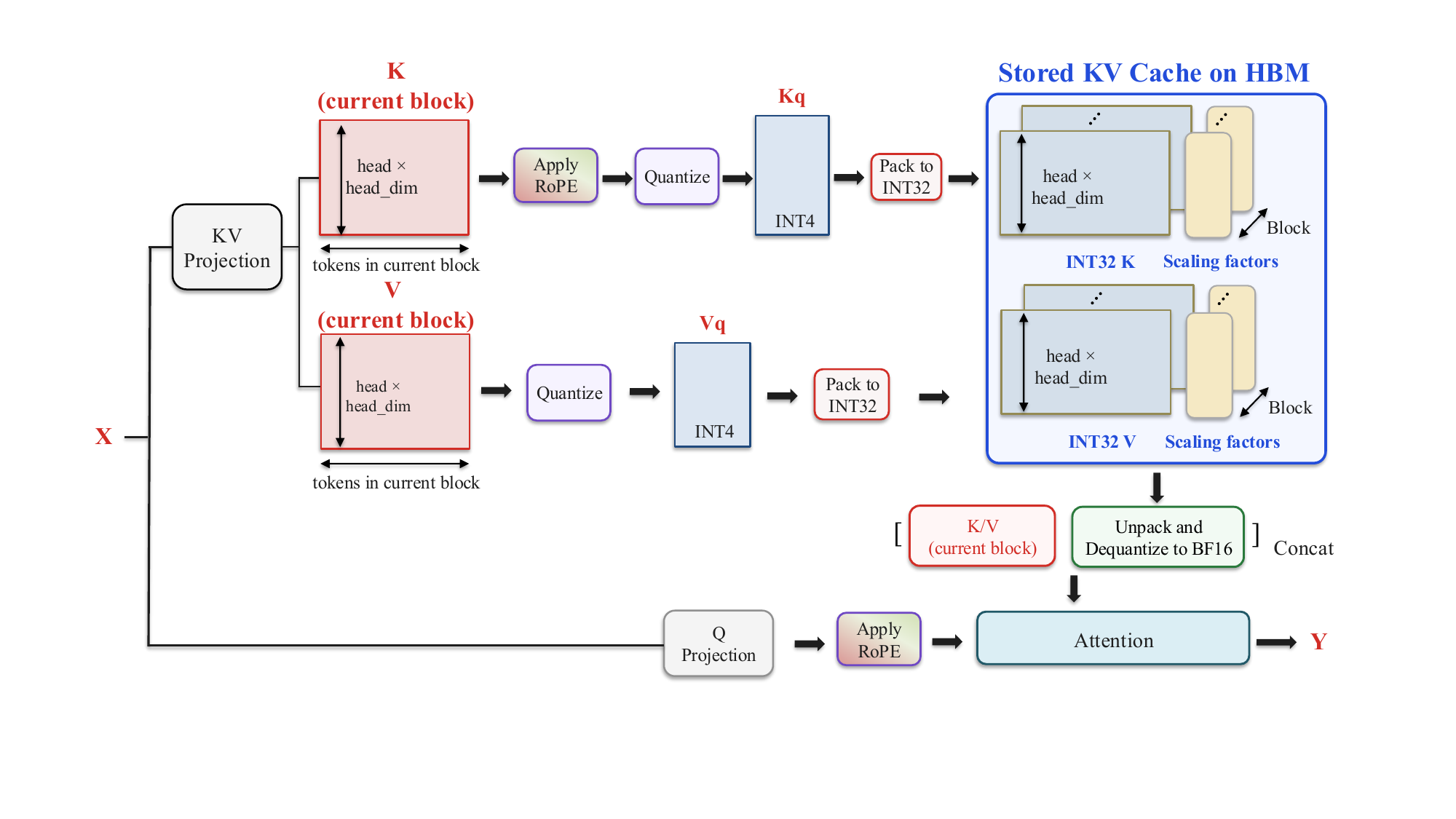}
    \caption{The proposed VAR-Q pipeline.}
    \label{fig:varq_init}
    \vspace{-10pt}
\end{wrapfigure}

\textbf{Algorithm.}
Section~\ref{sec:structure} shows that massive values in the KV cache concentrate along two axes: channels and blocks. VAR-Q aligns the quantization grouping with both. As shown in Fig.~\ref{fig:varq_init}, on top of the per-channel grouping, VAR-Q splits each group at block boundaries. The resulting quantization unit is indexed by $(i, j)$, where $i$ is the block index and $j$ is the channel; each unit carries a single scaling factor under uniform symmetric quantization.


The rule is online without requiring auxiliary detection or calibration and retraining. It is paradigm-agnostic, applying identically whether a block is a scale or a frame. And it is unified across keys and values, since both caches share the same two-axis structure.

We further expose a single hyperparameter $r \in (0, 1]$, the sub-block ratio. Let $L_{\max}$ denote the largest block size in the sequence. Each block of size $L$ is partitioned along the token axis into $\lceil L / (r L_{\max}) \rceil$ equal-sized sub-blocks; smaller $r$ yields finer groups. In next-scale AR earlier scales are geometrically smaller than $L_{\max}$ and remain unsplit at typical $r$, so $r$ effectively controls the granularity of the last scale; in next-frame AR every block has size $L_{\max}$, so $r$ partitions every frame uniformly. We default to $r=1$ throughout, recovering the per-block rule, and ablate $r$ in Section~\ref{sec:exp}.

\textbf{System Support.}
Quantization alone does not reduce the KV-cache footprint, because current CUDA toolchains lack native sub-8-bit storage types. We pair the algorithm with a hardware-friendly Triton implementation that packs sub-byte quantized values into byte-aligned containers on write and unpacks them during attention; details in Appendix~\ref{app:system_support_details}.

\section{Experiments}\label{sec:exp}

We evaluate \ourmethodandspace on nine models across five architectures, covering both paradigms and modalities: Infinity-2B/8B~\citep{han2025infinity} for next-scale image generation, InfinityStar-8B~\citep{infinitystar} at 720p and 480p for next-scale video, and Self-Forcing~\citep{huang2025selfforcing} on 5-second video clips and LongLive~\citep{yang2025longlive} on 30-second video clips for next-frame video generation. We additionally include VAR-d20/d24/d30~\citep{tian2024visual} on class-conditional ImageNet-1K. 
We compare against KIVI~\citep{liu2024kivi} and FlexGen~\citep{sheng2023flexgen}. KIVI uses per-channel grouping for K and per-token grouping for V; FlexGen uses $16{\times}16$ tile-wise grouping; VAR-Q uses per-channel and per-block grouping with $r=1$. All methods share identical bit-packing kernels, and each method's scale-factor overhead is included in the reported KV.
For class-conditional image generation we report FID, IS, Precision, and Recall, together with PSNR, SSIM, and LPIPS against the BF16 output under the same seeds, isolating quantization fidelity from sampling stochasticity. 
For text-to-image we report GenEval~\citep{ghosh2023geneval} and DPG~\citep{hu2024ella}; for text-to-video we report VBench~\citep{huang2024vbench} and CLIP Score on MovieGen prompts~\citep{polyak2024movie}, and additionally report VLM-as-judge~\citep{zheng2023judging} scores using Claude Opus 4.7 on long-video clips. 
The tuning-free methods require no calibration data, and INT8 packed kernels are implemented in Triton and evaluated on NVIDIA A100 80\,GB.

\input{Table/next_scale_image}
\subsection{Next-Scale Image Generation}
\label{sec:exp-next-scale-image}

Table~\ref{tab:next-scale-image} reports next-scale image generation on VAR (class-conditional) and Infinity (text-to-image). The qualitative examples are shown in Appendix~\ref{app:examples}.

On VAR-d20/d24/d30, \ourmethodandspace at INT4 matches BF16 in FID across all three depths and leads KIVI and FlexGen on PSNR, SSIM, and LPIPS computed against matched-seed BF16 outputs. FID is noisy at this scale; the consistent PSNR/SSIM/LPIPS advantage shows that \ourmethodandspace tracks the BF16 generation trajectory at the sample level rather than coincidentally landing on a similar distribution. The gap sharpens at INT3: \ourmethodandspace stays within 1.0 FID of BF16 (3.77/3.60/3.05) while KIVI degrades by 1.2 to 2.3 FID and FlexGen collapses (FID 15.3/14.9/9.5). 

On Infinity-2B and Infinity-8B, \ourmethodandspace preserves GenEval and DPG Overall within 0.3 points of BF16 at INT4. The 8B variant additionally holds at INT3 (GenEval 83.28, $\Delta=0.26$) while KIVI loses 0.79 points, and remains usable at INT2 (GenEval 74.38, DPG 83.83), confirming that larger backbones tolerate more aggressive KV-cache quantization.

\subsection{Next-Scale and Next-Frame Video Generation}
\label{sec:exp-video}
\input{Table/next_scale_frame_video}

We evaluate \ourmethodandspace on next-scale video with InfinityStar and on next-frame video with Self-Forcing and LongLive; all three backbones use RoPE. The qualitative examples are shown in Appendix~\ref{app:examples}.

\textbf{InfinityStar.}
At INT4 on both 720p and 480p, \ourmethodandspace matches BF16 in VBench Total Quality (81.66 and 81.32 vs.\ 82.33 and 81.30) and ranks first on Total at every precision and resolution. The gap widens at INT3 on the higher-resolution 720p setting: KIVI loses 6.87 points to 75.46, while \ourmethodandspace stays within 1.74 points of BF16 at 80.59. KV memory drops by 73 to 79\% across precisions.

\textbf{Self-Forcing and LongLive.}
At INT6, \ourmethodandspace is indistinguishable from BF16 on VBench Total (83.34 and 80.94 vs.\ 82.94 and 80.92). Opus4.7-VL ratings on LongLive corroborate this: 3.84 / 3.31 / 3.71 against BF16's 3.75 / 3.25 / 3.69. At INT4, \ourmethodandspace's gap to BF16 stays within 2 VBench points (1.95 and 1.32), against 2.61 and 1.97 for KIVI and 7.03 and 4.33 for FlexGen; cache savings reach 75\%.
Together with the VAR result in Section~\ref{sec:exp-next-scale-image}, these results establish that \ourmethodandspace's two-axis grouping transfers across both visual AR paradigms.


\subsection{Ablation Studies}

\input{Tables/AblationStudy_Group}

\textbf{Grouping rule.} Table~\ref{tab:grouping_ablation} ablates the two axes of \ourmethod{} at 4-bit on VAR-d30 (next-scale) and Self-Forcing INT4 (next-frame). On VAR-d30, removing the channel axis collapses generation (FID 12.07, IS 159.52); removing the block axis raises FID by 0.86 (1.97 to 2.83). The channel axis is the dominant factor, the block axis secondary. The trend persists on Self-Forcing: removing the channel axis drops Subject Consistency from 91.22 to 79.17 and Dynamic Degree from 74.00 to 27.00. \ourmethod{} covers both axes by construction and is therefore the safe paradigm-agnostic default. Among LLM-style alternatives, per-head per-token trails by 0.24 FID on VAR-d30 and 1.04 points on Subject; per-token alone collapses on both backbones with FID of 7.13 and Subject of 78.55.



\begin{wrapfigure}{r}{0.42\columnwidth}
    \centering
    \vspace{-15pt}
    \includegraphics[width=\linewidth,trim=0mm 6mm 0mm 6mm,clip]{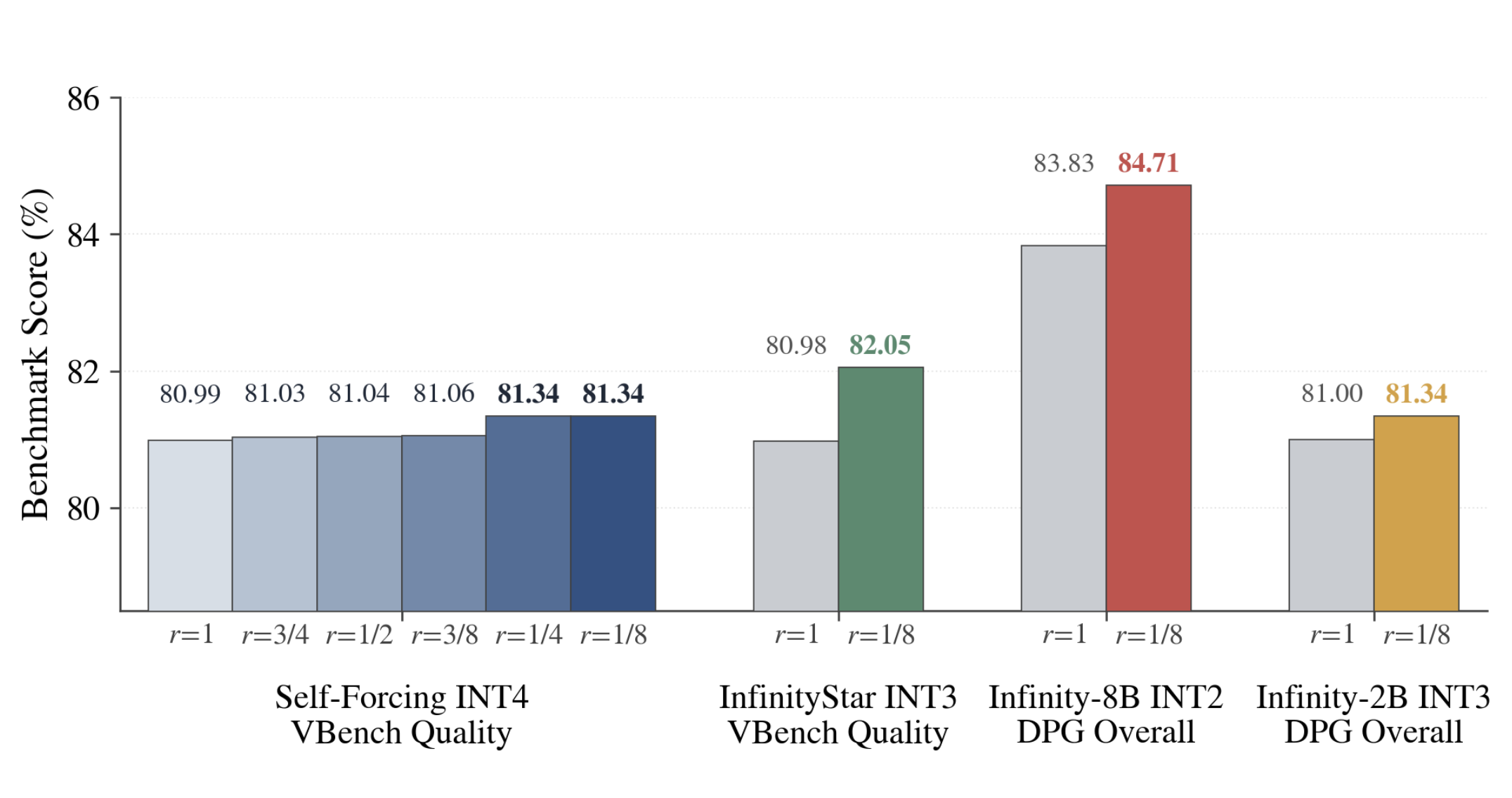}
    \caption{The effect of sub-block ratio $r$.}
    \label{fig:varq_sweep_r}
    \vspace{-13pt}
\end{wrapfigure}

\textbf{Effect of sub-block ratio $r$.}
Beyond the default per-block grouping ($r=1$), finer sub-blocks improve quality on every backbone we test (Figure~\ref{fig:varq_sweep_r}). On Self-Forcing INT4, VBench Quality rises from $80.99$ at $r=1$ to $81.34$ at $r=1/4$ and saturates at $r=1/8$. At $r=1/8$, gains range from $+0.34$ for Infinity-2B INT3 to $+1.07$ for InfinityStar INT3, with Infinity-8B INT2 at $+0.88$. Crucially, $r$ partitions within existing block boundaries, so this refinement adds no detection or calibration cost.

\input{Table/boundry_ablation}

\textbf{Effect of Boundary Alignment.}
We define two schemes with exactly matched groupings. \emph{Aligned} places group boundaries on block boundaries; \emph{misaligned} shifts every boundary by half a block. This minimal perturbation reduces the generation quality of Self-Forcing by 0.45 in Table~\ref{tab:boundary-ablation} and uniformly degrades all dimensions.

\section{Related Work}
Visual autoregression has consolidated around two block-wise paradigms: \emph{next-scale} prediction~\citep{tian2024visual, han2025infinity, infinitystar}, which decodes images coarse-to-fine across resolutions, and \emph{next-frame} prediction~\citep{huang2025selfforcing, yang2025longlive}, which decodes video frame-by-frame. Both emit a block of tokens per autoregressive step, so the KV cache scales with horizon and dominates memory in long-form synthesis.
KV-cache compression for language models is mature, spanning low-precision quantization~\citep{liu2024kivi, hooper2024kvquant, liu2024spinquant, su2025rotatekv, baek2025fireq, yao2025vecinfer}, attention-guided eviction~\citep{li2024snapkv, zhang2023h2o}, and hybrid quantization-sparsification~\citep{kang2024gear}. These methods inherit token-centric assumptions of uniform importance, linear cache growth, and stationary statistics across steps, all of which fail under the block-wise dynamics of visual AR, where magnitudes shift sharply at block boundaries and concentrate on a small set of channels. Extensions to diffusion transformers~\citep{ma2024deepcache} and multimodal LLMs~\citep{wan2024look} target adjacent regimes but inherit the same statistics.
The closest concurrent work addresses visual AR directly. KV eviction~\citep{li2025scalekv, xu2025amskv} and sparse-attention accelerators~\citep{xi2025sparsevideogen} reduce \emph{which} entries are retained rather than their precision, and thus compose orthogonally with \ourmethod. Concurrent quantization efforts~\citep{moonshift, xi2026quant} target next-scale image and next-frame video respectively, but neither attributes the cache structure to its architectural sources nor exploits the joint channel-block grouping; \ourmethod{} reuses block boundaries from the generation loop and applies one calibration-free rule across both paradigms.

\section{Conclusion}
\ourmethodandspace groups KV activations along channel and block axes, reusing boundaries supplied by the generation loop and requiring no calibration. Across nine visual AR models spanning both next-scale and next-frame paradigms, it cuts KV memory by up to 78.5\% with negligible quality degradation, enables 6$\times$ larger inference batches, and makes large visual autoregressive models more practical to deploy.

\bibliographystyle{plainnat}
\bibliography{main}

\clearpage

\appendix
\section{Asymmetric Roles of $\mathbf{K}$ and $\mathbf{V}$ Across Modalities}
\label{app:kv_separate}
\subsection{$\mathbf{K}$/$\mathbf{V}$ asymmetry between image and video AR}

The joint result above masks a per-tensor split: \textbf{video models are more sensitive to $\mathbf{K}$ outliers, while image models are more sensitive to $\mathbf{V}$ outliers}.
Mechanistically, $\mathbf{K}$ determines which past tokens are attended to and $\mathbf{V}$ determines what content is retrieved. The split therefore suggests that video generation places a heavier demand on precise temporal addressing into the cache, whereas image generation tolerates coarser addressing but depends more on the fidelity of the content retrieved at each cached position. The same per-tensor ablation on LLMs reproduces an LLM-side $\mathbf{K}$-dominance consistent with prior reports~\citep{liu2024kivi, tao2025asymkv}, placing video AR closer to the LLM regime and image AR farther from it.

This appendix provides per-tensor evidence for the $\mathbf{K}$/$\mathbf{V}$ asymmetry discussed in the main paper, separating $\mathbf{K}$-only and $\mathbf{V}$-only ablations across image and video AR, and extending the comparison to LLaMA-3-8B-Instruct. We follow the channel-wise ablation protocol of the main paper but mask $\mathbf{K}$ and $\mathbf{V}$ independently. Channels are ranked by per-token absolute maximum aggregated over a calibration set; the top, bottom, or random $p\%$ are zeroed at inference, with $p \in \{0.1, 0.5, 3, 5\}$. Four visual AR backbones (Infinity-2B, Infinity-8B, LongLive-30s, InfinityStar-720p) are evaluated with the perceptual and VBench metrics used in the main paper, and LLaMA-3-8B-Instruct is evaluated on WikiText-2 perplexity (4096-token chunks), MMLU (200 sampled questions), and GSM8K (50 sampled questions, chat-template greedy generation). Random and inlier ablation arms remain within metric-level noise of the unmasked baseline across all backbones and ratios, confirming that the observed degradation is selective to high-magnitude channels rather than to channel count. One measurement caveat applies to video: the VBench dimensions \emph{dynamic\_degree} and \emph{temporal\_flickering} register increases under $\mathbf{K}$-outlier ablation that do not reflect genuine motion, since the per-frame chroma drift induced by $\mathbf{K}$-side collapse is misclassified as motion by the underlying classifiers; we therefore rely on \emph{imaging\_quality} (MUSIQ) and \emph{overall\_consistency} (ViCLIP) as the most reliable VBench dimensions for this perturbation regime.

\begin{figure}[t]
    \centering
    \includegraphics[width=\columnwidth]{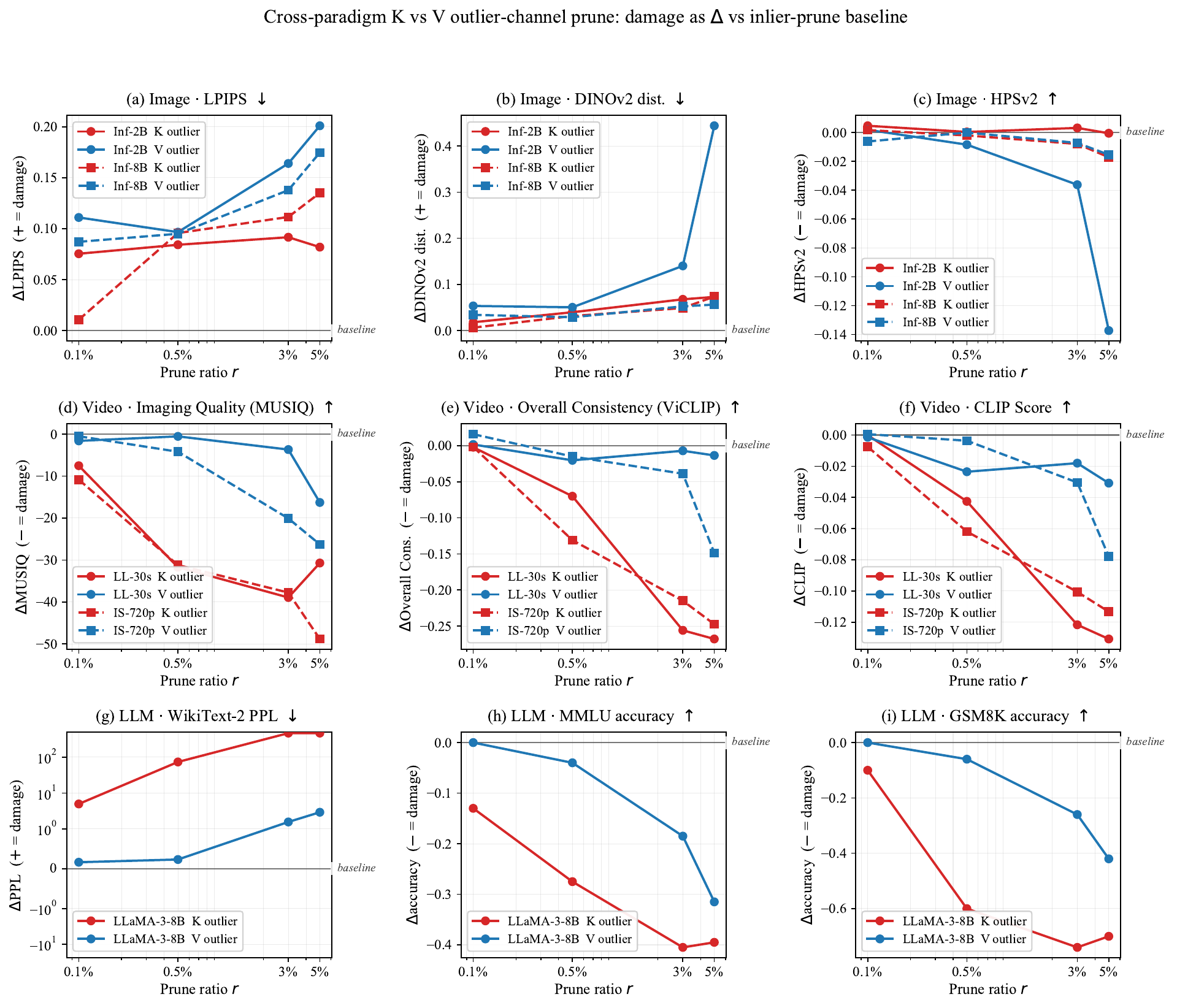}
    \caption{$\mathbf{K}$-only (red) versus $\mathbf{V}$-only (blue) outlier ablation. \textbf{Top row}: image AR backbones on perceptual and aesthetic metrics. \textbf{Middle row}: video AR backbones on perceptual and VBench-MUSIQ metrics. \textbf{Bottom row}: LLaMA-3-8B-Instruct on language benchmarks (PPL on log scale). All metrics are oriented so that higher values indicate worse generation. The dominant color flips between image and video AR.}
    \label{fig:cross_modality_main}
\end{figure}

\begin{figure}[t]
    \centering
    \includegraphics[width=\columnwidth]{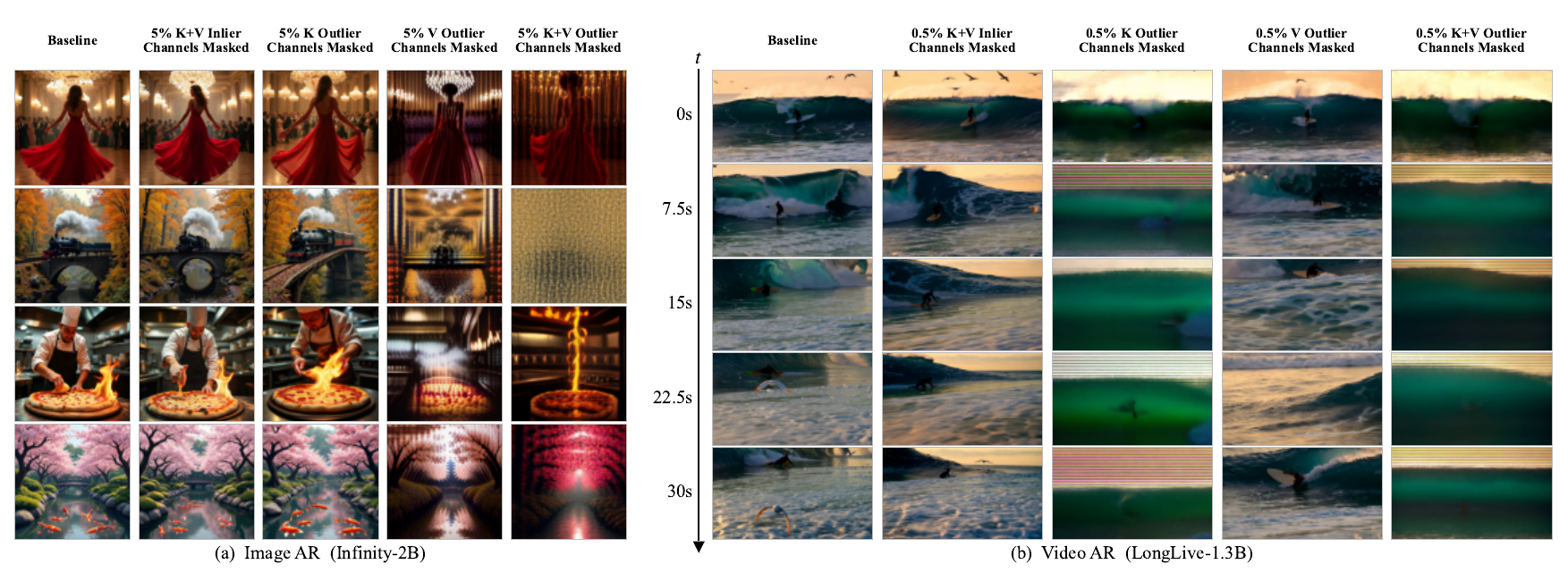}
    \caption{Generation under $\mathbf{K}$-only, $\mathbf{V}$-only, and joint $\mathbf{K}{+}\mathbf{V}$ outlier masking. \textbf{(a)} Infinity-2B at $p=5\%$: $\mathbf{V}$-only masking collapses outputs into texture or symmetric patterns, while $\mathbf{K}$-only masking leaves outputs visually close to the baseline. \textbf{(b)} LongLive-1.3B at $p=0.5\%$: $\mathbf{K}$-only masking collapses frames into solid color and chroma stripes, while $\mathbf{V}$-only masking preserves scene layout with degraded fine detail. Inlier masking matches the baseline in both cases.}
    \label{fig:qualitative_KV}
\end{figure}

\subsection{Extension to LLaMA-3-8B}
The top two rows of Figure~\ref{fig:cross_modality_main} show that image and video AR sit on opposite sides of the $\mathbf{K}$/$\mathbf{V}$ axis. Image AR is $\mathbf{V}$-dominant: on Infinity-2B at $p=5\%$, $\mathbf{V}$-only ablation drives $\Delta$LPIPS to $0.20$ against $0.08$ for $\mathbf{K}$-only, and the LAION aesthetic score drops by $1.3$ under $\mathbf{V}$-only against near-zero under $\mathbf{K}$-only. Video AR reverses the direction: on LongLive-30s at $p=5\%$, $\mathbf{K}$-only ablation drives $\Delta$LPIPS$_\text{mid}$ to $0.40$ against $0.14$ for $\mathbf{V}$-only, and the same direction holds on InfinityStar-720p. The image-side $\mathbf{V}$-dominance is consistent across both Infinity backbones and the video-side $\mathbf{K}$-dominance across both video backbones, indicating that the asymmetry tracks the modality rather than per-model idiosyncrasy. Figure~\ref{fig:qualitative_KV} renders the same contrast at the sample level: on Infinity-2B, $\mathbf{V}$-only masking eliminates recognizable scene content while $\mathbf{K}$-only masking is nearly invisible; on LongLive-1.3B, $\mathbf{K}$-only masking collapses frames into uniform color while $\mathbf{V}$-only masking preserves scene layout with degraded texture. The joint $\mathbf{K}{+}\mathbf{V}$ mask in both cases tracks the more damaging single-tensor mask, consistent with the dominant tensor carrying the load-bearing signal.

\begin{figure}[t]
    \centering
    \includegraphics[width=0.8\columnwidth]{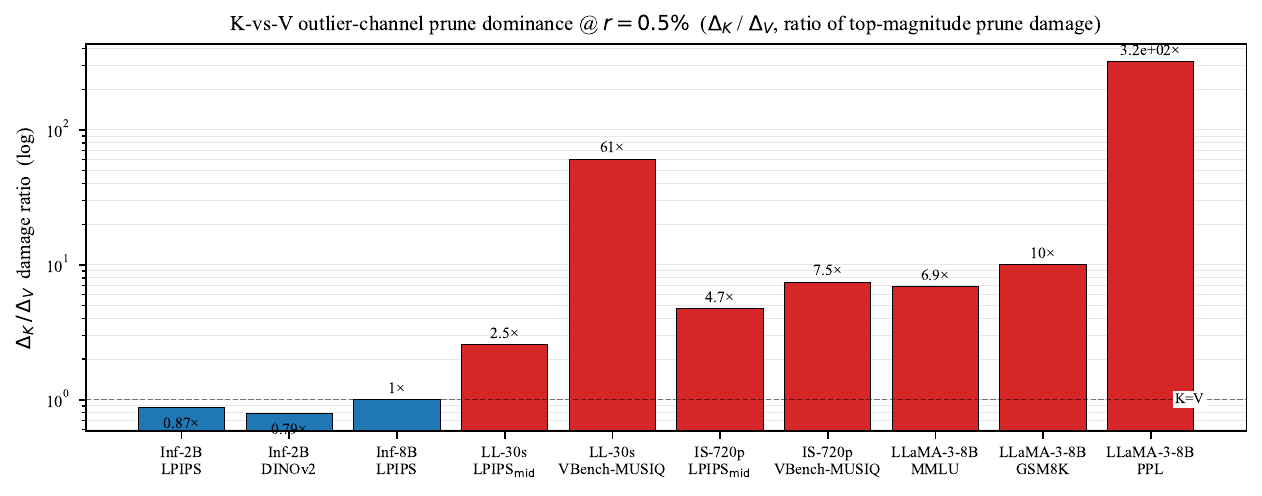}
    \caption{$\Delta_\mathbf{K}/\Delta_\mathbf{V}$ damage ratio at $p=0.5\%$ (log scale). Bars below the dashed $\mathbf{K}{=}\mathbf{V}$ line are $\mathbf{V}$-dominant; bars above are $\mathbf{K}$-dominant. Image AR sits at or below $1\times$, video AR ranges from $2.5\times$ to $61\times$, and LLaMA-3-8B reaches $320\times$ on perplexity.}
    \label{fig:cross_modality_dominance}
\end{figure}

The bottom row of Figure~\ref{fig:cross_modality_main} extends the same protocol to LLaMA-3-8B-Instruct. The model behaves like video AR in direction but more sharply in magnitude: at $p=0.5\%$, $\mathbf{K}$-only ablation increases WikiText-2 perplexity from $8.4$ to $82.4$ ($\sim\!10\times$), while $\mathbf{V}$-only ablation leaves it at $11.5$ ($1.4\times$); MMLU and GSM8K accuracy degrade in the same direction. Figure~\ref{fig:cross_modality_dominance} summarizes the damage ratio $\Delta_\mathbf{K}/\Delta_\mathbf{V}$ at $p=0.5\%$: image AR sits at or below $1\times$, video AR ranges from $2.5\times$ to $61\times$, and LLaMA-3-8B reaches $320\times$ on perplexity. The LLM observation aligns with prior reports of $\mathbf{K}$-side outlier dominance in LLM KV caches~\citep{liu2024kivi, tao2025asymkv}, placing video AR closer to the LLM regime while leaving image AR as the singular setting in which $\mathbf{V}$ outliers carry the bulk of the load.

\section{Persistency Across Different Prompts}
\begin{figure*}
  \vspace{-\intextsep}
  \centering
  \includegraphics[width=0.7\textwidth]{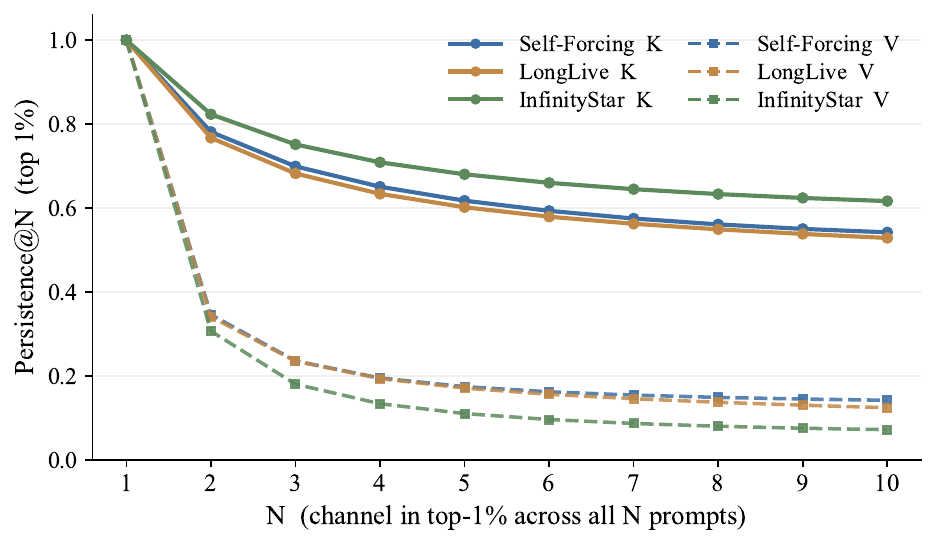}
  \caption{Outlier persistence across different amounts of prompts}
  \label{fig:persistence_curve}
  \vspace{-\baselineskip}
\end{figure*}

As shown in Figure~\ref{fig:persistence_curve}, the shapes of the K and V curves differ qualitatively, not just in magnitude. V (dashed) collapses on the first added prompt: even a single second prompt removes about two-thirds of the outlier channels (Persistence@2 $=0.31$ to $0.35$ across the three backbones), and the curve continues to decay toward $0.07$ to $0.14$. K (solid) instead bends gently, retaining $0.77$ to $0.82$ at $N\!=\!2$ and asymptoting at $0.53$ to $0.62$. The K-V gap is already $2.3\times$ at $N\!=\!2$ and widens monotonically with $N$, so the $N\!=\!10$ cross-section reported as Persistence@10 is the rightmost point of two
curves whose decay rates differ throughout. The plateau shape on
K is the signature of an architectural core: most outlier channels
survive every additional prompt, consistent with their being
selected by the static $W_K$, $\gamma_K$, and RoPE-pair chain
(Section~\ref{sec:interpretability}). The crash on V is the signature of
input-driven outliers: each new prompt is a fresh sample of token amplitudes, and the V channel identity tracks those samples.

\section{Approximate $\ell_2$ Invariance of Channel Max under RoPE}
\label{app:bound}

Within one block of $T$ tokens, let pair $(c_t, d_t)$ at token $t$ have pre-RoPE amplitude $M_t = \sqrt{c_t^2 + d_t^2}$, with order statistics $M_{(1)} \geq M_{(2)} \geq \cdots \geq M_{(T)}$. After RoPE rotation by a shared angle $\theta$, write $a_t = M_t\cos(\theta + \phi_t)$ and $b_t = M_t\sin(\theta + \phi_t)$, where $\phi_t$ is the per-token pre-RoPE angle. The pair of channel maxes satisfies
\begin{equation}
M_{(1)}^2 \;\leq\; \|a\|_\infty^2 + \|b\|_\infty^2 \;\leq\; M_{(1)}^2 + M_{(2)}^2.
\label{eq:bound}
\end{equation}

\textit{Lower bound.} Let $t^\star = \arg\max_t M_t$. Per-token RoPE invariance gives $a_{t^\star}^2 + b_{t^\star}^2 = M_{(1)}^2$. Since $\|a\|_\infty \geq |a_{t^\star}|$ and $\|b\|_\infty \geq |b_{t^\star}|$, the sum is at least $M_{(1)}^2$.

\textit{Upper bound.} Let $t_a = \arg\max_t |a_t|$ and $t_b = \arg\max_t |b_t|$. Per-token, $|a_t|, |b_t| \leq M_t$, so $\|a\|_\infty^2 \leq M_{t_a}^2$ and $\|b\|_\infty^2 \leq M_{t_b}^2$. If $t_a = t_b$, the sum equals $M_{t_a}^2 \leq M_{(1)}^2$. If $t_a \neq t_b$, at least one of them is not $t^\star$, so $\min(M_{t_a}, M_{t_b}) \leq M_{(2)}$ and the sum is at most $M_{(1)}^2 + M_{(2)}^2$.

\textit{Massive-value regime.} When pre-RoPE amplitudes are heavy-tailed within the block, $M_{(2)}^2 \ll M_{(1)}^2$ and the bounds collapse, leaving channel-max pair $\ell_2$ approximately invariant under RoPE rotation. This is the mathematical basis for the envelope-flat behavior in Fig.~\ref{fig:block_freq_link}(b).

\section{Prompt Template for VLM-based Video Evaluation}

\paragraph{Evaluation Prompt.}
For each video, we partition the timeline into 12 equal segments and sample one frame at the midpoint of each segment, i.e., at relative timestamps $(i+0.5)/12$ for $i=0,\ldots,11$. This produces samples spanning 4.17\% to 95.83\% of the video while excluding boundary regions, where transition artifacts are more common. Frames are extracted with \texttt{ffmpeg}, resized to 360 pixels in height with aspect ratio preserved, and stored as JPEG images using \texttt{-q:v 3}. To ensure a fair comparison, we keep the 12 sampling locations fixed across all quantization settings for the same prompt, so the VLM judge is exposed to identical temporal evidence. This corresponds to an average spacing of approximately 0.42 seconds for 5-second Self-Forcing videos and 2.5 seconds for 30-second LongLive videos.
We provide the exact prompt template used to query the VLM judge for rating text alignment, dynamics, and visual quality.

\begin{Verbatim}[breaklines=true,breakanywhere=true,fontsize=\small]
You are a video quality evaluation expert. Below are 12 frames uniformly sampled
from a generated video, plus the text prompt that was used to produce it. Rate
the video on three dimensions (1-5 integer each):

1. Text Alignment — consistency between video content and the text description.
   - 1: Completely Irrelevant. 2: Mostly Mismatched. 3: Partially Matched.
     4: Largely Consistent. 5: Perfectly Aligned.
2. Dynamics — dynamism/fluidity of the scene (camera, objects, transitions).
   - 1: Static/Disjointed. 2: Mostly Static. 3: Moderately Dynamic.
     4: Largely Dynamic. 5: Highly Dynamic.
3. Visual Quality — technical execution (clarity, color, composition, artifacts).
   - 1: Very Poor. 2: Poor. 3: Fair. 4: Good. 5: Excellent.

Output STRICTLY in this format (nothing else, no markdown fences):
Text Alignment: <1-5>
Dynamics: <1-5>
Visual Quality: <1-5>

Text Prompt: """{TEXT_PROMPT}"""
Frames (in temporal order): @f00.jpg @f01.jpg ... @f11.jpg
\end{Verbatim}

\section{VAR-Q System Supports}\label{app:system_support_details}
\input{Algorithms/pack}


\clearpage
\section{Pre-RoPE vs.\ post-RoPE quantization.} 
\input{Table/rope_and_bound}

\begin{figure}
    \centering
    \includegraphics[width=\columnwidth,
        trim={0 250pt 0 49pt},
        clip]{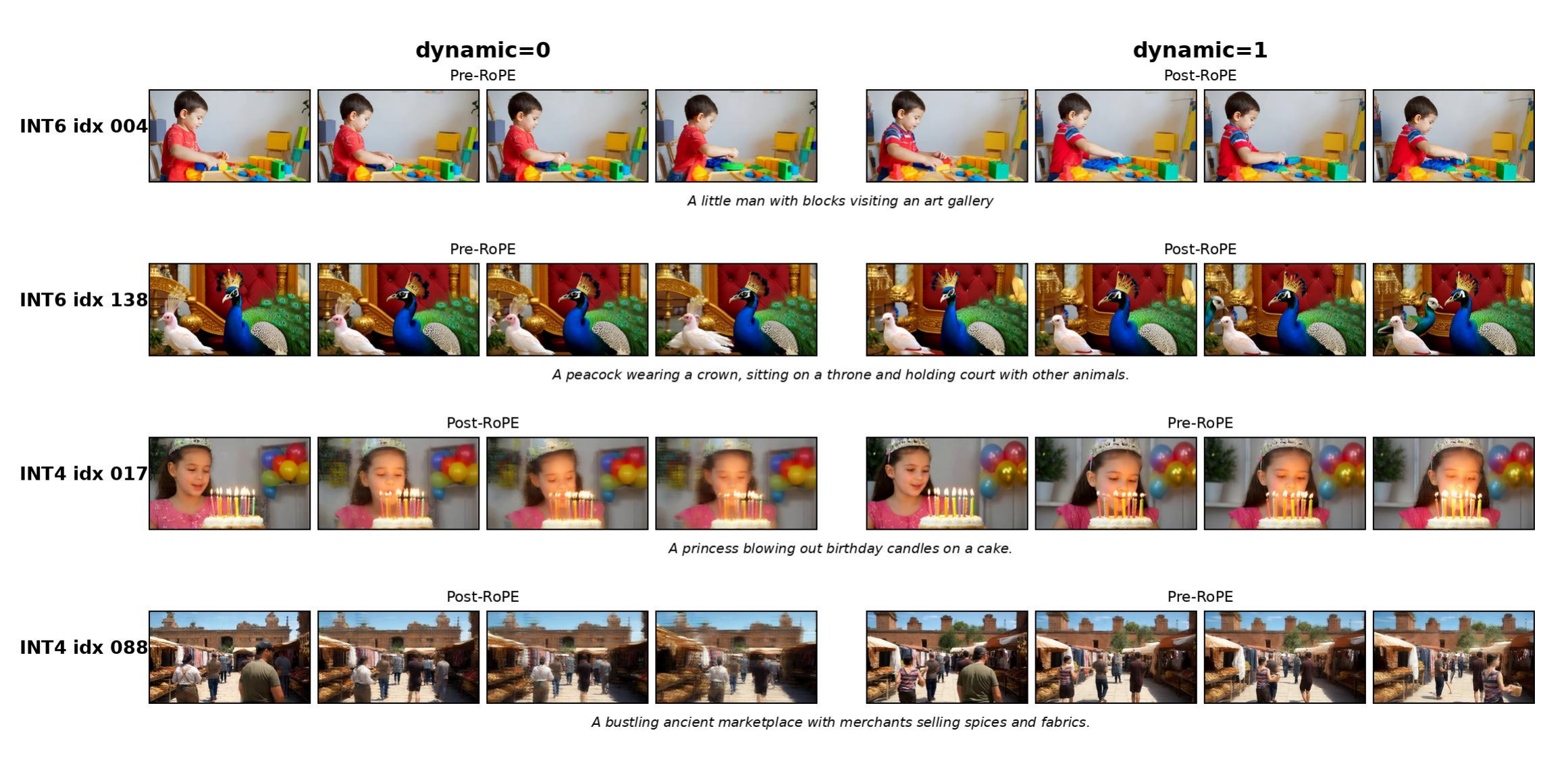}
    \caption{
    Pre-RoPE vs.\ post-RoPE quantization on Self-Forcing at INT6.
    Since RoPE preserves the pair-level envelope but redistributes magnitude across paired channels, quantizing after RoPE better matches the final attention-ready representation.
    Post-RoPE quantization better preserves dynamic behavior and overall visual quality in the shown examples.
    }
    \label{fig:prerope_dynamic}
\end{figure}

We ablate the quantization point of \ourmethod{} on Self-Forcing at INT6.
As shown in Table~\ref{tab:pre_rope}, post-RoPE quantization consistently outperforms pre-RoPE on the metrics most related to temporal behavior and overall generation quality: Subject Consistency, Background Consistency, Dynamic Degree, Aesthetic Quality, and Total Quality.
The gain in Dynamic Degree ($76.50$ vs.\ $74.50$) indicates that post-RoPE quantization better preserves motion cues instead of over-smoothing the generated sequence.

Meanwhile, the improvements in Subject and Background Consistency suggest that quantizing the attention-ready representation also stabilizes the appearance of both foreground objects and scene layout across frames.
Post-RoPE further improves Aesthetic Quality and Total Quality ($83.34$ vs.\ $82.93$), showing that the benefit is not limited to motion but also carries over to perceptual quality.

This supports quantizing the final post-RoPE cache, which is directly consumed by attention, rather than the pre-RoPE cache.
We therefore use post-RoPE quantization as the default.

\section{Qualitative Examples}
\label{app:examples}
This appendix provides additional qualitative comparisons across next-scale image generation, next-scale video generation, and next-frame video generation. 
For VAR, each figure arranges different quantization configurations by row and ImageNet classes by column, with the class ID shown above each column; INT8 examples include only BF16 and VAR-Q. 
For Infinity, each row corresponds to one text prompt and each column to one configuration, so images in the same row are generated from the same prompt. 
For video models, each page shows one prompt, with configurations arranged by row and temporally sampled frames arranged by column.

\paragraph{Qualitative observations.}
Across these examples, the visual trends are consistent with the quantitative results in the main paper. 
At moderate precision, VAR-Q largely preserves the appearance, layout, and temporal evolution of the BF16 outputs, while competing KV-cache quantization baselines more often introduce visible softening, local texture loss, or later-frame drift. 
We therefore keep individual figure captions concise and summarize the recurring visual patterns below, rather than describing every sample independently.

\paragraph{Next-scale image generation.}
For VAR and Infinity, the main visible difference appears in local fidelity rather than coarse semantics. 
Most methods preserve the high-level object category or prompt content, but FlexGen and KIVI more frequently lose fine textures, weaken small structures, or alter local details under low-bit KV caches. 
VAR-Q remains visually closer to BF16 across the shown samples, especially at INT4 and INT3, where it better preserves object boundaries, repeated patterns, and prompt-specific visual attributes.

\begin{center}
    \includegraphics[width=\textwidth,height=\textheight,keepaspectratio,page=1]{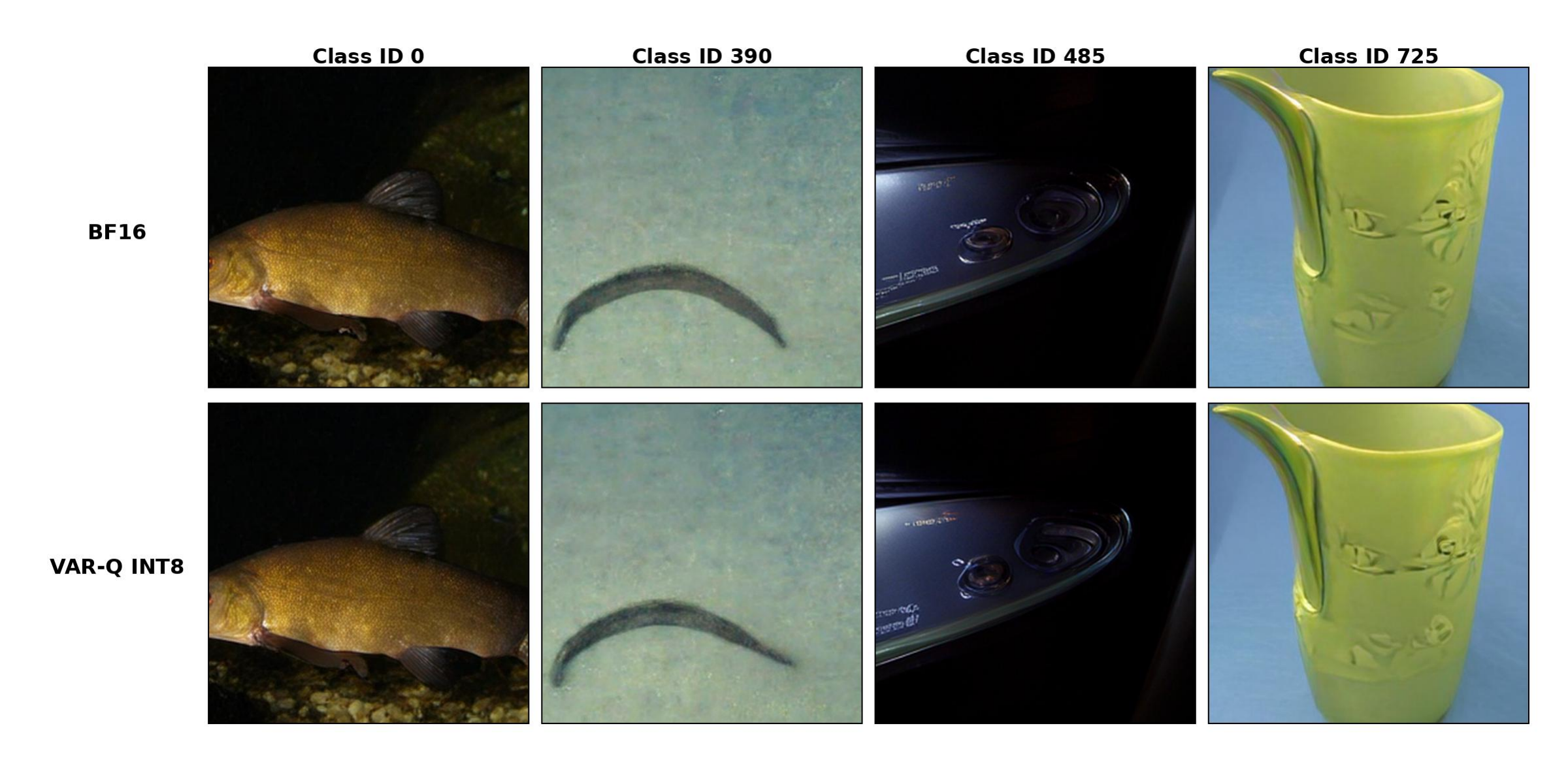}

    \refstepcounter{figure}
    \label{fig:var-d20-int8}
    \vspace{2mm}
    \small
    \textbf{Figure~\thefigure:} Qualitative comparison on VAR d20 with INT8 KV cache.
\end{center}
\begin{center}
    \includegraphics[width=\textwidth,height=\textheight,keepaspectratio,page=1]{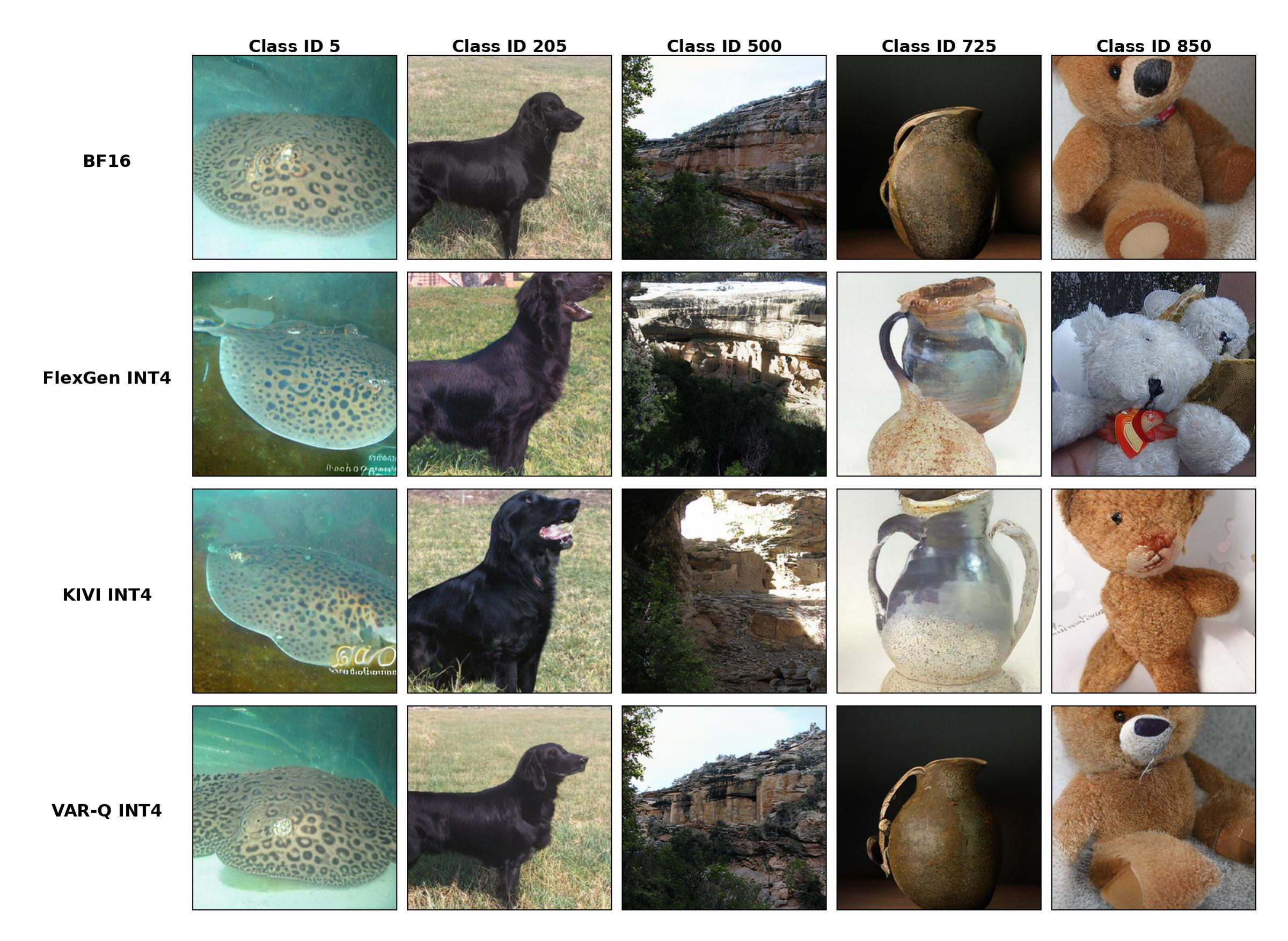}

    \refstepcounter{figure}
    \label{fig:var-d20-int4}
    \vspace{2mm}
    \small
    \textbf{Figure~\thefigure:} Qualitative comparison on VAR d20 with INT4 KV cache.
\end{center}

\begin{figure*}[!t]
    \centering
    \includegraphics[width=\textwidth,page=1]{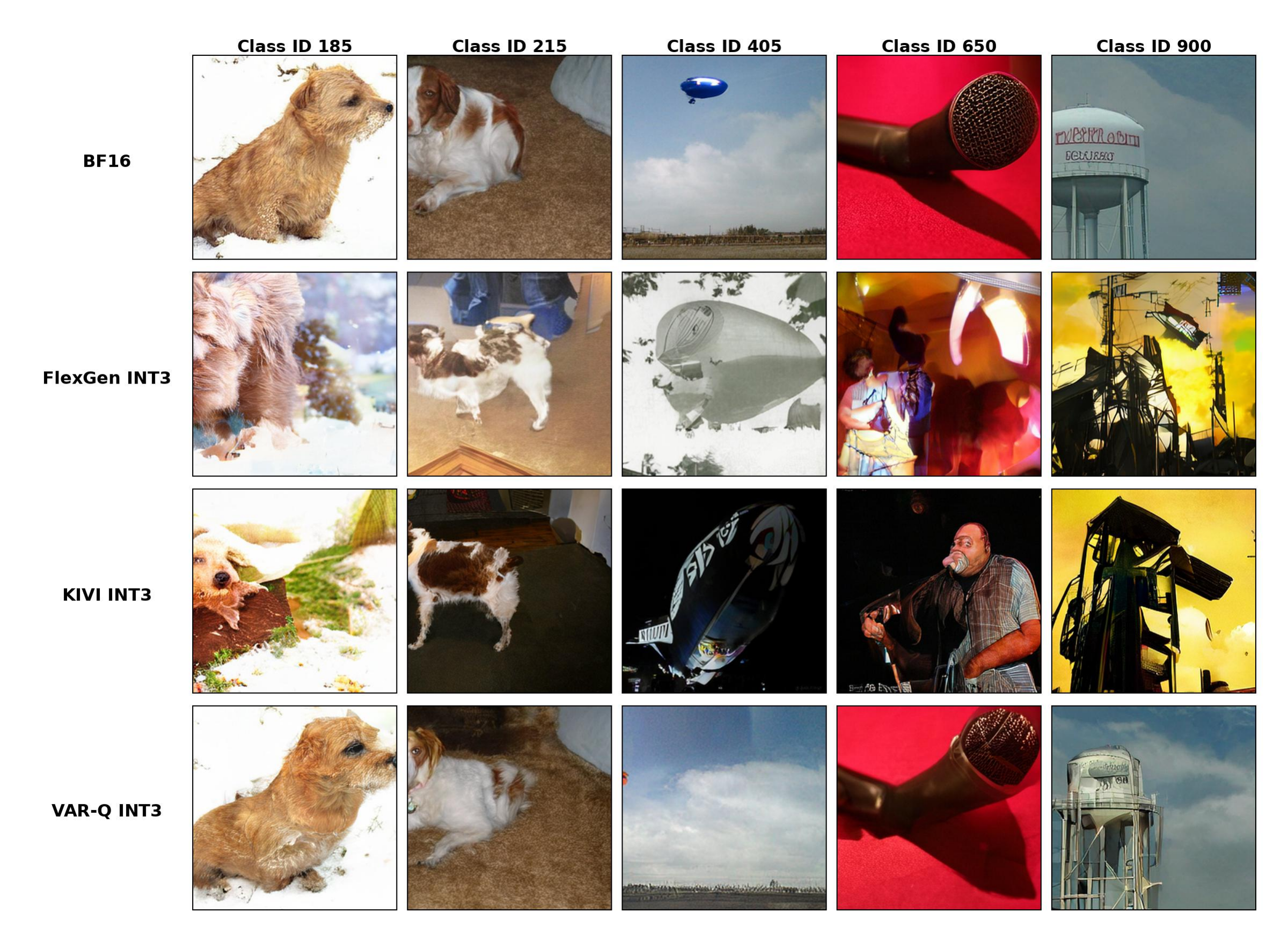}
    \caption{Qualitative comparison on VAR d20 with INT3 KV cache.}
    \label{fig:var-d20-int3}
\end{figure*}

\begin{figure*}[!t]
    \centering
    \includegraphics[width=\textwidth,page=1]{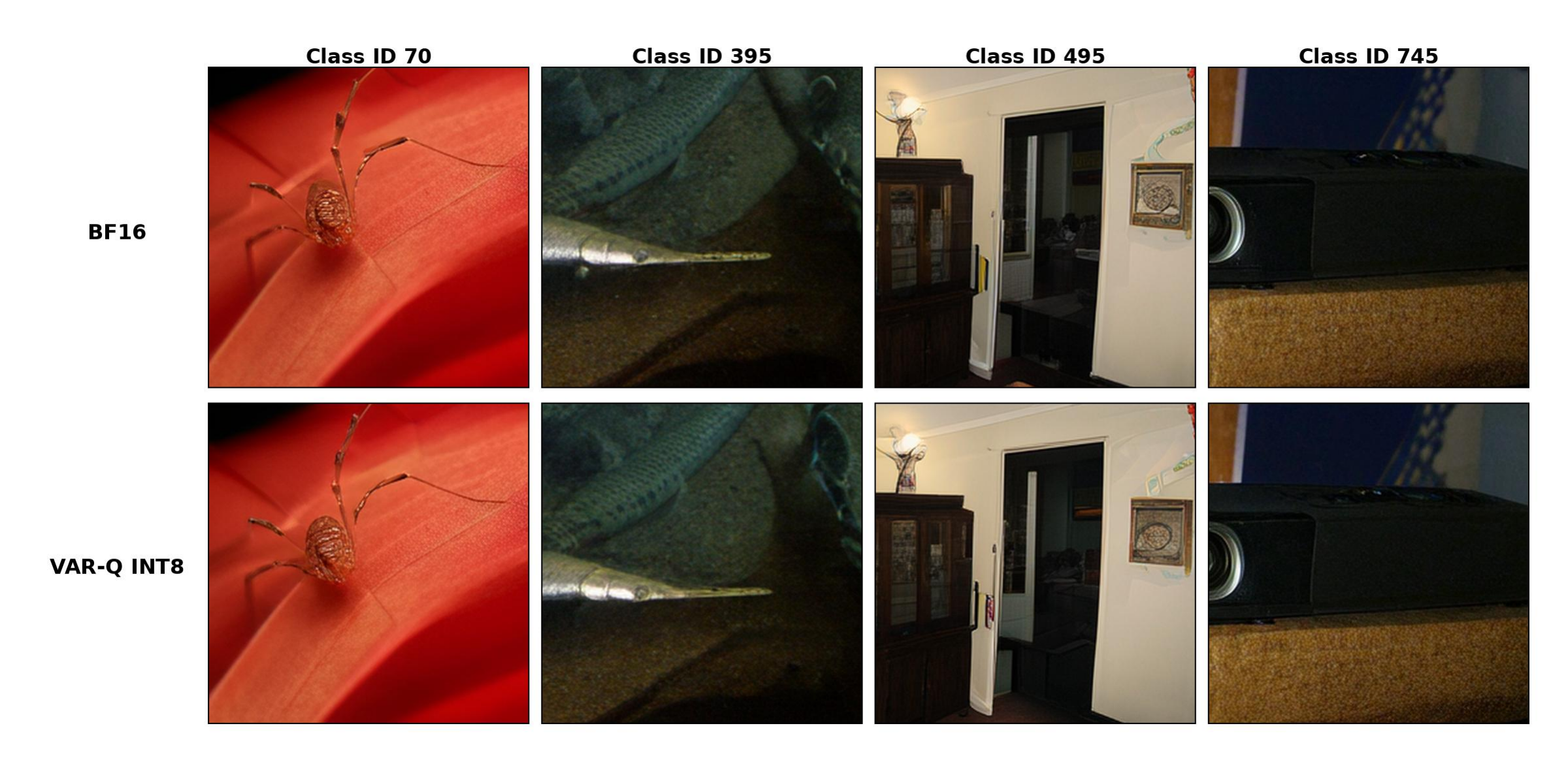}
    \caption{Qualitative comparison on VAR d24 with INT8 KV cache.}
    \label{fig:var-d24-int8}
\end{figure*}

\begin{figure*}[!t]
    \centering
    \includegraphics[width=\textwidth,page=1]{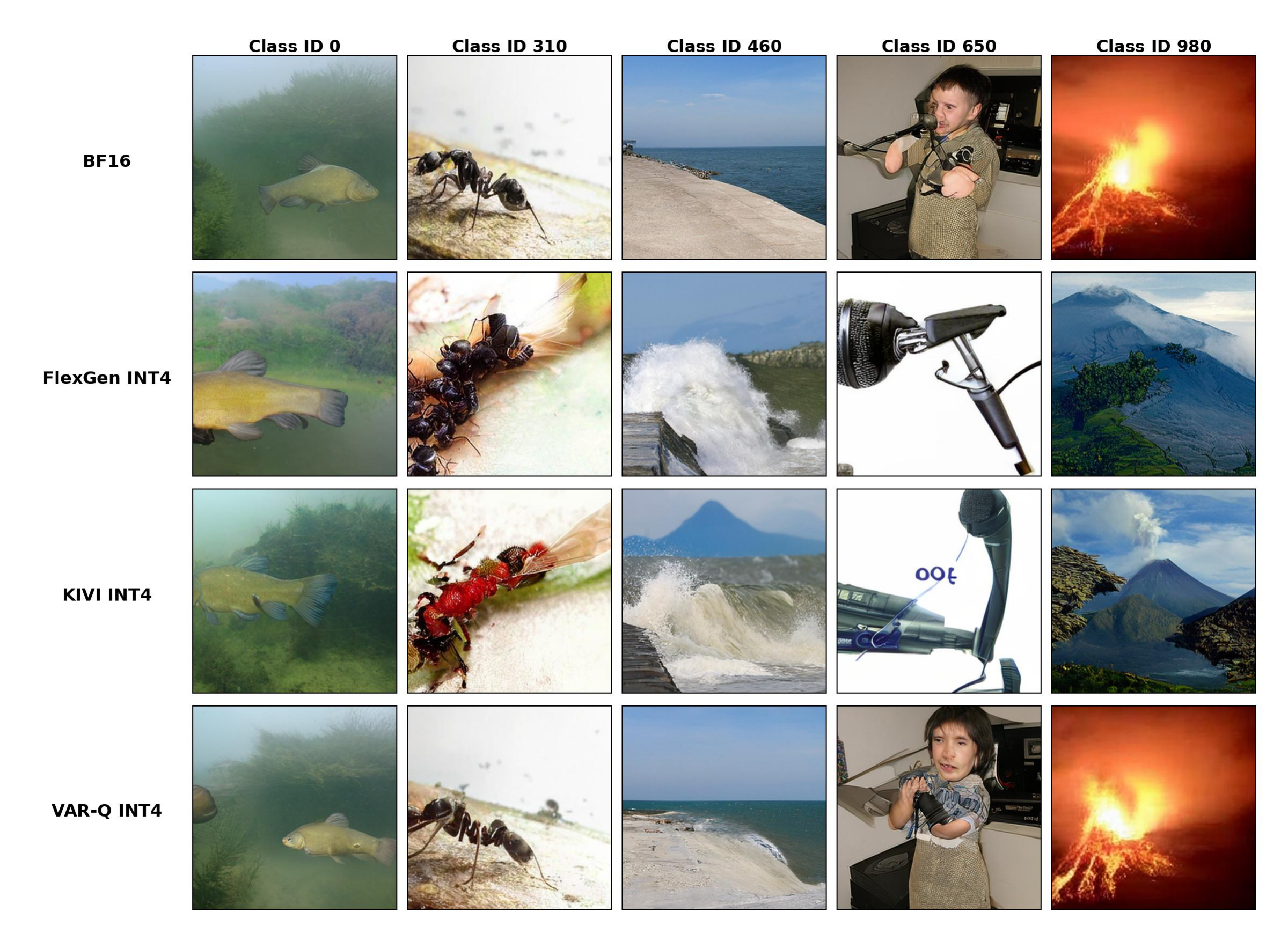}
    \caption{Qualitative comparison on VAR d24 with INT4 KV cache.}
    \label{fig:var-d24-int4}
\end{figure*}

\begin{figure*}[!t]
    \centering
    \includegraphics[width=\textwidth,page=1]{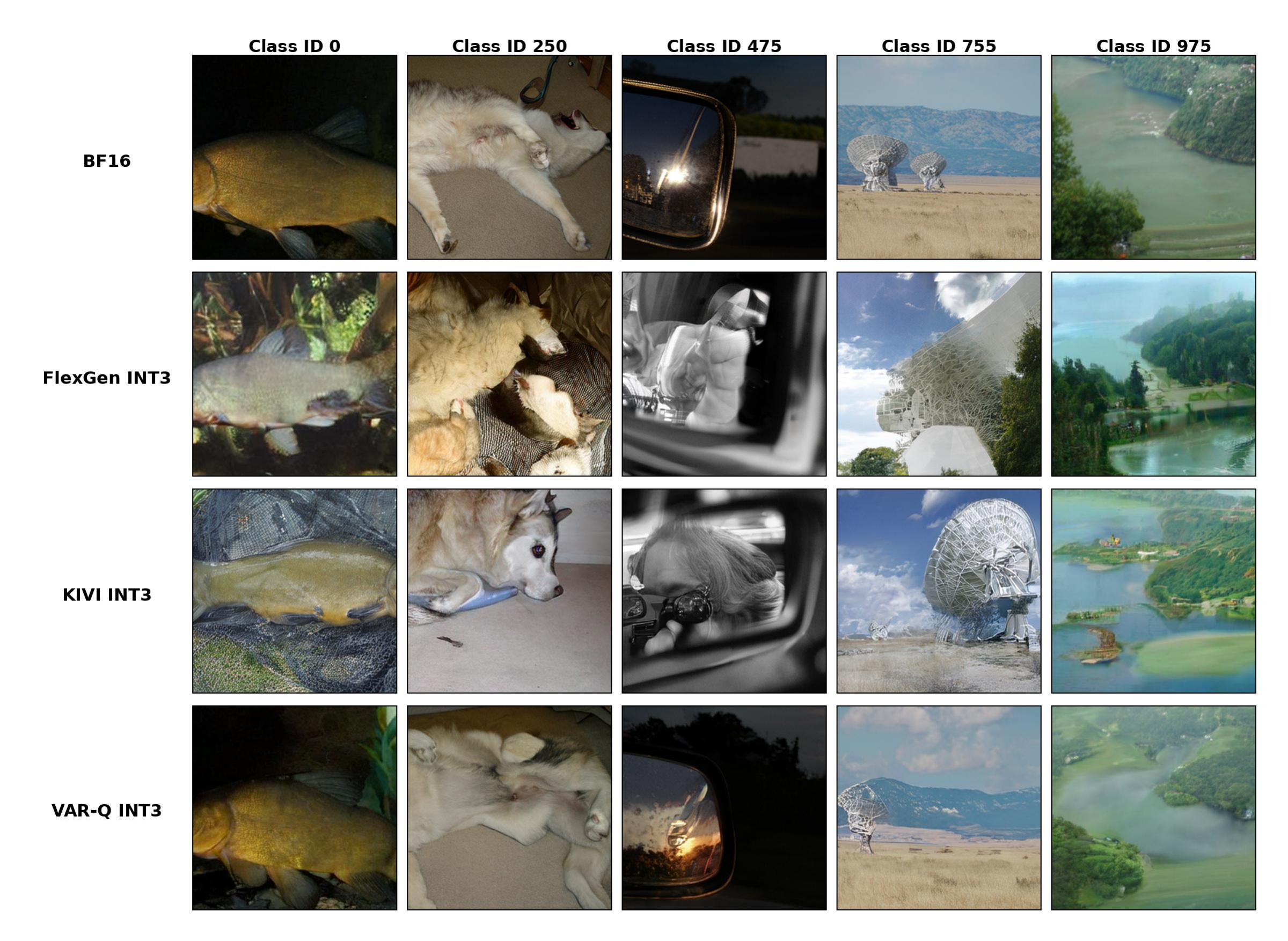}
    \caption{Qualitative comparison on VAR d24 with INT3 KV cache.}
    \label{fig:var-d24-int3}
\end{figure*}


\begin{figure*}[!t]
    \centering
    \includegraphics[width=\textwidth,page=1]{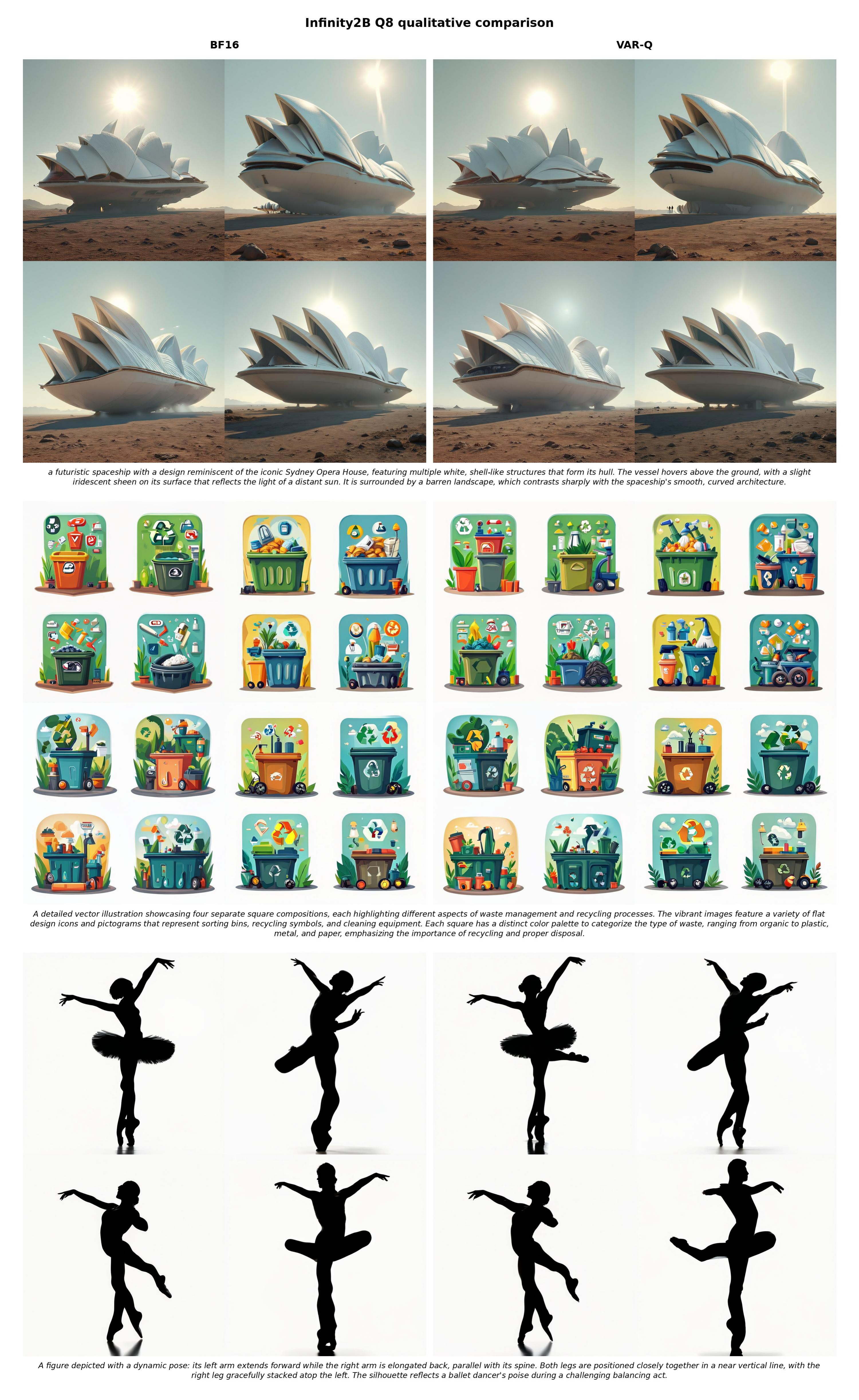}
    \caption{Qualitative comparison on Infinity 2B with INT8 KV cache.}
    \label{fig:infinity-2b-int8}
\end{figure*}

\begin{figure*}[!t]
    \centering
    \includegraphics[width=\textwidth,page=1]{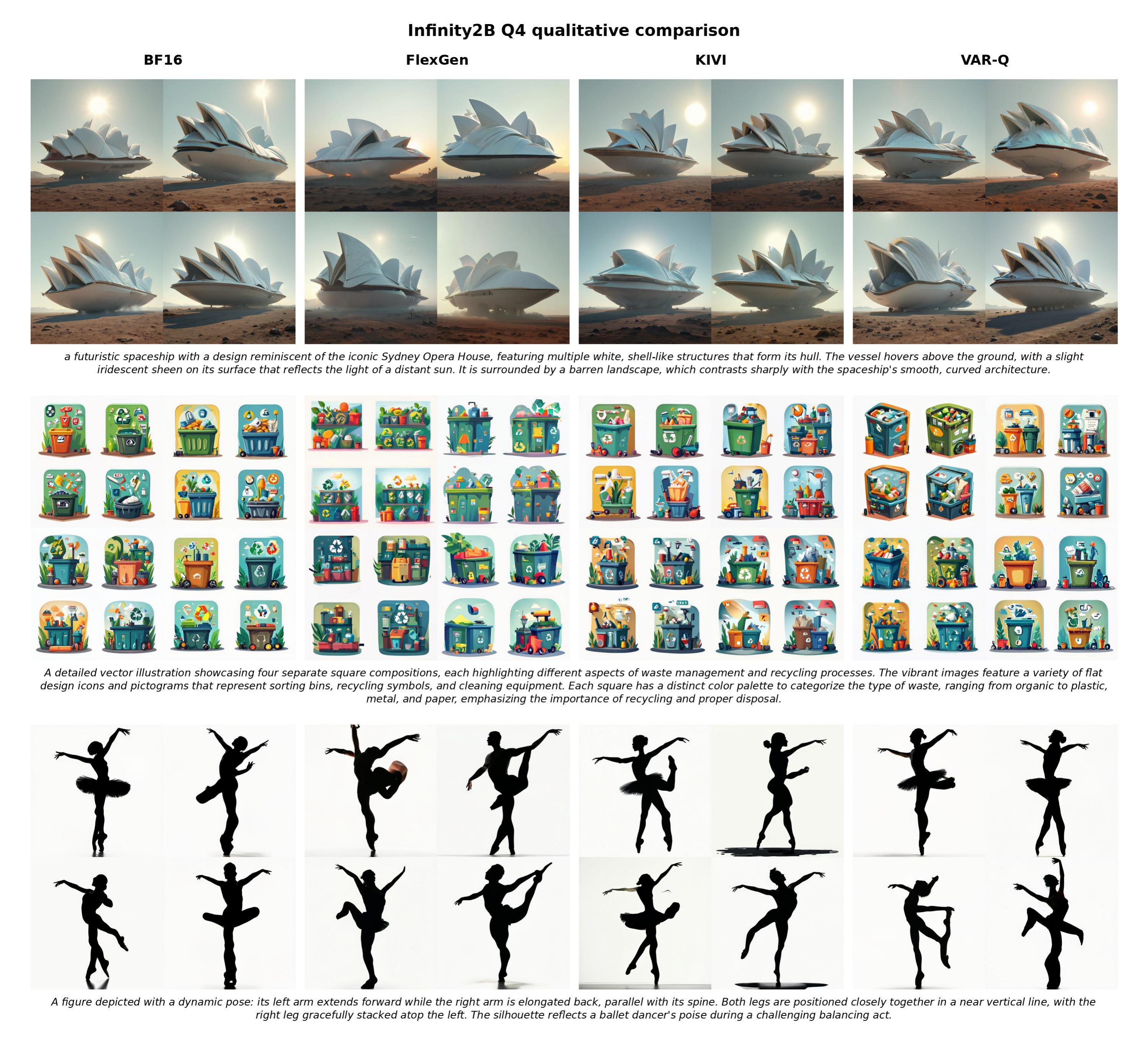}
    \caption{Qualitative comparison on Infinity 2B with INT4 KV cache.}
    \label{fig:infinity-2b-int4}
\end{figure*}

\begin{figure*}[!t]
    \centering
    \includegraphics[width=\textwidth,page=1]{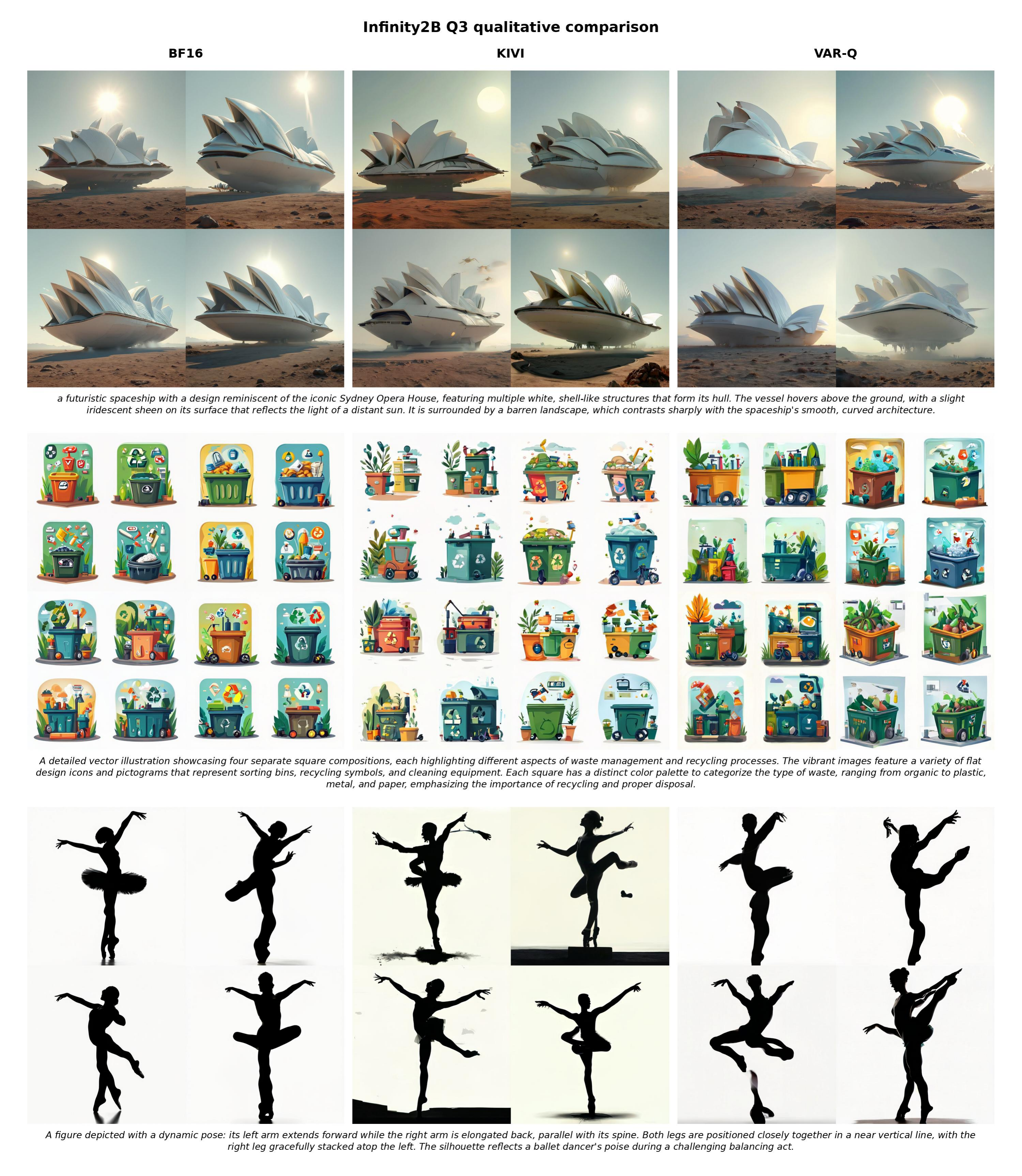}
    \caption{Qualitative comparison on Infinity 2B with INT3 KV cache.}
    \label{fig:infinity-2b-int3}
\end{figure*}

\begin{figure*}[!t]
    \centering
    \includegraphics[width=\textwidth,page=1]{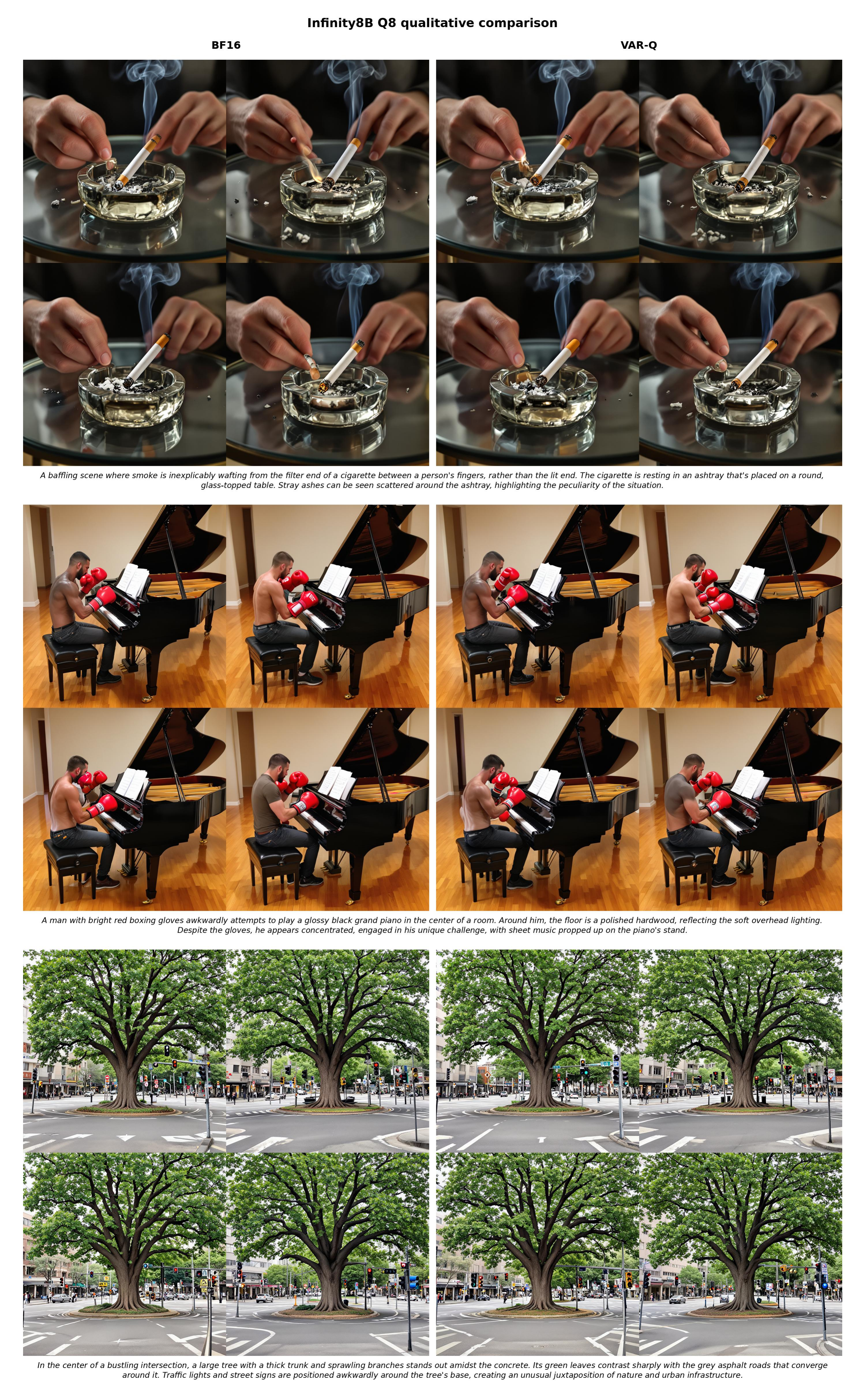}
    \caption{Qualitative comparison on Infinity 8B with INT8 KV cache.}
    \label{fig:infinity-8b-int8}
\end{figure*}

\begin{figure*}[!t]
    \centering
    \includegraphics[width=\textwidth,page=1]{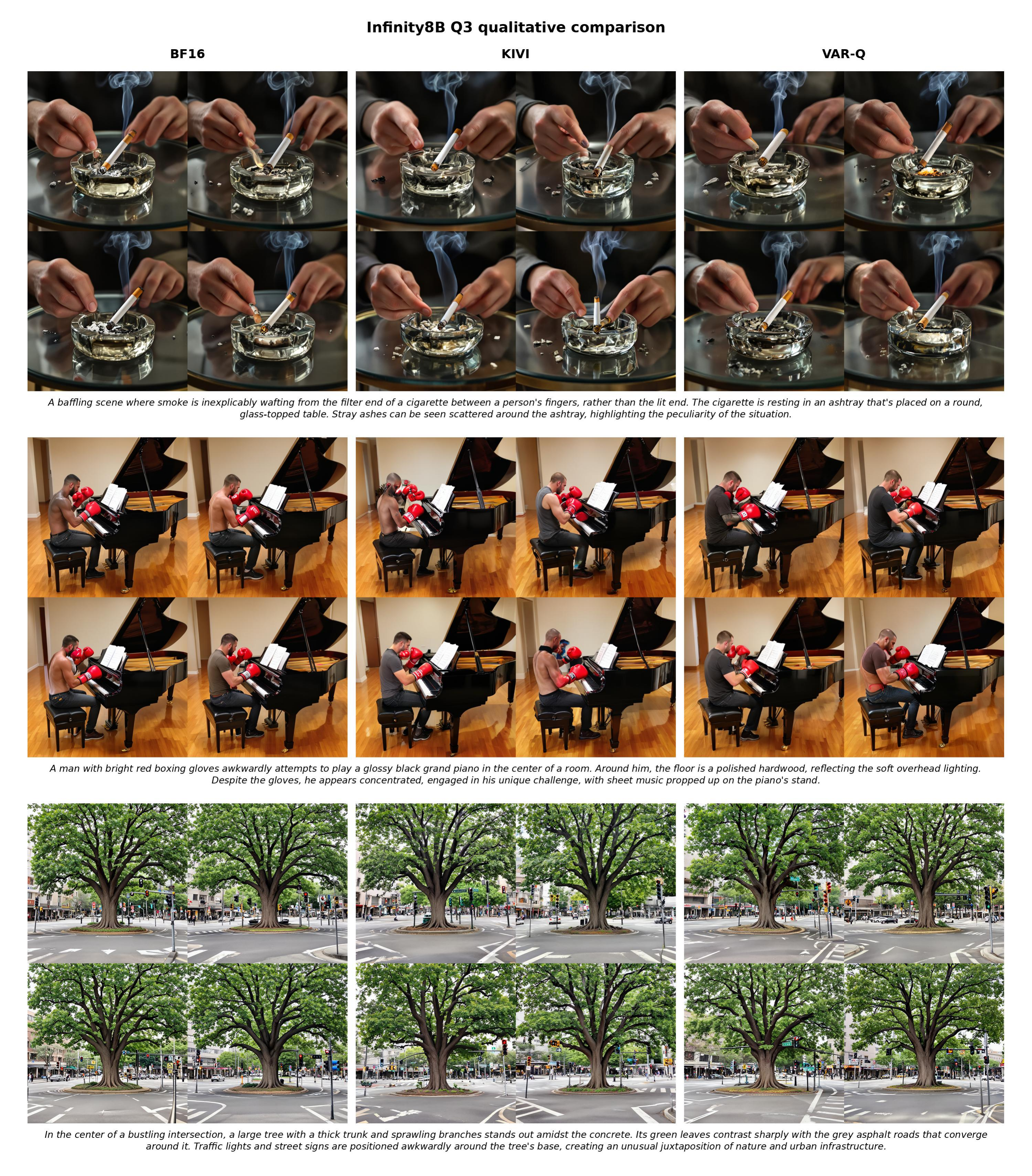}
    \caption{Qualitative comparison on Infinity 8B with INT3 KV cache.}
    \label{fig:infinity-8b-int3}
\end{figure*}

\begin{figure*}[!t]
    \centering
    \includegraphics[width=\textwidth,page=1]{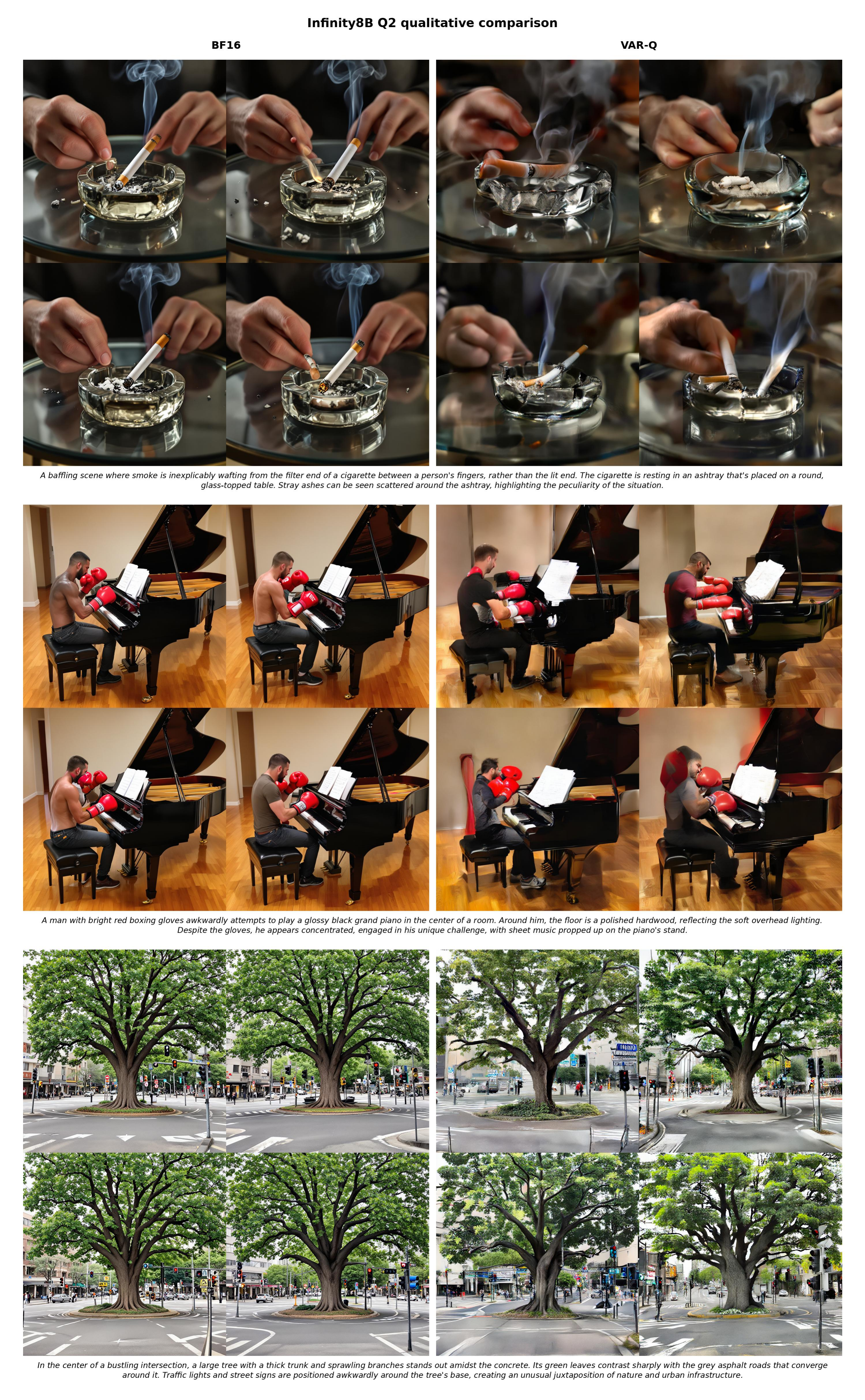}
    \caption{Qualitative comparison on Infinity 8B with INT2 KV cache.}
    \label{fig:infinity-8b-int2}
\end{figure*}
\clearpage
\paragraph{Next-scale video generation.}
For InfinityStar, the sampled frames show that quantization errors are more apparent in temporal consistency and imaging quality than in single-frame semantics. 
FlexGen and KIVI often preserve the rough scene content, but their later frames more frequently exhibit softened details or less stable background structure. 
VAR-Q better maintains the BF16 trajectory across frames, with smaller visible drift and more consistent subject/background appearance.


\begin{figure*}[!t]
    \centering
    \includegraphics[width=\textwidth,page=1]{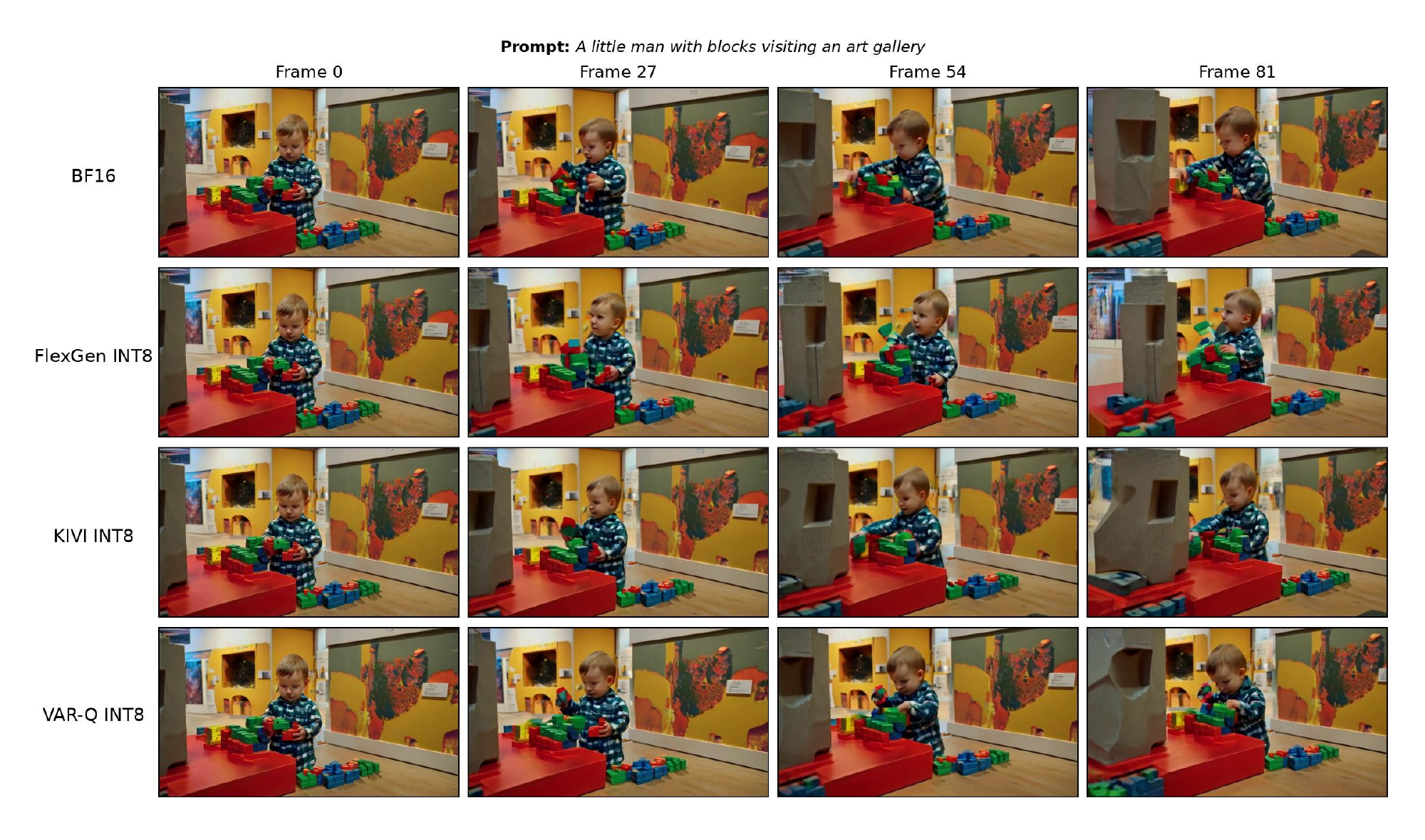}
    \caption{Qualitative comparison on InfinityStar 480p with INT8 KV cache.}
    \label{fig:infinitystar-480p-int8}
\end{figure*}

\begin{figure*}[!t]
    \centering
    \includegraphics[width=\textwidth,page=1]{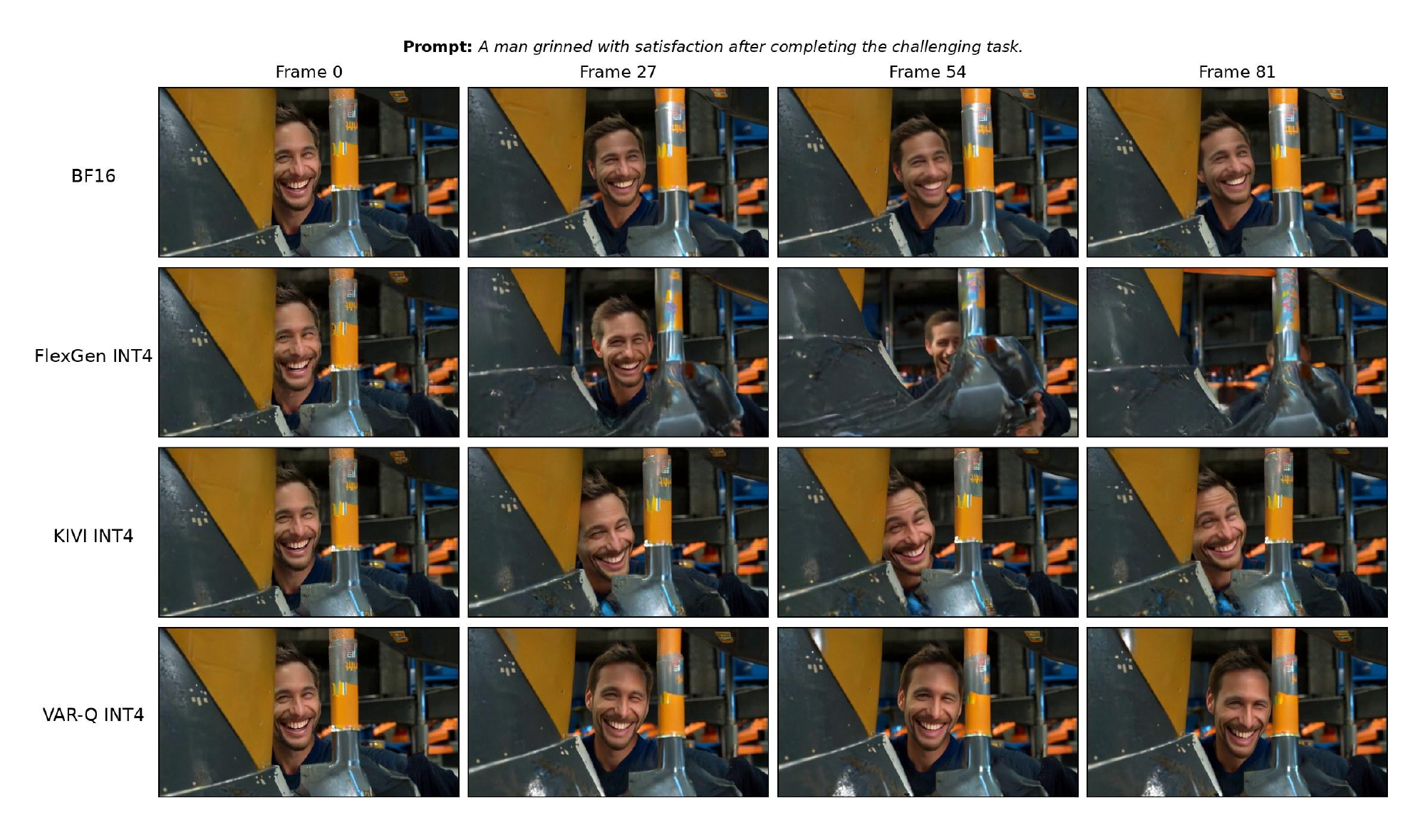}
    \caption{Qualitative comparison on InfinityStar 480p with INT4 KV cache.}
    \label{fig:infinitystar-480p-int4}
\end{figure*}

\begin{figure*}[!t]
    \centering
    \includegraphics[width=\textwidth,page=1]{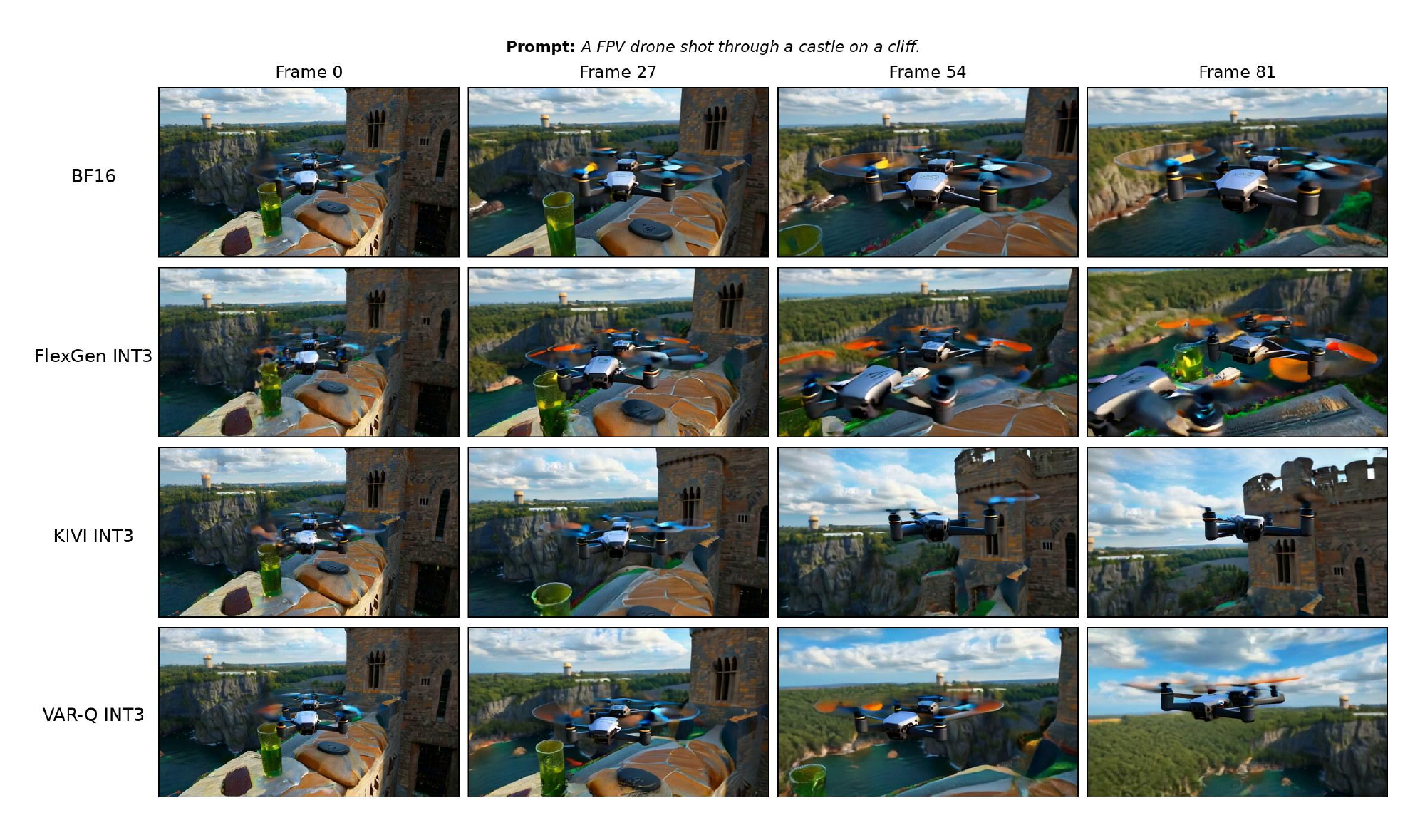}
    \caption{Qualitative comparison on InfinityStar 480p with INT3 KV cache.}
    \label{fig:infinitystar-480p-int3}
\end{figure*}

\begin{figure*}[!t]
    \centering
    \includegraphics[width=\textwidth,page=1]{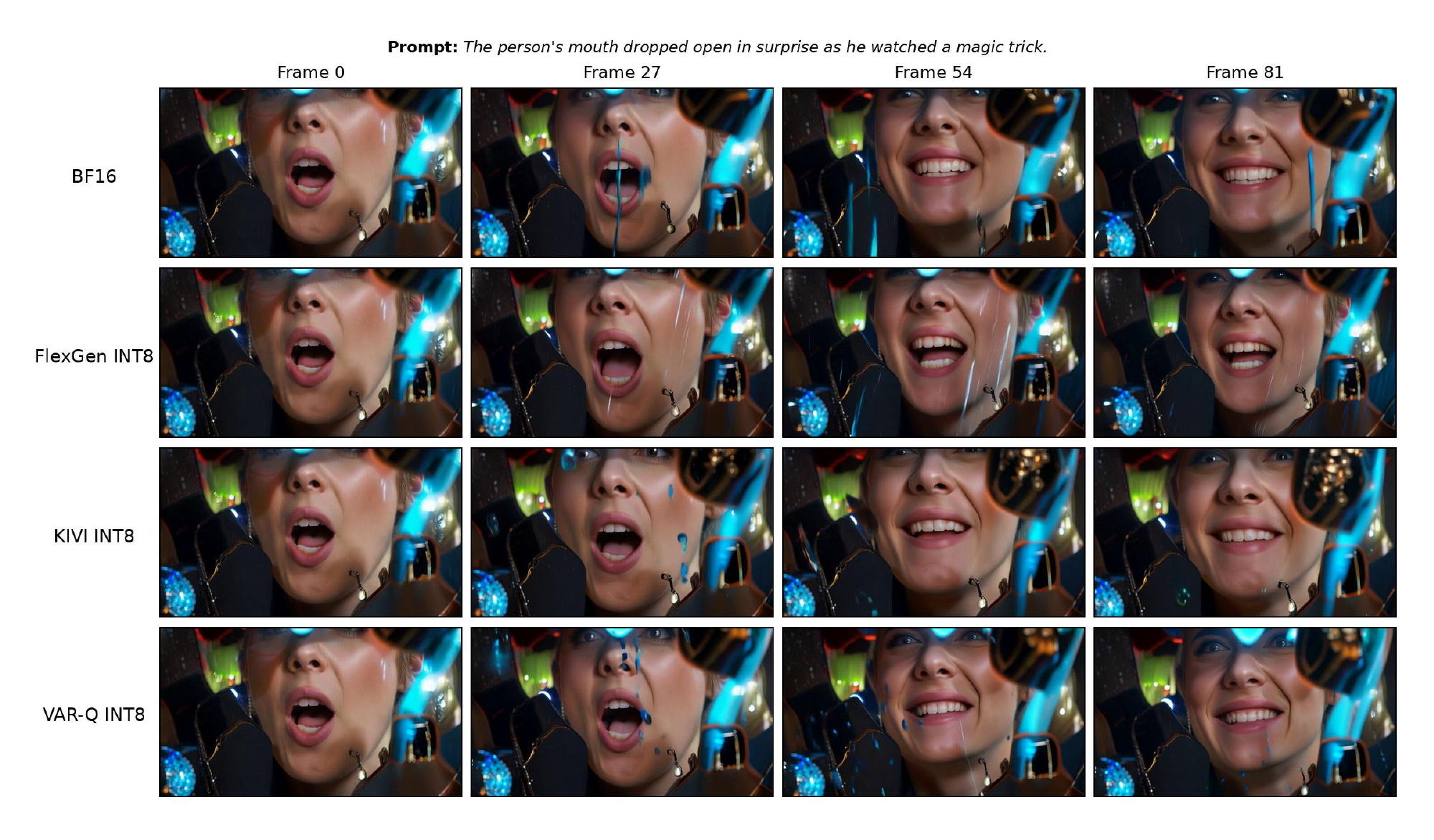}
    \caption{Qualitative comparison on InfinityStar 720p with INT8 KV cache.}
    \label{fig:infinitystar-720p-int8}
\end{figure*}

\begin{figure*}[!t]
    \centering
    \includegraphics[width=\textwidth,page=1]{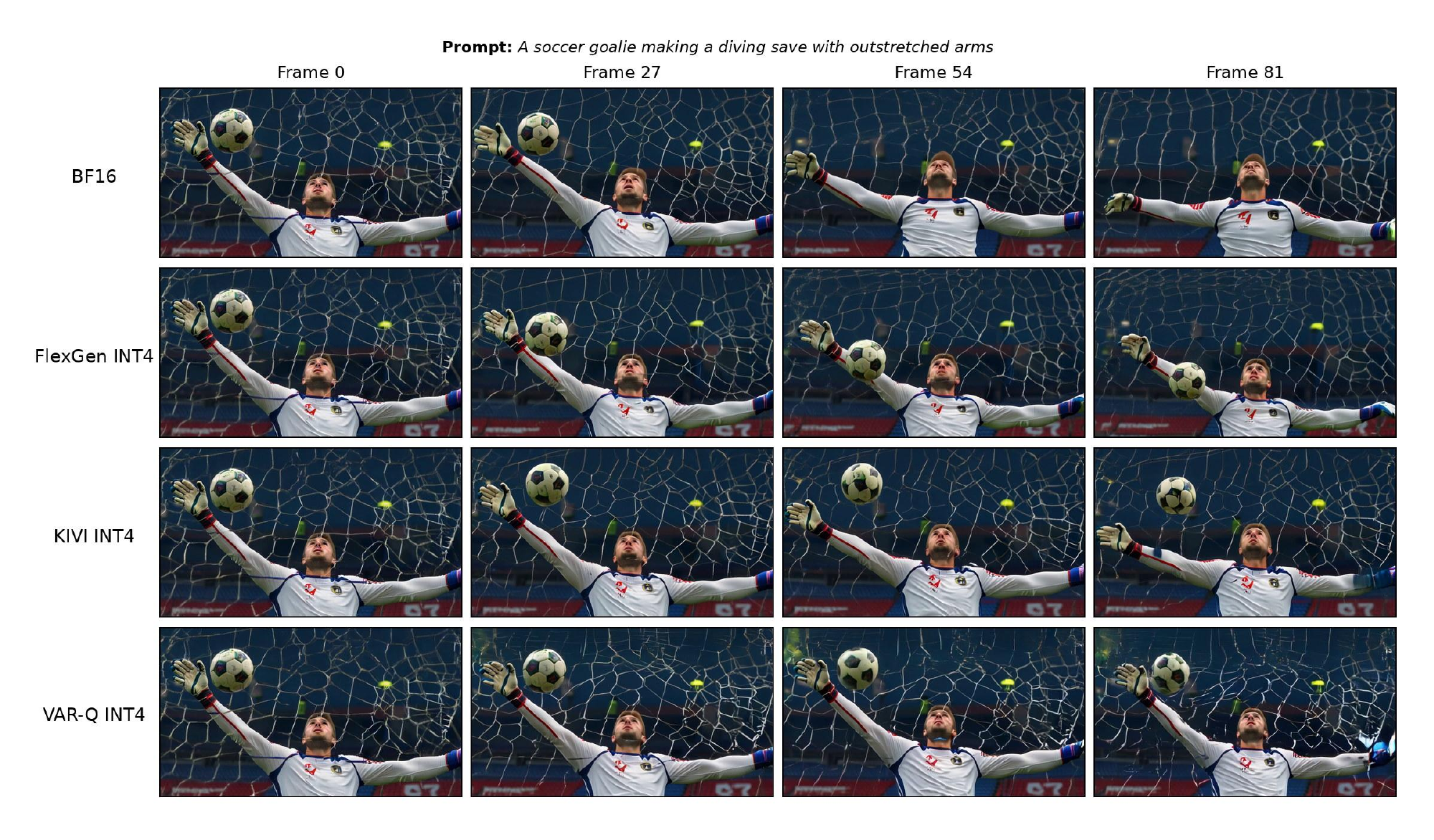}
    \caption{Qualitative comparison on InfinityStar 720p with INT4 KV cache.}
    \label{fig:infinitystar-720p-int4}
\end{figure*}

\begin{figure*}[!t]
    \centering
    \includegraphics[width=\textwidth,page=1]{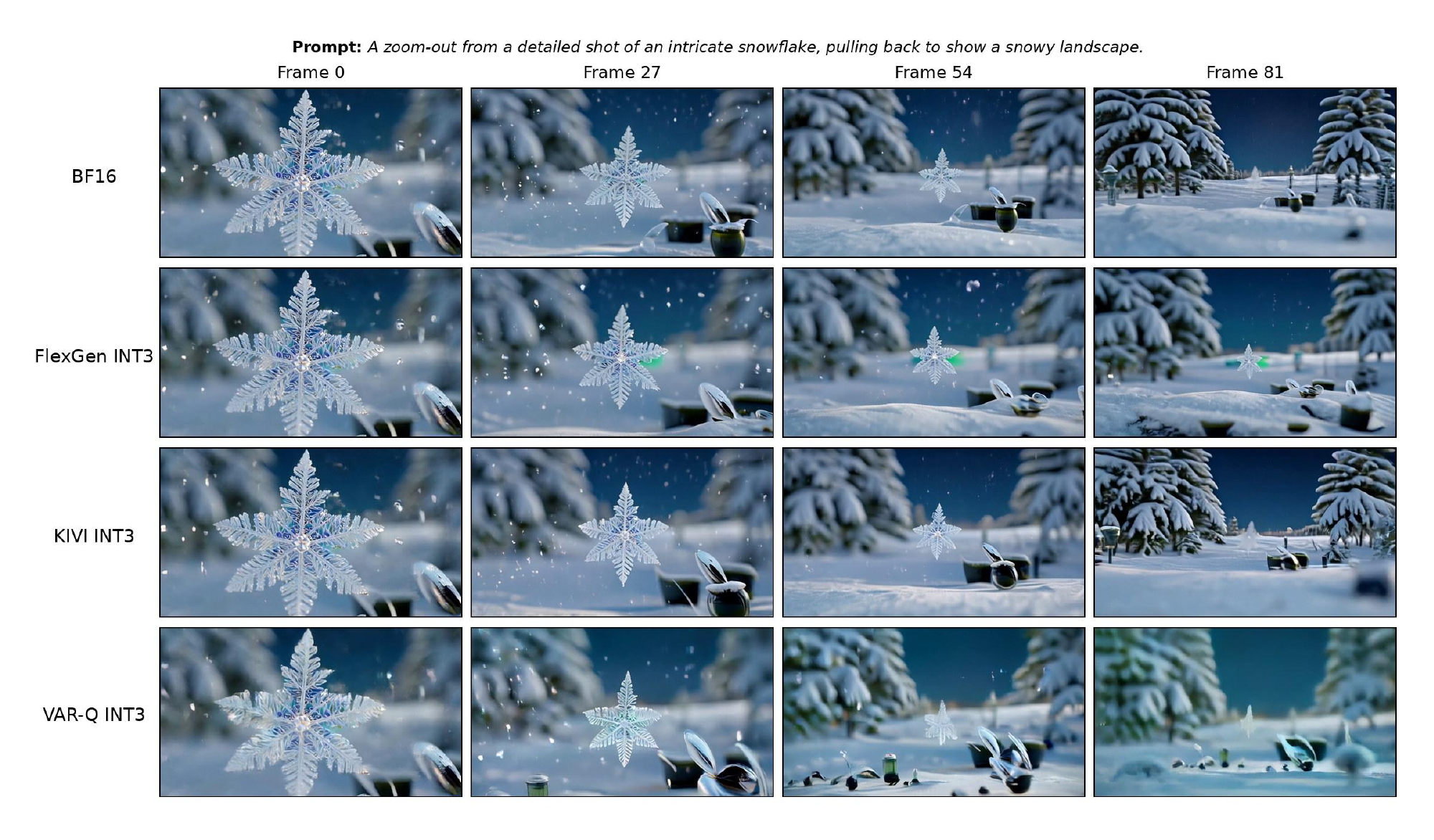}
    \caption{Qualitative comparison on InfinityStar 720p with INT3 KV cache.}
    \label{fig:infinitystar-720p-int3}
\end{figure*}
\clearpage
\paragraph{Next-frame video generation.}
For Self-Forcing and LongLive, the qualitative comparison emphasizes long-horizon stability. 
In the 5-second Self-Forcing examples, VAR-Q keeps motion and scene structure close to BF16 under INT6 and INT4, while the baselines show stronger degradation in texture sharpness and frame-to-frame consistency. 
In the 30-second LongLive examples, differences become more visible in later sampled frames: VAR-Q better preserves subject appearance and background configuration, whereas the baselines more often accumulate drift or smoothing over time.

\begin{figure*}[!t]
    \centering
    \includegraphics[width=\textwidth,page=3]{samples/self_forcing-q6.pdf}
    \caption{Qualitative comparison on Self-Forcing with INT6 KV cache.}
    \label{fig:self-forcing-int6-p3}
\end{figure*}

\begin{figure*}[!t]
    \centering
    \includegraphics[width=\textwidth,page=5]{samples/self_forcing-q6.pdf}
    \caption{Qualitative comparison on Self-Forcing with INT6 KV cache.}
    \label{fig:self-forcing-int6-p4}
\end{figure*}

\begin{figure*}[!t]
    \centering
    \includegraphics[width=\textwidth,page=8]{samples/self_forcing-q6.pdf}
    \caption{Qualitative comparison on Self-Forcing with INT6 KV cache.}
    \label{fig:self-forcing-int6-p8}
\end{figure*}

\begin{figure*}[!t]
    \centering
    \includegraphics[width=\textwidth,page=9]{samples/self_forcing-q6.pdf}
    \caption{Qualitative comparison on Self-Forcing with INT6 KV cache.}
    \label{fig:self-forcing-int6-p9}
\end{figure*}

\begin{figure*}[!t]
    \centering
    \includegraphics[width=\textwidth,page=1]{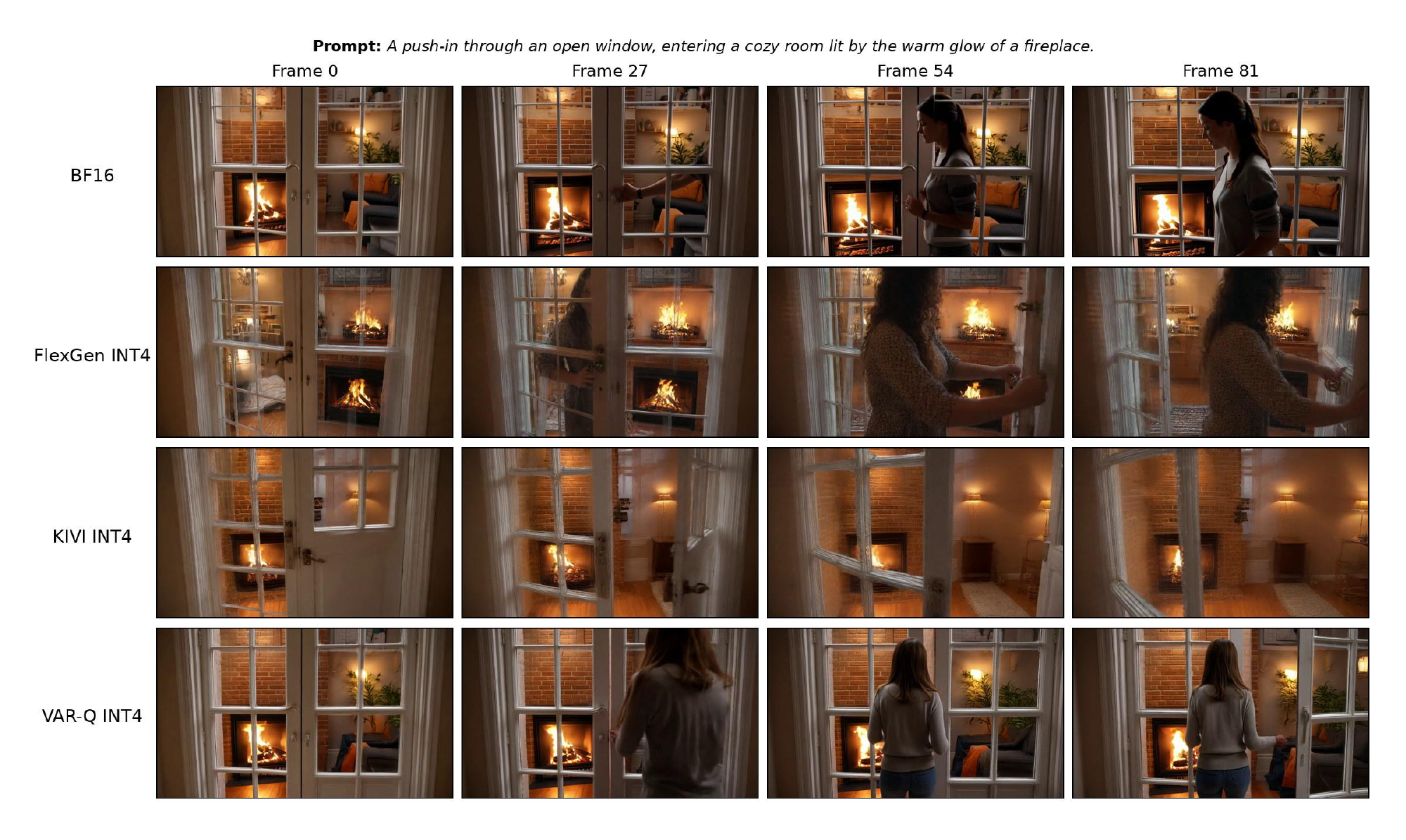}
    \caption{Qualitative comparison on Self-Forcing with INT4 KV cache.}
    \label{fig:self-forcing-int4-s1}
\end{figure*}

\begin{figure*}[!t]
    \centering
    \includegraphics[width=\textwidth,page=1]{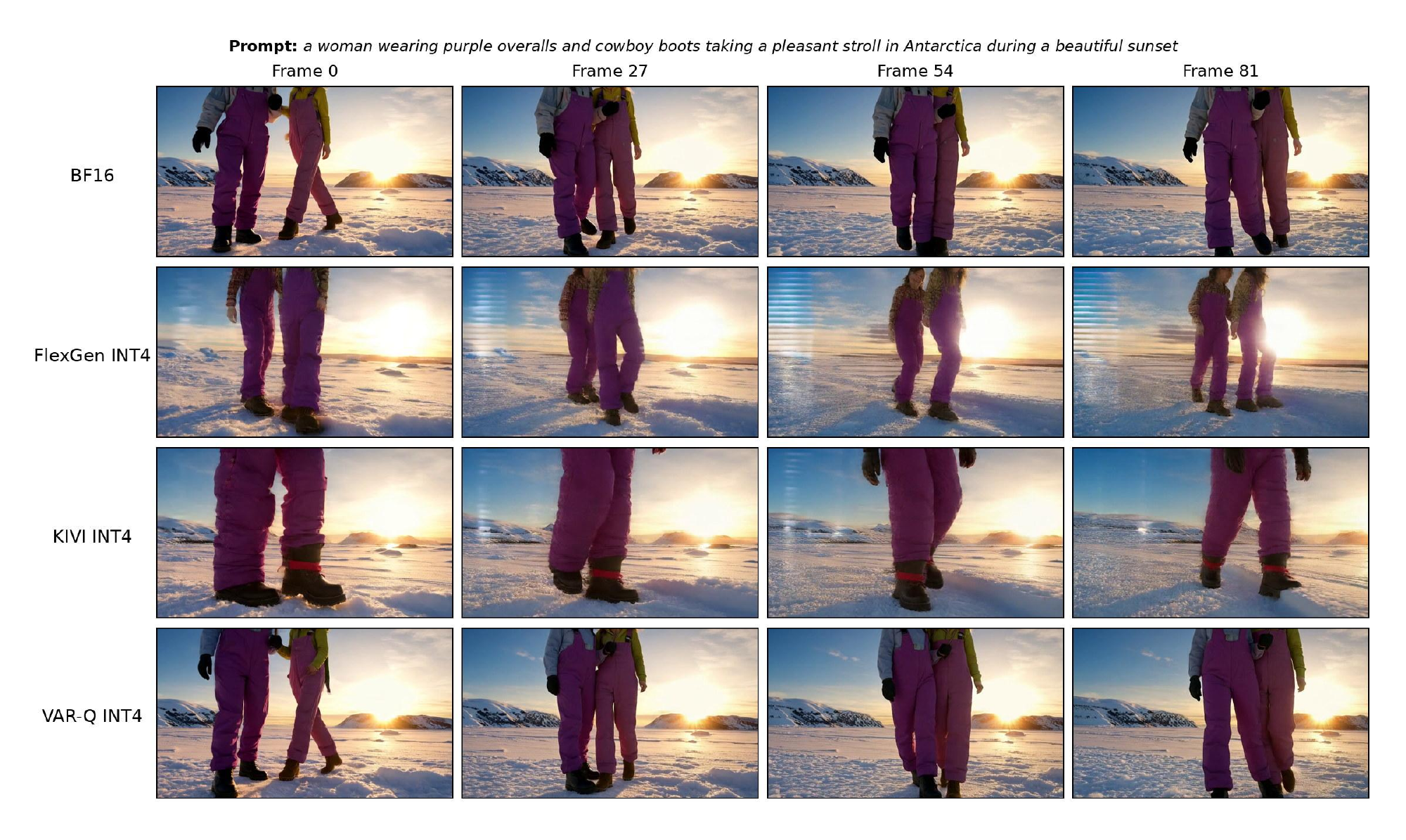}
    \caption{Qualitative comparison on Self-Forcing with INT4 KV cache.}
    \label{fig:self-forcing-int4-s2}
\end{figure*}

\begin{figure*}[!t]
    \centering
    \includegraphics[width=\textwidth,page=1]{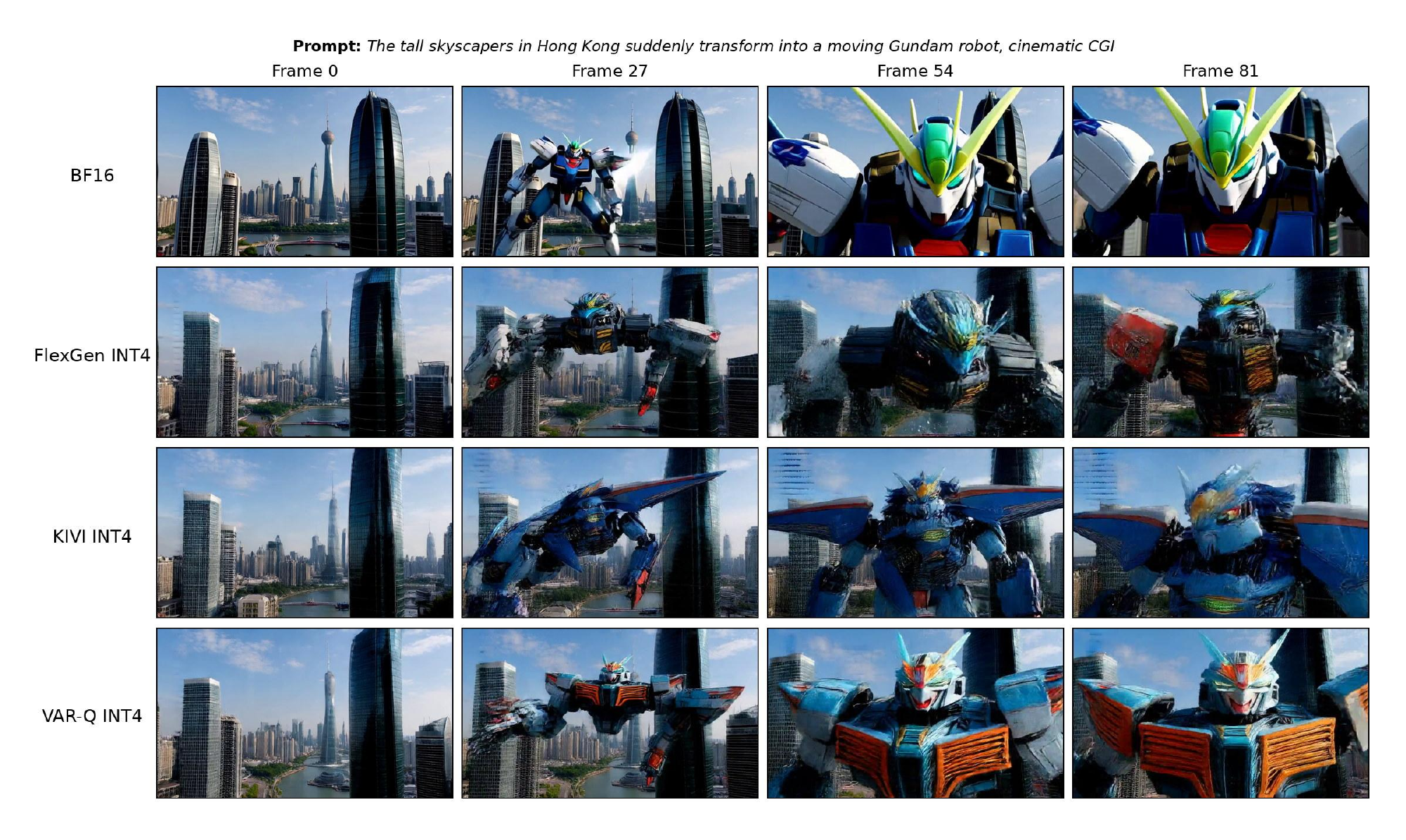}
    \caption{Qualitative comparison on Self-Forcing with INT4 KV cache.}
    \label{fig:self-forcing-int4-s3}
\end{figure*}


\begin{figure*}[!t]
    \centering
    \includegraphics[width=\textwidth,page=1]{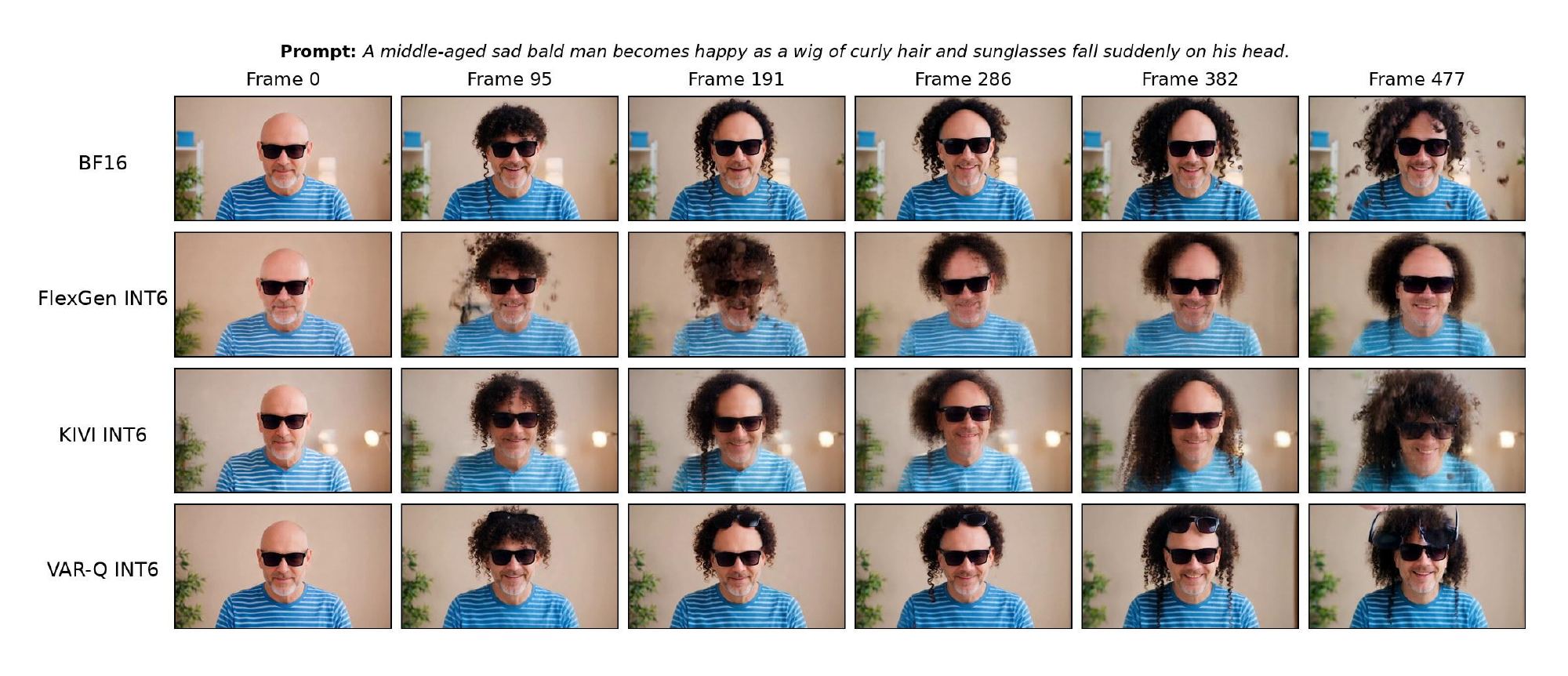}
    \caption{Qualitative comparison on LongLive with INT6 KV cache.}
    \label{fig:longlive-int6-p1}
\end{figure*}

\begin{figure*}[!t]
    \centering
    \includegraphics[width=\textwidth,page=2]{samples/longlive-q6.pdf}
    \caption{Qualitative comparison on LongLive with INT6 KV cache.}
    \label{fig:longlive-int6-p2}
\end{figure*}

\begin{figure*}[!t]
    \centering
    \includegraphics[width=\textwidth,page=6]{samples/longlive-q6.pdf}
    \caption{Qualitative comparison on LongLive with INT6 KV cache.}
    \label{fig:longlive-int6-p6}
\end{figure*}

\begin{figure*}[!t]
    \centering
    \includegraphics[width=\textwidth,page=9]{samples/longlive-q6.pdf}
    \caption{Qualitative comparison on LongLive with INT6 KV cache.}
    \label{fig:longlive-int6-p9}
\end{figure*}

\begin{figure*}[!t]
    \centering
    \includegraphics[width=\textwidth,page=1]{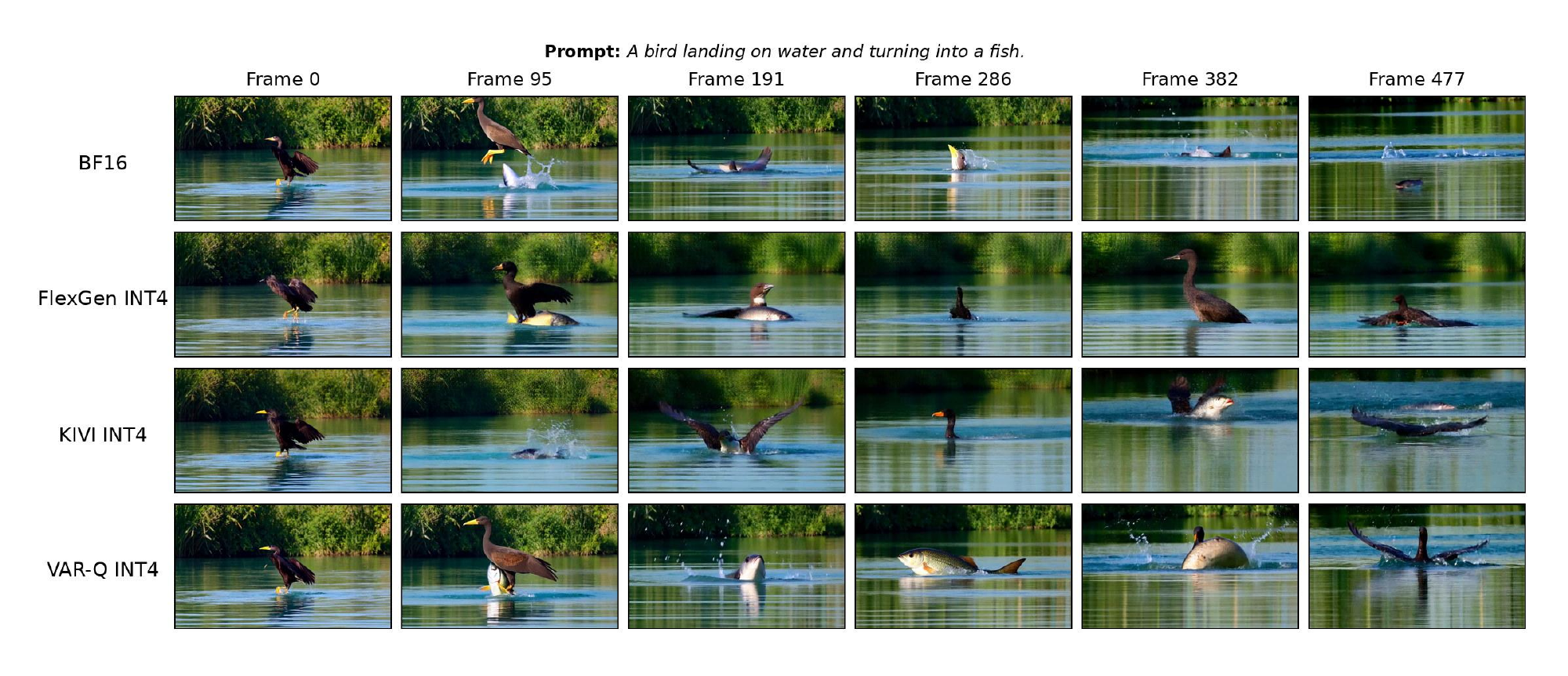}
    \caption{Qualitative comparison on LongLive with INT4 KV cache.}
    \label{fig:longlive-int4-s1}
\end{figure*}

\begin{figure*}[t]
    \centering
    \includegraphics[width=\textwidth,page=1]{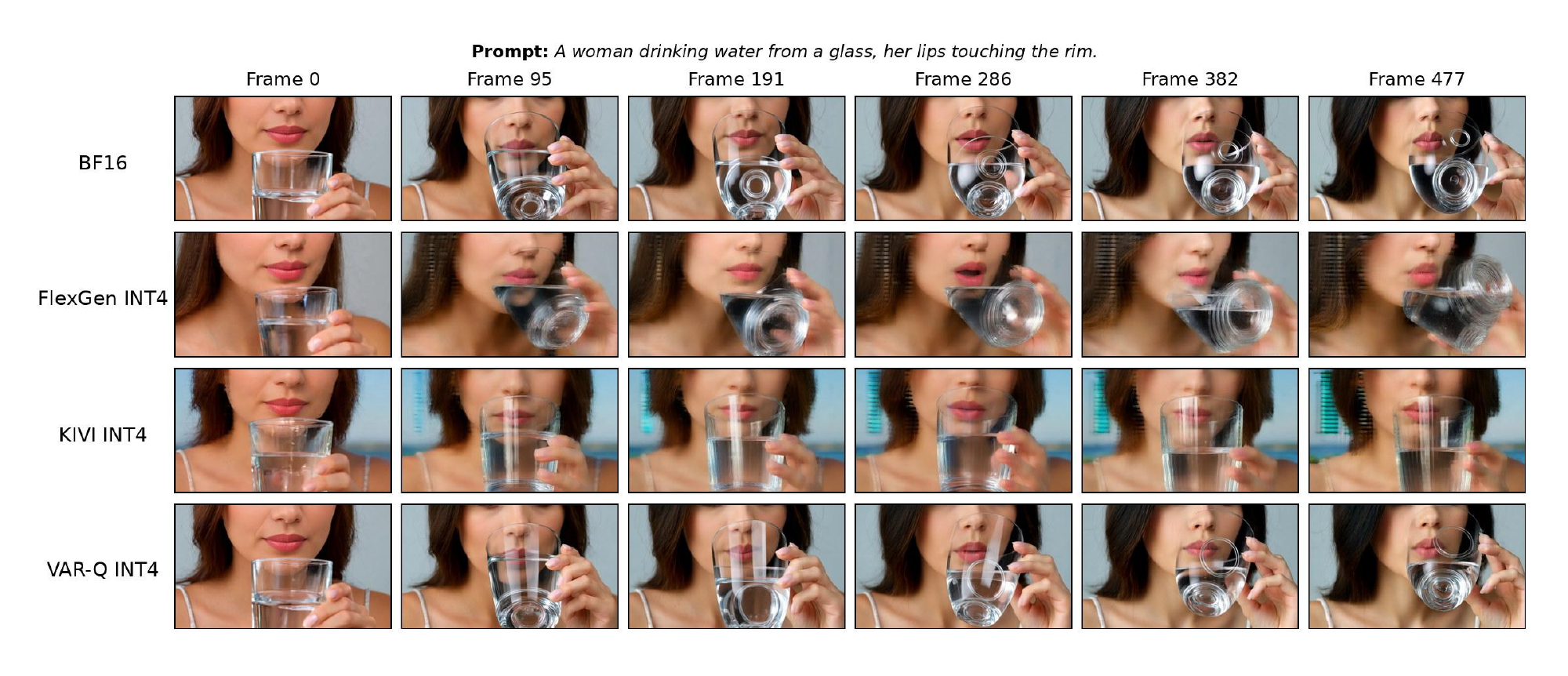}
    \caption{Qualitative comparison on LongLive with INT4 KV cache.}
    \label{fig:longlive-int4-s2}
\end{figure*}

\clearpage
\section{Evaluation Prompts Adapted from MovieGen}
\label{app:Prompts used in MovieGen}
For the MovieGen-based evaluation, we use a fixed subset of 200 prompts randomly sampled from the original \textbf{MovieGenVideoBench} prompt set. We adopt this reduced prompt set purely for efficiency, so that all methods can be evaluated under the same protocol with substantially lower inference cost and turnaround time. The sampled 200 prompts are kept identical across all experiments, models, and quantization settings to ensure fair comparison. No manual prompt selection or prompt-specific tuning is applied after sampling.
\input{Video_prompts}

\section{Broader Societal Impact}
\label{app:broader_impact}

The proposed VAR-Q framework significantly reduces the memory footprint and computational overhead of deploying large-scale visual autoregressive models. By compressing the KV cache by up to 78.5\% with negligible quality degradation, this work democratizes access to state-of-the-art image and video generation. Positive societal impacts include empowering researchers, independent creators, and under-resourced institutions to utilize high-fidelity visual AI on commodity hardware. This can foster innovation in creative industries, education, and accessibility tools. 

However, by drastically lowering the computational barrier to generating highly realistic synthetic media, this technology simultaneously exacerbates the risk of malicious applications. Bad actors can leverage these efficiency gains to mass-produce Deepfakes and realistic synthetic videos for disinformation campaigns, fraud, or non-consensual explicit content. Furthermore, because VAR-Q makes these models cheaper and faster to run, any inherent biases within the foundational models' training distributions (e.g., demographic misrepresentations) could be propagated more widely and rapidly, raising fairness concerns. 

To mitigate these negative impacts, we advocate that the deployment of highly efficient visual generators be coupled with robust provenance techniques, such as invisible watermarking and cryptographic media authentication, to ensure synthetic content remains identifiable. Additionally, developers of foundational visual AR models should consider gated access, strict licensing, or API-monitored deployments to prevent the automated, large-scale generation of harmful content, ensuring the broader community can safely benefit from the efficiency gains of methods like VAR-Q.

\clearpage

\end{document}

%% file: Tables/MSE.tex
\begin{wraptable}{r}{0.4\columnwidth}
\vspace{-12pt}
\caption{INT4 Quantization Errors (\%) on Infinity-2B, where $\epsilon_X \!=\! \|X - X'\|_F / \|X\|_F$.}
\label{tab:grouping_err}
\centering
\footnotesize
\setlength{\tabcolsep}{3pt}
\renewcommand{\arraystretch}{0.95}
\begin{tabular}{lcccc}
\toprule
& \multicolumn{2}{c}{Reconstruction} & \multicolumn{2}{c}{Attention} \\
\cmidrule(lr){2-3} \cmidrule(lr){4-5}
\textbf{Rule} & $\epsilon_K$ & $\epsilon_V$ & $\epsilon_A$ & $\epsilon_O$ \\
\midrule
Per-tensor               & 24.01 & 43.54 & 29.20 & 32.32 \\
Per-block                & 21.59 & 32.93 & 26.36 & 22.78 \\
Per-channel              & 12.11 & 18.07 & 16.17 & 13.13 \\
\textbf{Per-(ch., blk.)} & \textbf{10.04} & \textbf{15.28} & \textbf{13.38} & \textbf{10.13} \\
\bottomrule
\vspace{-10pt}
\end{tabular}
\end{wraptable}

%% file: Table/next_scale_image.tex
\begin{table*}[t]
\centering
\caption{Next-scale image generation quality under low-precision KV-cache quantization on VAR-d20/d24/d30 and Infinity-2B/8B. Our INT4 results closely match the BF16 baseline. Bold/underline: best/second-best per precision. KV (GB): realized cache size; subscripts: savings vs.\ BF16.}
\label{tab:next-scale-image}

\begin{subtable}{\textwidth}
\centering
\scriptsize
\setlength{\tabcolsep}{4pt}
\renewcommand{\arraystretch}{1.05}
\caption{VAR-d20/d24/d30~\cite{tian2024visual} on class-conditional ImageNet $256{\times}256$.}
\label{tab:next-scale-var}
\begin{tabular}{lclccccccccc}
\toprule
\multirow{2}{*}{Model} & \multirow{2}{*}{\#Params} & \multirow{2}{*}{Method} & \multirow{2}{*}{Prec.} & \multicolumn{4}{c}{Generation Quality} & \multicolumn{3}{c}{Fidelity to BF16} & \multirow{2}{*}{KV (GB)$\downarrow$} \\
\cmidrule(lr){5-8} \cmidrule(lr){9-11}
& & & & FID $\downarrow$ & IS $\uparrow$ & Precision $\uparrow$ & Recall $\uparrow$ & PSNR $\uparrow$ & SSIM $\uparrow$ & LPIPS $\downarrow$ & \\
\midrule
\multirow{7}{*}{VAR-d20} & \multirow{7}{*}{600M}
 & Baseline   & BF16 & 4.11 & 357.32 & 0.8650 & 0.5160 & --    & --    & --    & 16.60\sub{bs=128} \\
\cmidrule(lr){3-12}
 & & FlexGen    & INT4 & 4.03 & 325.39 & 0.8326 & 0.5230 & \underline{11.84} & \underline{0.264} & 0.553 & \textbf{4.29}\sub{-74.14\%} \\
 & & KIVI       & INT4 & \underline{3.55} & \underline{341.52} & \underline{0.8458} & \textbf{0.5363} & 11.81 & 0.256 & \underline{0.552} & 4.65\sub{-71.97\%} \\
 & & \ourmethod & INT4 & \textbf{3.44} & \textbf{342.77} & \textbf{0.8502} & \underline{0.5281} & \textbf{13.22} & \textbf{0.308} & \textbf{0.462} & \underline{4.64}\sub{-72.06\%} \\
 & & FlexGen    & INT3 & 15.33 & 139.59 & 0.5944 & \textbf{0.6308} & \underline{10.35} & \underline{0.212} & \underline{0.650} & \textbf{3.46}\sub{-79.14\%} \\
 & & KIVI       & INT3 & \underline{6.36} & \underline{233.84} & \underline{0.7260} & 0.5748 & 9.88  & 0.187 & 0.662 & 3.82\sub{-76.97\%} \\
 & & \ourmethod & INT3 & \textbf{3.77} & \textbf{262.24} & \textbf{0.7706} & \underline{0.5767} & \textbf{12.24} & \textbf{0.266} & \textbf{0.538} & \underline{3.81}\sub{-77.06\%} \\
\midrule
\multirow{7}{*}{VAR-d24} & \multirow{7}{*}{1.0B}
 & Baseline   & BF16 & 2.97 & 357.24 & 0.8444 & 0.5605 & --    & --    & --    & 23.91\sub{bs=128} \\
\cmidrule(lr){3-12}
 & & FlexGen    & INT4 & 2.74 & 325.69 & 0.8175 & \textbf{0.5816} & 11.78 & 0.256 & 0.560 & \textbf{6.18}\sub{-74.14\%} \\
 & & KIVI       & INT4 & \underline{2.56} & \underline{340.55} & \underline{0.8295} & \underline{0.5786} & \underline{11.92} & \underline{0.268} & \underline{0.545} & 6.70\sub{-71.97\%} \\
 & & \ourmethod & INT4 & \textbf{2.52} & \textbf{341.60} & \textbf{0.8321} & 0.5718 & \textbf{13.03} & \textbf{0.296} & \textbf{0.479} & \underline{6.68}\sub{-72.06\%} \\
  & & FlexGen    & INT3 & 14.91 & 144.09 & 0.6066 & \textbf{0.6424} & 10.63 & 0.201 & 0.643 & \textbf{4.99}\sub{-79.14\%} \\
 & & KIVI       & INT3 & \underline{4.29} & \underline{239.40} & \underline{0.7271} & \underline{0.6340} & \underline{10.88} & \underline{0.226} & \underline{0.631} & 5.51\sub{-76.97\%} \\
 & & \ourmethod & INT3 & \textbf{3.60} & \textbf{258.33} & \textbf{0.7477} & 0.6130 & \textbf{11.90} & \textbf{0.252} & \textbf{0.558} & \underline{5.48}\sub{-77.06\%} \\
\midrule
\multirow{7}{*}{VAR-d30} & \multirow{7}{*}{2.0B}
 & Baseline   & BF16 & 2.07 & 335.10 & 0.8260 & 0.5812 & --    & --    & --    & 37.35\sub{bs=128} \\
\cmidrule(lr){3-12}
 & & FlexGen    & INT4 & 2.00 & 307.92 & 0.8061 & \textbf{0.6006} & 12.47 & 0.277 & 0.510 & \textbf{9.66}\sub{-74.14\%} \\
 & & KIVI       & INT4 & \underline{1.99} & \underline{321.46} & \textbf{0.8162} & \underline{0.5974} & \underline{12.86} & \underline{0.290} & \underline{0.484} & 10.47\sub{-71.97\%} \\
 & & \ourmethod & INT4 & \textbf{1.97} & \textbf{323.67} & \underline{0.8156} & 0.5961 & \textbf{13.54} & \textbf{0.312} & \textbf{0.437} & \underline{10.44}\sub{-72.06\%} \\
 & & FlexGen    & INT3 & 9.52 & 180.57 & 0.6738 & \textbf{0.6417} & 11.33 & 0.232 & 0.597 & \textbf{7.79}\sub{-79.14\%} \\
 & & KIVI       & INT3 & \underline{3.26} & \underline{264.93} & \underline{0.7536} & 0.6189 & \underline{11.69} & \underline{0.249} & \underline{0.573} & 8.60\sub{-76.97\%} \\
 & & \ourmethod & INT3 & \textbf{3.05} & \textbf{268.16} & \textbf{0.7612} & \underline{0.6307} & \textbf{12.41} & \textbf{0.270} & \textbf{0.517} & \underline{8.57}\sub{-77.06\%} \\
\bottomrule
\end{tabular}
\end{subtable}

\vspace{8pt}

\begin{subtable}{\textwidth}
\centering
\scriptsize
\setlength{\tabcolsep}{2pt}
\renewcommand{\arraystretch}{1.05}
\caption{Infinity-2B/8B~\cite{han2025infinity} on text-to-image generation.}
\label{tab:next-scale-infinity}
\begin{tabular}{llc*{11}{c}}
\toprule
\multirow{2}{*}{Model} & \multirow{2}{*}{Method} & \multirow{2}{*}{Prec.} & \multicolumn{4}{c}{GenEval~\cite{ghosh2023geneval} $\uparrow$} & \multicolumn{6}{c}{DPG~\cite{hu2024ella} $\uparrow$} & \multirow{2}{*}{KV (GB)$\downarrow$} \\
\cmidrule(lr){4-7} \cmidrule(lr){8-13}
& & & Two-Object & Position & Color & Overall & Entity & Relation & Attribute & Global & Other & Overall & \\
\midrule

\multirow{8}{*}{Infinity-2B}
 & Baseline   & BF16 & 80.61 & 25.51 & 88.03 & 69.61 & 86.55 & 89.51 & 86.62 & 83.24 & 90.14 & 81.97 & 41.10\sub{bs=8} \\
\cmidrule(lr){2-14}
 & \ourmethod & INT8 & 79.34 & 26.26 & 87.23 & 69.06 & 87.69 & 89.39 & 88.70 & 86.40 & 85.85 & 82.21 & 20.65\sub{-49.75\%} \\
 & FlexGen    & INT4 & 78.57 & \underline{24.75} & 54.50 & \underline{67.75} & \underline{87.39} & \textbf{91.56} & 85.38 & 82.24 & \underline{82.04} & \underline{81.01} & \underline{10.60}\sub{-74.22\%} \\
 & KIVI       & INT4 & \textbf{80.61} & 24.49 & \textbf{87.23} & 67.48 & \textbf{88.18} & \underline{91.13} & \underline{86.51} & \underline{83.18} & 80.81 & 80.04 & 10.78\sub{-73.76\%} \\
 & \ourmethod & INT4 & \underline{79.34} & \textbf{26.01} & \underline{86.97} & \textbf{69.35} & 87.24 & 89.60 & \textbf{87.43} & \textbf{86.75} & \textbf{90.13} & \textbf{82.02} & \textbf{10.38}\sub{-74.75\%} \\
 & KIVI       & INT3 & \underline{77.30} & \textbf{26.26} & \underline{86.70} & \underline{67.56} & \underline{87.12} & \textbf{92.39} & \underline{85.66} & \underline{82.94} & \underline{79.31} & \underline{80.76} & \underline{8.73}\sub{-78.76\%} \\
 & \ourmethod & INT3 & \textbf{78.32} & \underline{24.24} & \textbf{88.03} & \textbf{68.04} & \textbf{87.74} & \underline{89.92} & \textbf{87.01} & \textbf{85.13} & \textbf{87.01} & \textbf{81.18} & \textbf{8.32}\sub{-79.75\%} \\

\midrule

\multirow{8}{*}{Infinity-8B}
 & Baseline   & BF16 & 95.41 & 62.12 & 90.43 & 83.54 & 88.27 & 89.30 & 91.57 & 92.83 & 88.61 & 85.72 & 67.43\sub{bs=6} \\
\cmidrule(lr){2-14}
 & \ourmethod & INT8 & 96.43 & 62.88 & 89.89 & 84.00 & 90.57 & 92.57 & 89.23 & 87.12 & 92.68 & 85.97 & 33.88\sub{-49.75\%} \\
 & FlexGen    & INT4 & 95.41 & 60.89 & 89.02 & 82.45 & \underline{90.75} & \underline{92.33} & \underline{89.15} & 85.04 & \underline{84.49} & 84.58 & \underline{17.38}\sub{-74.22\%} \\
 & KIVI       & INT4 & \underline{95.66} & \textbf{65.40} & \underline{91.22} & \underline{83.26} & \textbf{91.13} & \textbf{94.10} & 88.93 & \underline{85.67} & 82.04 & \underline{85.74} & 17.69\sub{-73.76\%} \\
 & \ourmethod & INT4 & \textbf{95.92} & \underline{63.64} & \textbf{91.49} & \textbf{83.67} & 90.70 & 91.99 & \textbf{91.07} & \textbf{87.47} & \textbf{89.23} & \textbf{85.90} & \textbf{17.03}\sub{-74.75\%} \\
 & KIVI       & INT3 & \textbf{95.66} & \textbf{63.38} & \textbf{91.49} & \underline{82.75} & \underline{87.74} & \textbf{94.33} & \underline{88.97} & \underline{84.05} & \underline{85.71} & \underline{84.89} & \underline{14.32}\sub{-78.76\%} \\
 & \ourmethod & INT3 & \underline{95.41} & \underline{63.13} & \underline{90.69} & \textbf{83.28} & \textbf{89.65} & \underline{88.84} & \textbf{90.82} & \textbf{89.16} & \textbf{91.80} & \textbf{85.89} & \textbf{13.65}\sub{-79.75\%} \\
 & \ourmethod & INT2 & 85.97 & 56.57 & 86.70 & 74.38 & 87.80 & 90.09 & 89.31 & 91.22 & 88.12 & 83.83 & 8.60\sub{-87.25\%} \\

\bottomrule
\end{tabular}
\end{subtable}

\end{table*}

%% file: Table/next_scale_frame_video.tex
\begin{table*}[t]
\centering
\caption{\textbf{Next-scale video generation on InfinityStar.} KV reports cache size (GB) with reduction vs.\ BF16 in subscript. \textbf{Bold} and \underline{underline} indicate best and second-best, respectively.}
\label{tab:video-nextscale}
\scriptsize
\setlength{\tabcolsep}{2pt}
\renewcommand{\arraystretch}{1.06}
\begin{tabular}{ll*{10}{c}}
\toprule
\multirow{3}{*}{Model} & \multirow{3}{*}{Method} & \multirow{3}{*}{Prec.} & \multicolumn{8}{c}{VBench~\cite{huang2024vbench} $\uparrow$} & \multirow{3}{*}{KV (GB)$\downarrow$} \\
\cmidrule(lr){4-11}
& & & Subject & Background & Temporal & Motion & Dynamic & Aesthetic & Imaging & Total & \\
& & & Consistency & Consistency & Flickering & Smoothness & Degree & Quality & Quality & Quality & \\
\midrule
\multirow{7}{*}{\shortstack{InfinityStar\\8B 720p}}
 & Baseline   & BF16 & 93.69 & 94.62 & 97.71 & 98.86 & 67.50 & 58.49 & 66.79 & 82.33 & 62.72\sub{bs=3} \\
\cmidrule(lr){2-12}
 & FlexGen    & INT4 & \textbf{93.83} & \textbf{94.76} & \textbf{97.83} & \textbf{98.91} & 63.00 & 55.50 & \textbf{64.80} & 80.11 & \underline{16.48}\sub{-73.72\%} \\
 & KIVI       & INT4 & 92.01 & \underline{94.61} & 97.27 & \underline{98.70} & \underline{68.00} & \underline{56.12} & 63.60 & \underline{81.24} & 16.73\sub{-73.33\%} \\
 & \ourmethod & INT4 & \underline{92.65} & 94.09 & \underline{97.37} & 98.59 & \textbf{74.50} & \textbf{56.87} & \underline{64.35} & \textbf{81.66} & \textbf{16.00}\sub{-74.49\%} \\
 & FlexGen    & INT3 & 90.12 & \underline{92.68} & \textbf{97.84} & \underline{96.91} & 62.00 & \textbf{56.01} & 60.48 & \underline{79.95} & \underline{13.56}\sub{-78.37\%} \\
 & KIVI       & INT3 & \underline{90.45} & 90.42 & \underline{97.65} & 96.79 & \underline{74.00} & 50.97 & \textbf{61.92} & 75.46 & 13.81\sub{-77.99\%} \\
 & \ourmethod & INT3 & \textbf{91.41} & \textbf{93.56} & 97.25 & \textbf{98.37} & \textbf{75.00} & \underline{55.55} & \underline{61.75} & \textbf{80.59} & \textbf{13.08}\sub{-79.15\%} \\

\midrule

\multirow{7}{*}{\shortstack{InfinityStar\\8B 480p}}
 & Baseline   & BF16 & 93.01 & 94.53 & 97.25 & 98.68 & 66.00 & 58.13 & 64.01 & 81.30 & 55.87\sub{bs=6} \\
\cmidrule(lr){2-12}
 & FlexGen    & INT4 & 90.59 & 94.39 & 95.16 & \underline{98.64} & \underline{66.50} & 56.48 & 62.14 & 78.94 & \underline{15.03}\sub{-73.10\%} \\
 & KIVI       & INT4 & \underline{92.44} & \underline{94.42} & \underline{96.65} & 97.79 & 64.00 & \underline{56.97} & \underline{62.92} & \underline{80.46} & 15.25\sub{-72.71\%} \\
 & \ourmethod & INT4 & \textbf{92.75} & \textbf{94.51} & \textbf{97.21} & \textbf{98.67} & \textbf{68.50} & \textbf{57.84} & \textbf{63.62} & \textbf{81.32} & \textbf{14.60}\sub{-73.86\%} \\
 & FlexGen    & INT3 & 91.89 & 93.91 & 97.24 & 98.62 & \underline{64.00} & \underline{57.54} & \textbf{63.69} & \underline{80.64} & \underline{12.45}\sub{-77.71\%} \\
 & KIVI       & INT3 & \textbf{93.01} & \textbf{94.46} & \textbf{97.33} & \textbf{98.71} & \underline{64.00} & \textbf{57.79} & \underline{63.56} & 78.05 & 12.67\sub{-77.33\%} \\
 & \ourmethod & INT3 & \underline{92.86} & \underline{94.30} & \underline{97.26} & \underline{98.64} & \textbf{65.50} & 57.52 & 63.37 & \textbf{80.98} & \textbf{12.03}\sub{-78.47\%} \\

\bottomrule
\end{tabular}
\end{table*}

\begin{table*}[t]
\centering
\caption{\textbf{Next-frame video generation on Self-Forcing (5\,s) and LongLive (30\,s).} 
KV reports cache size (GB) with reduction vs.\ BF16 in subscript. On generation quality, \textbf{bold} and \underline{underline} indicate best and second-best, respectively.}
\label{tab:video-nextframe}
\scriptsize
\setlength{\tabcolsep}{2pt}
\renewcommand{\arraystretch}{1.06}
\begin{tabular}{ll*{11}{c}}
\toprule
\multirow{3}{*}{Model} & \multirow{3}{*}{Method} & \multirow{3}{*}{Prec.} & \multicolumn{6}{c}{VBench~\cite{huang2024vbench} $\uparrow$} & \multicolumn{3}{c}{Opus4.7-VL $\uparrow$} & \multirow{3}{*}{KV (GB)$\downarrow$} \\
\cmidrule(lr){4-9} \cmidrule(lr){10-12}
& & & Subject & Background & Motion & Dynamic & Aesthetic & Total & Visual & Dynamic & Text & \\
& & & Consistency & Consistency & Smoothness & Degree & Quality & Quality & & & & \\
\midrule

\multirow{7}{*}{\shortstack{Self-Forcing\\1.3B\\(5-second)}}
 & Baseline   & BF16 & 93.57 & 94.07 & 98.43 & 75.00 & 60.07 & 82.94 & 3.81 & 3.10 & 3.62 & 67.49\sub{bs=12} \\
\cmidrule(lr){2-13}
 & FlexGen    & INT6 & 93.52 & 94.11 & 98.44 & \underline{76.00} & 60.02 & \underline{82.98} & 3.79 & \underline{3.15} & \textbf{3.73} & \underline{27.95}\sub{-58.59\%} \\
 & KIVI       & INT6 & \underline{93.58} & \underline{94.12} & \textbf{98.49} & 73.00 & \underline{60.07} & 82.85 & \underline{3.82} & 3.14 & \underline{3.63} & 28.22\sub{-58.18\%} \\
 & \ourmethod & INT6 & \textbf{93.96} & \textbf{94.76} & \textbf{98.49} & \textbf{76.50} & \textbf{61.25} & \textbf{83.34} & \textbf{3.89} & \textbf{3.17} & 3.62 & \textbf{27.50}\sub{-59.25\%} \\

 & FlexGen    & INT4 & 86.39 & 85.84 & 95.40 & 73.00 & 57.39 & 75.91 & \textbf{3.60} & 3.40 & \underline{3.75} & \underline{17.40}\sub{-74.22\%} \\
 & KIVI       & INT4 & \underline{90.53} & \underline{91.86} & \underline{98.29} & \underline{74.50} & \textbf{57.64} & \underline{80.33} & 3.54 & \underline{3.44} & \textbf{3.76} & 17.68\sub{-73.81\%} \\
 & \ourmethod & INT4 & \textbf{91.28} & \textbf{92.55} & \textbf{98.35} & \textbf{75.50} & \underline{57.40} & \textbf{80.99} & \underline{3.55} & \textbf{3.50} & 3.73 & \textbf{16.95}\sub{-74.88\%} \\

\midrule

\multirow{7}{*}{\shortstack{LongLive\\1.3B\\(30-second)}}
 & Baseline   & BF16 & 91.44 & 93.83 & 98.94 & 44.50 & 60.74 & 80.92 & 3.75 & 3.25 & 3.69 & 64.28\sub{bs=20} \\
\cmidrule(lr){2-13}
 & FlexGen    & INT6 & \textbf{91.33} & \underline{93.49} & \textbf{98.95} & \underline{46.50} & \underline{60.29} & 80.60 & \underline{3.78} & \textbf{3.33} & \textbf{3.78} & \underline{26.62}\sub{-58.59\%} \\
 & KIVI       & INT6 & 91.14 & 93.45 & \underline{98.94} & 43.50 & \textbf{60.36} & \underline{80.65} & 3.73 & 3.27 & 3.70 & 26.88\sub{-58.18\%} \\
 & \ourmethod & INT6 & \underline{91.26} & \textbf{93.57} & 98.93 & \textbf{47.50} & 60.17 & \textbf{80.94} & \textbf{3.84} & \underline{3.31} & \underline{3.71} & \textbf{26.19}\sub{-59.25\%} \\

 & FlexGen    & INT4 & \underline{86.72} & 90.19 & 95.83 & 45.00 & 55.16 & 76.59 & \underline{3.48} & 3.48 & 3.77 & \underline{16.57}\sub{-74.22\%} \\
 & KIVI       & INT4 & 85.83 & \textbf{92.09} & \underline{97.80} & \textbf{49.50} & \underline{56.59} & \underline{78.95} & \underline{3.48} & \underline{3.55} & \underline{3.79} & 16.84\sub{-73.81\%} \\
 & \ourmethod & INT4 & \textbf{88.39} & \underline{91.97} & \textbf{98.80} & \underline{47.50} & \textbf{57.11} & \textbf{79.60} & \textbf{3.54} & \textbf{3.57} & \textbf{3.80} & \textbf{16.15}\sub{-74.88\%} \\

\bottomrule
\end{tabular}
\end{table*}

%% file: Tables/AblationStudy_Group.tex
\begin{table*}[t]
\centering
\caption{Grouping ablation at INT4 on next-scale VAR-d30 and next-frame Self-Forcing. \ourmethod~(full) is the proposed method. Shaded rows ablate one axis; the bottom rows compare to LLM-style alternatives. Bold marks the best in each column.}
\label{tab:grouping_ablation}
\scriptsize
\setlength{\tabcolsep}{3pt}
\renewcommand{\arraystretch}{1.05}
\begin{tabular}{lcccccc|ccccc}
\toprule
\multirow{2}{*}{Grouping} & \multicolumn{6}{c|}{VAR-d30 (next-scale)} & \multicolumn{5}{c}{Self-Forcing (next-frame)} \\
\cmidrule(lr){2-7} \cmidrule(lr){8-12}
& FID$\downarrow$ & IS$\uparrow$ & PSNR$\uparrow$ & SSIM$\uparrow$ & LPIPS$\downarrow$ & KV (GB)$\downarrow$ & Subject$\uparrow$ & Dynamic$\uparrow$ & Aesthetic$\uparrow$ & Total$\uparrow$ & KV (GB)$\downarrow$ \\
\midrule
\ourmethod~(full) & \textbf{1.97} & \textbf{323.67} & \textbf{13.54} & \textbf{0.312} & \textbf{0.437} & 10.44\sub{-72.06\%} & \textbf{91.28} & \textbf{75.50} & \textbf{57.40} & \textbf{80.99} & 16.95 \\
\space \rowcolor{gray!12} w/o block-axis & 2.83 & 292.13 & 12.07 & 0.257 & 0.538 & 9.45\sub{-74.71\%} & 88.49 & 69.00 & 54.51 & 79.38 & 16.90 \\
\space \rowcolor{gray!12} w/o channel-axis & 12.07 & 159.52 & 10.42 & 0.198 & 0.645 & \textbf{9.34}\sub{-75.00\%} & 79.17 & 27.00 & 33.86 & 67.22 & \textbf{16.87}\\
\midrule
\multicolumn{12}{l}{\textit{Additional baselines (LLM-style groupings):}} \\
Per-token \& per-head & 2.21 & 292.12 & 11.97 & 0.261 & 0.545 & 10.51\sub{-71.87\%} & 90.18 & 67.00 & 54.96 & 79.46 & 17.93 \\
Per-token & 7.13 & 203.50 & 10.51 & 0.208 & 0.635 & 9.38\sub{-74.90\%} & 78.55 & 25.50 & 39.51 & 67.72 & 16.96 \\
\bottomrule
\end{tabular}
\end{table*}

%% file: Table/boundry_ablation.tex
\begin{wraptable}{r}{0.4\columnwidth}
\centering
\vspace{-10pt}
\caption{Boundary alignment ablation.}
\label{tab:boundary-ablation}
\scriptsize
\setlength{\tabcolsep}{3pt}
\renewcommand{\arraystretch}{1.05}
\begin{tabular}{l*{4}{c}}
\toprule
Grouping & Subj $\uparrow$ & Dyn $\uparrow$ & Aes $\uparrow$ & Quality $\uparrow$ \\
\midrule
Aligned    & \textbf{91.23} & \textbf{74.50} & \textbf{57.41} & \textbf{80.91} \\
Misaligned & 90.80 & 73.50 & 56.75 & 80.46 \\
\bottomrule
\end{tabular}
\vspace{-10pt}
\end{wraptable}

%% file: Algorithms/pack.tex
\begin{algorithm}
\caption{VAR-Q Quantization}
\label{alg:varq-quant}
\DontPrintSemicolon
\KwIn{KV tensor $X$, bit-width $b$, quantization method $M$, qkv layout $F$, packing flag $\mathrm{pack}$}
\KwOut{quantized tensor $\widehat{Q}$, scale $S$, quantization metadata $m_q$, packing metadata $m_p$}

$q_{\min} \leftarrow -2^{b-1}$,\quad $q_{\max} \leftarrow 2^{b-1}-1$\;
$D \leftarrow \textsc{ReduceDims}(M,F)$\;

$S \leftarrow \max(|X|,\mathrm{dim}=D,\mathrm{keepdim}=\mathrm{True}) / q_{\max}$\;
$S \leftarrow \max(S,\epsilon)$\;

$Q \leftarrow \mathrm{round}(X/S)$\;
$Q \leftarrow \mathrm{clamp}(Q,q_{\min},q_{\max})$\;
$Q \leftarrow \mathrm{int8}(Q)$\;

\eIf{$\mathrm{pack}$}{
    $(\widehat{Q},m_p) \leftarrow \textsc{PackLastDim}(Q,b)$\;
}{
    $\widehat{Q} \leftarrow Q$,\quad $m_p \leftarrow \varnothing$\;
}

$m_q \leftarrow (M,F,\mathrm{shape}(X))$\;
\Return{$\widehat{Q},S,m_q,m_p$}\;
\end{algorithm}

\begin{algorithm}
\caption{VAR-Q Dequantization}
\label{alg:varq-dequant}
\DontPrintSemicolon
\KwIn{quantized tensor $\widehat{Q}$, scale $S$, quantization metadata $m_q$, packing metadata $m_p$, output dtype $\tau$}
\KwOut{dequantized tensor $\widetilde{X}$}

\eIf{$m_p \neq \varnothing$}{
    $Q \leftarrow \textsc{UnpackLastDim}(\widehat{Q},m_p)$\;
}{
    $Q \leftarrow \widehat{Q}$\;
}

$S \leftarrow \textsc{BroadcastScale}(S,m_q,\mathrm{shape}(Q))$\;
$\widetilde{X} \leftarrow \mathrm{Cast}_{\tau}\bigl(\mathrm{float32}(Q) \odot S\bigr)$\;

\Return{$\widetilde{X}$}\;
\end{algorithm}

The full quantization stack consists of four operators (Algorithms~\ref{alg:varq-quant}--\ref{alg:varq-unpack}) that together turn the per-(channel, block) grouping rule of Section~\ref{sec:algorithm} into actual KV-memory savings on GPU. Algorithm~\ref{alg:varq-quant} performs uniform symmetric quantization on each newly emitted block: the reduction axes $D$ are derived from the quantization method $M$ and qkv layout $F$ so that one scaling factor is computed per quantization unit, an $\epsilon$ floor guards against numerically zero groups, and the rounded integers are clamped to the signed range $[-2^{b-1}, 2^{b-1}-1]$ and held in \texttt{int8}. Algorithm~\ref{alg:varq-dequant} is the symmetric inverse: it broadcasts $S$ back to the cache shape recorded in $m_q$, multiplies in float32, and casts to the requested attention dtype $\tau$. Both operators are stateless with respect to other blocks, which matches the online property of the algorithm: a block is quantized once at write time and never revisited.

Quantization alone does not shrink the cache, because current CUDA toolchains expose no native sub-8-bit storage type. Algorithms~\ref{alg:varq-pack} and~\ref{alg:varq-unpack} close this gap by packing $w = \lfloor 32/b \rfloor$ signed $b$-bit values into a single \texttt{uint32} word along the last (channel) dimension. We support packed storage for INT8, INT6, INT4, INT3, and INT2: INT8, INT4, and INT2 exactly fill a 32-bit word, whereas INT6 and INT3 leave two unused bits per word, yielding effective storage ratios of $40\%$ and $20\%$ relative to BF16, respectively. The packer pads the channel axis up to a multiple of $w$, masks each lane to $b$ bits, and ORs the lanes into their assigned slots via the shifts $kb$ for $k = 0, \ldots, w-1$; the original channel count $C_{\mathrm{orig}}$, pad amount $p$, lane count $w$, and bit-width $b$ are recorded in $m_p$ so that the unpacker can shift, mask, sign-extend, and trim back to the exact original shape. This converts the nominal bit-width reduction into a real cache-byte reduction, and the lane layout is byte-aligned so loads remain coalesced.

The four operators compose into the runtime pipeline expected by visual AR inference. At write time, a freshly produced block is passed through Algorithm~\ref{alg:varq-quant} with $\mathrm{pack}=\mathrm{True}$, which internally invokes Algorithm~\ref{alg:varq-pack}; the resulting \texttt{uint32} tensor, scale, and metadata $(m_q, m_p)$ are appended to the cache. At attention time, Algorithm~\ref{alg:varq-dequant} is invoked on the cached blocks, calling Algorithm~\ref{alg:varq-unpack} only when $m_p \neq \varnothing$ and otherwise falling back to a plain \texttt{int8} dequant path. The whole stack is implemented in Triton and runs without calibration data, retraining, or model-specific kernel variants, so it drops into Infinity, InfinityStar, Self-Forcing, LongLive, and VAR with no code changes outside the cache module.

\begin{algorithm}[t]
\caption{Pack Last Dimension into 32-bit Words}
\label{alg:varq-pack}
\DontPrintSemicolon
\KwIn{quantized tensor $Q$ stored in int8, bit-width $b$}
\KwOut{packed tensor $P$ stored in uint32, packing metadata $m_p$}

$C_{\mathrm{orig}} \leftarrow \mathrm{LastDim}(Q)$\;
$w \leftarrow 32/b$ \tcp*{number of $b$-bit values per 32-bit word}
$p \leftarrow (-C_{\mathrm{orig}}) \bmod w$\;
$C_{\mathrm{pad}} \leftarrow C_{\mathrm{orig}} + p$\;
$C_w \leftarrow C_{\mathrm{pad}}/w$\;

$Q' \leftarrow \textsc{PadLastDim}(Q,p)$\;
$Q' \leftarrow \textsc{ReshapeLastDim}(Q',[C_w,w])$\;

$P \leftarrow 0 \in \mathrm{uint32}^{\mathrm{shape}(Q)[:-1] \times C_w}$\;

\For{$k \leftarrow 0$ \KwTo $w-1$}{
    $U_k \leftarrow Q'[...,k] \;\&\; ((1 \ll b)-1)$\;
    $P \leftarrow P \;|\; \bigl(\mathrm{uint32}(U_k) \ll (kb)\bigr)$\;
}

$m_p \leftarrow (C_{\mathrm{orig}},p,w,b)$\;
\Return{$P,m_p$}\;
\end{algorithm}

\begin{algorithm}[t]
\caption{Unpack 32-bit Words along the Last Dimension}
\label{alg:varq-unpack}
\DontPrintSemicolon
\KwIn{packed tensor $P$ stored in uint32, packing metadata $m_p=(C_{\mathrm{orig}},p,w,b)$}
\KwOut{recovered quantized tensor $Q$ stored in int8}

$mask \leftarrow (1 \ll b)-1$\;
$U \leftarrow 0 \in \mathrm{uint32}^{\mathrm{shape}(P) \times w}$\;

\For{$k \leftarrow 0$ \KwTo $w-1$}{
    $U[...,k] \leftarrow (P \gg (kb)) \;\&\; mask$\;
}

$Q' \leftarrow \textsc{SignExtend}(U,b)$\;
$Q' \leftarrow \textsc{FlattenLastTwoDims}(Q')$\;
$Q \leftarrow Q'[...,0:C_{\mathrm{orig}}]$\;

\Return{$Q$}\;
\end{algorithm}

%% file: Table/rope_and_bound.tex
\begin{wraptable}{r}{0.50\textwidth}
\centering
\caption{Pre-RoPE vs.\ post-RoPE quantization at INT6.}
\label{tab:pre_rope}
\scriptsize
\setlength{\tabcolsep}{2pt}
\vspace{-8pt}
\renewcommand{\arraystretch}{1.05}
\begin{tabular}{llccccc|c}
\toprule
\multirow{2}{*}{Variant} & \multirow{2}{*}{Prec.} & \multicolumn{5}{c|}{VBench~\cite{huang2024vbench} $\uparrow$} & \multirow{2}{*}{KV (GB)$\downarrow$} \\
\cmidrule(lr){3-7}
& & Subj. & Bg. & Dyn. & Aes. & Quality & \\
\midrule
post-RoPE & INT6 & \textbf{93.96} & \textbf{94.76} & \textbf{76.50} & \textbf{61.25} & \textbf{83.34} & 27.50\sub{-59.25\%} \\
\rowcolor{gray!12} pre-RoPE & INT6 & 93.63 & 93.97 & 74.50 & 60.00 & 82.93 & 27.50\sub{-59.25\%} \\
\bottomrule
\vspace{-18pt}
\end{tabular}
\end{wraptable}

%% file: Video_prompts.tex
\begingroup
\scriptsize
\setlength{\tabcolsep}{3pt}
\renewcommand{\arraystretch}{1.08}
\setlength{\LTleft}{0pt}
\setlength{\LTright}{0pt}
\emergencystretch=2em

\begin{longtable}{@{}r r p{0.74\textwidth}@{}}
\caption{Complete list of the 200 prompts used for MovieGen-based evaluation. The original line number is 1-based in the full MovieGenVideoBench prompt file.}
\label{tab:moviegen-200-prompts} \\
\toprule
\textbf{ID} & \textbf{Original Line} & \textbf{Prompt} \\
\midrule
\endfirsthead

\multicolumn{3}{c}{\tablename~\thetable{} -- continued from previous page} \\
\toprule
\textbf{ID} & \textbf{Original Line} & \textbf{Prompt} \\
\midrule
\endhead

\midrule
\multicolumn{3}{r}{Continued on next page} \\
\endfoot

\bottomrule
\endlastfoot

0 & 621 & A person rowing a boat through a river of liquid gold, with shimmering banks. \\
1 & 117 & Over the shoulder shot of a woman running and watching a rocket in the distance. \\
2 & 705 & A trampoline surface bending under the weight of a person jumping on it, then springing back up as they jump off. \\
3 & 338 & A woman's face twisted in disgust when she tasted the spoiled food. \\
4 & 50 & A little man with blocks visiting an art gallery \\
5 & 331 & A man's eyes filled with tears of frustration after failing the exam. \\
6 & 394 & a man and a woman in their 20s are dining in a futuristic restaurant materialized out of nanotech and ferrofluids \\
7 & 513 & A deep underground cave filled with glowing crystals and hidden treasures. \\
8 & 172 & A spaceship being pulled into a blackhole. \\
9 & 98 & A middle-aged sad bald man becomes happy as a wig of curly hair and sunglasses fall suddenly on his head. \\
10 & 850 & A tracking shot following a skateboarder performing tricks down a city street, keeping pace with their fluid movements. \\
11 & 906 & a toy robot wearing purple overalls and cowboy boots taking a pleasant stroll in Mumbai India during a colorful festival \\
12 & 923 & a woman wearing blue jeans and a white t shirt taking a pleasant stroll in Mumbai India during a beautiful sunset \\
13 & 378 & Gold coins are falling out when elevator door opens \\
14 & 756 & A low-angle view of a majestic lion standing on a rocky outcrop, looking regal and powerful against the horizon. \\
15 & 984 & an old man wearing blue jeans and a white t shirt taking a pleasant stroll in Antarctica during a colorful festival \\
16 & 305 & A gardener planting seeds in a garden bed, their hands gently pressing the soil over the seeds. \\
17 & 440 & A princess blowing out birthday candles on a cake. \\
18 & 920 & a toy robot wearing a green dress and a sun hat taking a pleasant stroll in Antarctica during a beautiful sunset \\
19 & 834 & A push-in through an open window, entering a cozy room lit by the warm glow of a fireplace. \\
20 & 901 & a toy robot wearing blue jeans and a white t shirt taking a pleasant stroll in Johannesburg South Africa during a winter storm \\
21 & 875 & Macro shot. Man in an antique scuba helmet with dark glass walking out of a flower \\
22 & 991 & an old man wearing purple overalls and cowboy boots taking a pleasant stroll in Johannesburg South Africa during a winter storm \\
23 & 738 & A single ice cube placed in a warm drink, slowly melting and sending gentle ripples through the liquid as it transforms. \\
24 & 407 & A surreal collage of a whirlwind of colorful fabrics and clothing items, fluttering and swirling in mid-air. The scene is dynamic and fashionable, with vibrant textile patterns. A sense of motion and style create a visually striking and complex scene. Pitch black background. \\
25 & 883 & A mother dog gently picks up a piece of meat and carefully places it in her puppy's bowl, her eyes filled with warmth and care as she watches her little one eat. \\
26 & 14 & A beautiful homemade video showing the people of Lagos, Nigeria in the year 2056. Shot with a mobile phone camera. \\
27 & 785 & A pan left across a serene beach at sunrise, moving from the darkened shore to the brightening horizon. \\
28 & 231 & The person's nose wrinkled in distaste as he smelled something unpleasant. \\
29 & 496 & A quaint village nestled in a valley is surrounded by blooming cherry blossoms, with petals drifting through the air as villagers go about their daily activities, adding life to the scene. \\
30 & 424 & Close-up of a man eating an apple. \\
31 & 701 & A rubber band being stretched to its maximum length and then released, snapping back to its original shape. \\
32 & 403 & Cinematic trailer for a group of adventurous puppies exploring ruins in the sky \\
33 & 908 & a toy robot wearing purple overalls and cowboy boots taking a pleasant stroll in Johannesburg South Africa during a beautiful sunset \\
34 & 828 & A zoom-in on a drop of morning dew on a leaf, showing the reflection of the surrounding world within it. \\
35 & 927 & a woman wearing blue jeans and a white t shirt taking a pleasant stroll in Johannesburg South Africa during a colorful festival \\
36 & 933 & a woman wearing purple overalls and cowboy boots taking a pleasant stroll in Mumbai India during a colorful festival \\
37 & 177 & A softball player sliding safely into second base. \\
38 & 35 & Five gray wolf pups frolicking and chasing each other around a remote gravel road, surrounded by grass. The pups run and leap, chasing each other, and nipping at each other, playing. \\
39 & 301 & A person is rapidly typing on a keyboard \\
40 & 313 & A woman is search her bag trying to find something. \\
41 & 398 & a man BASE jumping over tropical hawaii waters. His pet macaw flies alongside him \\
42 & 728 & A crystal ball sitting on a table, with sunlight streaming through it and casting a circle of rainbow colors on the floor. \\
43 & 892 & A girl is peeling an orange \\
44 & 173 & a real girl franatically running a dense forest with bushes, trees, in rainy day, the animals are running after her and she is screaming and shouting \\
45 & 843 & A pull-out from a child’s hands holding a small seashell, moving back to show the beach and the waves around them. \\
46 & 590 & A panda in a scientist's lab coat, conducting experiments with beakers and test tubes. \\
47 & 831 & A push-in through a garden archway, revealing a secret, tranquil garden filled with blooming flowers. \\
48 & 631 & A person holding an umbrella that turns rain into colorful confetti. \\
49 & 622 & A person playing a harp strung with rainbows, creating music that colors the air. \\
50 & 260 & In a vibrant theater, a magician in dazzling attire stands center stage, pulling a comically oversized rubber chicken from an ornate, old-fashioned box. His costume shimmers under the stage lights, adding to the spectacle. The crowd erupts in laughter and applause, their faces filled with joy and amazement. The magician's expression hints at mischievous delight as he holds up the rubber chicken, his performance bringing cheer to the audience. \\
51 & 296 & A man is putting a ring on a woman's finger \\
52 & 783 & An aerial shot of a field of blooming wildflowers, creating a patchwork of colors in the landscape. \\
53 & 750 & An arc shot around a couple standing under a cherry blossom tree, petals falling around them as they embrace. \\
54 & 673 & A cyclist accelerating out of the saddle during a steep climb. \\
55 & 968 & an adorable kangaroo wearing a green dress and a sun hat taking a pleasant stroll in Mumbai India during a beautiful sunset \\
56 & 460 & A video of a diver creating bubbles underwater, with bubbles rising and interacting with each other. \\
57 & 151 & A chimp on the sidewalk doing a backflip on a skateboard. \\
58 & 752 & An arc shot around a lone tree in a vast, foggy field at dawn, revealing the changing light and shadows. \\
59 & 279 & A bunny puts the bright moon on its back and flies into the distance. \\
60 & 409 & A dynamic motion shot of a broom morphing surreal and magically into a peacock. The handle shortens and curves into a slender neck, the bristles fanning out into a magnificent tail. The camera moves around as vibrant colors and eye-shaped patterns emerge on the expanding feathers. A small head forms at the top, complete with a delicate crest. The transformation completes as the peacock proudly displays its newly formed plumage. \\
61 & 134 & a cactus with googly eyes dancing in the breeze \\
62 & 306 & A pair of hands skillfully knitting a colorful scarf, the yarn winding through their fingers with each stitch. \\
63 & 283 & A car crashes into a barrier at high speed.  \\
64 & 323 & A man is sipping a hot cup of coffee, steam rising from the mug. \\
65 & 91 & A cinematic wide portrait of a man with his face lit by the glow of a TV. \\
66 & 777 & An over-the-shoulder shot of a photographer adjusting their camera, framing a beautiful sunset scene. \\
67 & 374 & A man is running in the forest and transforms into a wolf. \\
68 & 652 & A person on a hoverboard colliding with a wall, the board stopping abruptly. \\
69 & 861 & Camera tracking shot. An early 19th century scuba diver with a huge iron helmet and an iron body suit lounges on an antique lawn chair. The light is diffused and gray, casting soft shadows along the scene. The diver brings a martini glass to his helmet and puts it back down. The year is 1912. The diver is in a grassy tree-filled park. People in period-accurate dress mill around, wearing long dresses and suits, holding parasols. The diver's suit is burnished and old, held together with rusted bolts. The diver tips the martini toward his helmet and clinks the glass against the glass. The scene should feel serene and beautiful, evoking the feeling of an impressionist painting. \\
70 & 114 & Aerial shot of the ocean. a maelstrom forms in the water swirling around until it reveals the fiery depths below. \\
71 & 604 & A person playing chess with a robot on a floating platform above the ocean. \\
72 & 910 & a toy robot wearing purple overalls and cowboy boots taking a pleasant stroll in Johannesburg South Africa during a winter storm \\
73 & 790 & A truck left through a bustling city street, following the flow of traffic and pedestrians during rush hour. \\
74 & 965 & an adorable kangaroo wearing purple overalls and cowboy boots taking a pleasant stroll in Antarctica during a beautiful sunset \\
75 & 42 & This close-up shot of a chameleon showcases its striking color changing capabilities. The background is blurred, drawing attention to the animal’s striking appearance. \\
76 & 299 & A woman is ascending to the sky from the ground \\
77 & 369 & A paraglider descending to a landing zone. \\
78 & 137 & A police helicopter hovers above a high-speed chase, guiding officers on the ground to apprehend a suspect. \\
79 & 385 & a giant duck walks through the streets in Boston \\
80 & 832 & A push-in towards a lone figure standing at the edge of a cliff, overlooking a vast, fog-covered valley. \\
81 & 146 & A cook flipping pancakes on a griddle. \\
82 & 139 & a woman jumps over a bear \\
83 & 784 & An aerial view of a snow-covered mountain range, with the peaks and valleys forming intricate patterns in the snow. \\
84 & 978 & an old man wearing blue jeans and a white t shirt taking a pleasant stroll in Mumbai India during a colorful festival \\
85 & 297 & A man is playing the drums under the water \\
86 & 308 & A person using a screwdriver to assemble a piece of furniture, carefully tightening each screw. \\
87 & 220 & The person's lips curled up in a sly grin as he shared a secret joke. \\
88 & 510 & A bustling ancient marketplace with merchants selling spices and fabrics. \\
89 & 408 & A dynamic motion shot of a lamp transforming into a flamingo. The curved neck of the lamp elongates, its shade flattening into a delicate head. The camera circles as the base splits into two spindly legs, the bulb socket becoming a beak. Pink hues wash over the metal surface, transforming into soft feathers. The power cord coils and disappears as the transformation completes, revealing a graceful flamingo balancing on one leg. \\
90 & 207 & A man is eating spaghetti \\
91 & 821 & A pedestal down from the top of a waterfall, descending to show the pool of water and mist at its base. \\
92 & 104 & A slow cinematic push in on an ostrich standing in a 1980s kitchen. \\
93 & 77 & A FPV shot zooming through a tunnel into a vibrant underwater space. \\
94 & 566 & A fish driving a tiny submarine, exploring an underwater city. \\
95 & 443 & A woman drinking water from a glass, her lips touching the rim. \\
96 & 107 & A cyclone of broken glass in an urban alleyway. dynamic movement. \\
97 & 325 & A singer is belting out a high note on stage. \\
98 & 81 & Zooming in hyper-fast to a dandelion to reveal macro dream-like abstract world. \\
99 & 737 & The slow crystallization of a water droplet turning into ice on a frosty morning, with delicate patterns forming across its surface. \\
100 & 155 & A chef drizzling sauce onto a plate with precision. \\
101 & 856 & A little boy is riding a dragon in the sky to a castle \\
102 & 400 & a dark neon rainforest aglow with fantastical fauna and animals \\
103 & 452 & A young boy is playing a harmonica at sunset, with his dog sitting quietly beside him, listening. \\
104 & 286 & The tall skyscapers in Hong Kong suddenly transform into a moving Gundam robot, cinematic CGI \\
105 & 969 & an adorable kangaroo wearing a green dress and a sun hat taking a pleasant stroll in Mumbai India during a colorful festival \\
106 & 333 & A man sighed in relief as the doctor delivered the good news. \\
107 & 390,695 & pov footage of an ant navigating the inside of an ant nest \\
108 & 836 & A zoom-out from a detailed shot of an intricate snowflake, pulling back to show a snowy landscape. \\
109 & 228 & The person's mouth twisted in disgust as he tasted something bitter. \\
110 & 232 & The person's forehead furrowed in concern as he listened to a friend's problems. \\
111 & 803 & A truck right through a tranquil garden, moving past blooming flowers, trees, and a small fountain. \\
112 & 420 & A marble goes through a glass cup, breaking it into pieces. \\
113 & 937 & a woman wearing purple overalls and cowboy boots taking a pleasant stroll in Johannesburg South Africa during a winter storm \\
114 & 51 & A white cat driving in a car through a busy downtown street with tall buildings and pedestrians in the background \\
115 & 335 & A man looked away in shame when confronted with his wrongdoing. \\
116 & 499 & Robots move efficiently through a futuristic laboratory, adjusting holographic displays and conducting experiments, while scientists observe and interact with the high-tech equipment. \\
117 & 357 & A high-speed video of a soap bubble popping, with the soapy liquid dispersing in all directions. \\
118 & 918 & a toy robot wearing a green dress and a sun hat taking a pleasant stroll in Johannesburg South Africa during a colorful festival \\
119 & 699 & A magician takes off his performing mask. \\
120 & 271 & Penguins roller skating \\
121 & 879 & An astronaut fighting a monster \\
122 & 493 & In an ancient library, books float and glow as they drift through the air, occasionally landing softly on the tables, where curious individuals reach out to read their contents. \\
123 & 164 & A chef skillfully tossing a salad in a bowl. \\
124 & 891 & A mother is peeling an apple for her daughter \\
125 & 106 & Zooming through a dark forest with neon light flora lighting up. \\
126 & 565 & A horse wearing roller skates, gracefully gliding through a city park. \\
127 & 215 & The person walked into the room, his face lighting up with a warm smile. \\
128 & 193 & A pair of lovebirds preening each other's feathers. \\
129 & 634 & A person skydiving from a hot air balloon into a sea of clouds. \\
130 & 46 & Tiltshift of a construction site filled with workers, equipment, and heavy machinery. \\
131 & 444 & A woman singing softly to a baby, her lips forming gentle words. \\
132 & 156 & A frog that gets kissed and turns into a chocolate milkshake. \\
133 & 581 & A kangaroo wearing boxing gloves, sparring with a punching bag in a gym. \\
134 & 131 & Hyperrealistic monster that closes its mouth \\
135 & 458 & A time-lapse video of a river flowing through a forest, with changing water levels and currents. \\
136 & 797 & A pan right across a field of tall grass swaying gently in the wind, with a setting sun in the background. \\
137 & 725 & Sunlight piercing through a canopy of bamboo, casting long, linear shadows and patches of light on the forest floor. \\
138 & 574 & A peacock wearing a crown, sitting on a throne and holding court with other animals. \\
139 & 90 & A FPV drone shot through a castle on a cliff. \\
140 & 38 & A grandmother with neatly combed grey hair stands behind a colorful birthday cake with numerous candles at a wood dining room table, expression is one of pure joy and happiness, with a happy glow in her eye. She leans forward and blows out the candles with a gentle puff, the cake has pink frosting and sprinkles and the candles cease to flicker, the grandmother wears a light blue blouse adorned with floral patterns, several happy friends and family sitting at the table can be seen celebrating, out of focus. The scene is beautifully captured, cinematic, showing a 3/4 view of the grandmother and the dining room. Warm color tones and soft lighting enhance the mood. \\
141 & 774 & A close-up of a key turning in a lock, showing the subtle movements of the key and the intricate details of the mechanism as it turns into place. \\
142 & 23 & A cat waking up its sleeping owner demanding breakfast. The owner tries to ignore the cat, but the cat tries new tactics and finally the owner pulls out a secret stash of treats from under the pillow to hold the cat off a little longer. \\
143 & 782 & An aerial perspective of a group of dolphins swimming near the surface of a crystal-clear ocean, their movements synchronized. \\
144 & 596 & A person walking up a staircase made of clouds leading to a floating castle. \\
145 & 322 & A little child let out a big yawn \\
146 & 863 & Over the shoulder camera shot. A huge lizard creature sits in a midcentury orange swivel chair. The light is dim and volumetric, casting an eerie glow across the scene. The creature uses its arms to maniacally push buttons on a gigantic control panel. Above the control panel is a panoramic window looking out and down on 1940s new york city. The room should invoke midcentury science fiction aesthetics, like rusty orange colors, bright flashing control buttons, and space-age flair. As the creature continues to quickly push buttons, the New York City scene out of the window moves closer, as though the creature is in a gigantic robot stomping through the city. The scene should give the feeling of frantic action, highlighting the intensity of piloting a giant robot. The scene should take inspiration from midcentury japanese monster films. \\
147 & 281 & Whilst a man and woman are walking through a city street in a dream, the man shows the woman how to fold the entire street upwards at a 90-degree angle and connect it with the sky. This creates a visually stunning effect, with the buildings and road bending and defying gravity. The scene highlights the limitless possibilities and creativity within the dream world. \\
148 & 764 & A first-person view of an artist sketching in a notebook, the pencil moving swiftly across the page as the drawing takes shape. \\
149 & 393 & an alien blending in naturally with new york city, paranoia thriller style, 35mm film \\
150 & 687 & A woman is playing a soft melody on his lute while sitting by the fountain in the castle courtyard. \\
151 & 830 & A push-in through a dense crowd at a festival, moving towards a performer on stage who is captivating the audience. \\
152 & 82 & Internal window of a train moving at hyper-speed in an old European city. \\
153 & 32 & New York City submerged like Atlantis. Fish, whales, sea turtles and sharks swim through the streets of New York. \\
154 & 199 & A martial artist breaking a board with a powerful punch. \\
155 & 153 & A fish walking into a coffee shop and asking for a cup of coffee. \\
156 & 872 & A computer mouse with legs running on a treadmill \\
157 & 206 & a woman eating ice scream \\
158 & 63 & Carefully pouring the milk into the cup, the milk flowed smoothly and the cup was gradually filled with a milky white color \\
159 & 184 & A soccer goalie making a diving save with outstretched arms \\
160 & 171 & A penguin flies into the mouth of a blue whale breaking the surface of the water. \\
161 & 471 & A close-up of a drink being poured over ice, capturing the detailed flow and interaction with the ice cubes. \\
162 & 128 & A curious cat peering out from a cozy hiding spot. \\
163 & 975 & an adorable kangaroo wearing a green dress and a sun hat taking a pleasant stroll in Antarctica during a colorful festival \\
164 & 257 & A bird made of fresh oranges rushes out of the orange \\
165 & 570 & A bear wearing a superhero cape, flying through the sky over a bustling city. \\
166 & 718 & Light filtering through a canopy of autumn leaves, casting warm, dappled patterns of yellow, orange, and red onto the ground. \\
167 & 654 & A person on a scooter colliding with a park bench, the scooter tipping over. \\
168 & 677 & A bobsled team accelerating down an icy track. \\
169 & 816 & A tilt-down from the towering peak of a mountain to the winding path leading up to it. \\
170 & 255 & Camera tracking shot. A man walking down a city street, holding a coffee cup in his hand. He is wearing a dark suit and red tie. \\
171 & 379 & A rose is growing out of a stone \\
172 & 886 & A mother bird feeding her chicks in the nest, delicately placing food into their wide-open beaks as they chirp eagerly. \\
173 & 217 & The person's eyebrows furrowed in concentration as he worked on a puzzle. \\
174 & 218 & The person's mouth dropped open in surprise as he watched a magic trick. \\
175 & 929 & a woman wearing blue jeans and a white t shirt taking a pleasant stroll in Antarctica during a beautiful sunset \\
176 & 578 & A fox wearing a pirate hat and eyepatch, steering a ship through a stormy sea. \\
177 & 823 & A pedestal down through a field of sunflowers, showing their tall stalks and bright petals against the sky. \\
178 & 337 & A man grinned with satisfaction after completing the challenging task. \\
179 & 747 & A thin sheet of ice on a lake cracking and breaking as the sun warms it, creating a mosaic of shifting patterns. \\
180 & 122 & A bird landing on water and turning into a fish. \\
181 & 384 & a photorealistic video of a butterfly that can swim navigating underwater through a beautiful coral reef \\
182 & 266 & A red panda taking a bite of a pizza \\
183 & 202 & A child's face lighting up with joy as they blow out the candles on their birthday cake. \\
184 & 191 & A person opens a book and turns it upside-down and characters from the book begin to fall out of it. \\
185 & 244 & The person jumps up and down excitedly, expressing happiness through dance moves. \\
186 & 208 & A person takes a big bite of a juicy burger, the meat and cheese filling his mouth. \\
187 & 935 & a woman wearing purple overalls and cowboy boots taking a pleasant stroll in Johannesburg South Africa during a beautiful sunset \\
188 & 431 & A man pushes away a huge stone with superhuman strength. \\
189 & 109 & Highly detailed close up of a bacteria. \\
190 & 573 & A goat operating a food truck, serving gourmet grilled cheese sandwiches to a line of animals. \\
191 & 472 & A mesmerizing video of a whirlpool forming in a sink as water drains. \\
192 & 97 & An astronaut walking between stone buildings. \\
193 & 610 & A person doing yoga on top of a giant lily pad in the middle of a serene pond. \\
194 & 636 & A person riding a giant tortoise through a desert of glass sand. \\
195 & 614 & A person playing hopscotch on the rings of Saturn. \\
196 & 938 & a woman wearing purple overalls and cowboy boots taking a pleasant stroll in Antarctica during a beautiful sunset \\
197 & 844 & A pull-out from a dancer’s feet moving gracefully, expanding to show the entire stage and audience. \\
198 & 463 & A close-up of paint being mixed, showing the detailed interaction of colors and textures. \\
199 & 442 & A person chewing on a pencil while deep in thought. \\

\end{longtable}
\endgroup